\documentclass[11pt]{article}

\usepackage[margin=1in]{geometry}
\usepackage[T1]{fontenc}
\usepackage[utf8]{inputenc}
\usepackage{lmodern}
\usepackage{amsmath,amssymb}
\usepackage{graphicx}
\usepackage{booktabs}
\usepackage{array}
\usepackage{longtable}
\usepackage{caption}
\usepackage[numbers,sort&compress]{natbib}
\usepackage{xcolor}
\usepackage{hyperref}

\hypersetup{
  colorlinks=true,
  linkcolor=blue,
  citecolor=blue,
  urlcolor=blue
}
\providecommand{\JournalTitle}[1]{#1}
\DeclareUnicodeCharacter{2212}{\ensuremath{-}}
\DeclareUnicodeCharacter{2248}{\ensuremath{\approx}}
\DeclareUnicodeCharacter{2010}{-}
\DeclareUnicodeCharacter{2011}{-}
\newcommand{\dropcap}[1]{#1}
\newcommand{\FloatBarrier}{\clearpage}
\newcommand{\sisetup}[1]{}
\newenvironment{threeparttable}{\begin{center}}{\end{center}}
\newenvironment{tablenotes}[1][]{\begin{minipage}{0.95\linewidth}\footnotesize\begin{list}{}{\leftmargin=0pt\itemindent=0pt\labelwidth=0pt\labelsep=0pt}}{\end{list}\end{minipage}}
\newcommand{\keywords}[1]{\par\noindent\textbf{Keywords:} #1}
\makeatletter
\newcommand{\cogsynDefineTcolorboxFallback}{%
  \define@key{cogsynbox}{breakable}[]{}%
  \define@key{cogsynbox}{title}{\def\cogsynboxtitle{##1}}%
  \define@key{cogsynbox}{colback}{\def\cogsynboxback{##1}}%
  \define@key{cogsynbox}{colframe}{\def\cogsynboxframe{##1}}%
  \define@key{cogsynbox}{fonttitle}{}%
  \define@key{cogsynbox}{arc}{}%
  \define@key{cogsynbox}{boxrule}{}%
  \newenvironment{tcolorbox}[1][]{%
    \par\medskip
    \begingroup
    \small
    \def\cogsynboxtitle{}%
    \def\cogsynboxback{blue!5!white}%
    \def\cogsynboxframe{blue!75!black}%
    \setkeys{cogsynbox}{##1}%
    \noindent{\color{\cogsynboxframe}\rule{\linewidth}{1pt}}\par
    \ifx\cogsynboxtitle\@empty\else
      \noindent\begingroup
      \setlength{\fboxsep}{4pt}%
      \colorbox{\cogsynboxframe}{%
        \parbox{\dimexpr\linewidth-2\fboxsep\relax}{\color{white}\strut\cogsynboxtitle}%
      }%
      \endgroup\par
    \fi
    \noindent{\color{\cogsynboxback}\rule{\linewidth}{3pt}}\par
    \vspace{-0.25\baselineskip}
    \begin{list}{}{%
      \setlength{\leftmargin}{1em}%
      \setlength{\rightmargin}{1em}%
      \setlength{\topsep}{0.25\baselineskip}%
      \setlength{\parsep}{0.25\baselineskip}%
      \setlength{\itemsep}{0pt}%
    }%
    \item\relax\ignorespaces
  }{%
    \end{list}%
    \vspace{-0.4\baselineskip}
    \noindent{\color{\cogsynboxback}\rule{\linewidth}{3pt}}\par
    \noindent{\color{\cogsynboxframe}\rule{\linewidth}{1pt}}\par
    \endgroup
    \medskip
  }%
}
\cogsynDefineTcolorboxFallback
\newsavebox{\cogsynsignificancebox}

\makeatother
\NewDocumentEnvironment{promptbox}{O{blue!5!white} O{blue!75!black} m}{%
  \begin{tcolorbox}[
    breakable,
    title={#3},
    colback=#1,
    colframe=#2,
    fonttitle=\bfseries,
    arc=2mm,
    boxrule=1pt,
  ]%
}{\end{tcolorbox}}

\title{AI models can predict and collaboratively \\ modulate human memory search}
\author{Eric Lacosse$^{1,2,*}$, Mariana Duarte$^{1,2,*}$, Graham Todd$^3$,\\
Peter M. Todd$^4$, and Daniel C. McNamee$^{1,2}$\\[0.75em]
\small $^1$Champalimaud Research, Centre for the Unknown, Lisbon, Portugal\\
\small $^2$Champalimaud Centre for Restorative Neurotechnology, Lisbon, Portugal\\
\small $^3$New York University Tandon, Brooklyn, New York, USA\\
\small $^4$Indiana University, Bloomington, IN, USA\\
\small $^*$Equal contributors\\
\small Correspondence: daniel.mcnamee@research.fchampalimaud.org}
\date{}

\begin{document}

\maketitle

\begin{abstract}
Large language models (LLMs) exhibit unprecedented natural language generation and many text-based problem-solving capabilities.
Indeed, in many language-based tasks, for example routine coding, these artificial intelligence models have reduced, or even eliminated, the need for human input.
But rather than replacing human cognitive effort, LLMs may instead serve as cognitive tools to extend human abilities, particularly when they are engaged in a task requiring open-ended conceptual exploration and creative ideation.
However, we are yet to understand how these models may enhance such generative human cognitive abilities in human--AI interactions.
In this study, we explore and evaluate the ability of LLMs to follow and enhance human mental trajectories during semantic memory search.
To test this, we use the semantic fluency task (SFT),
a classic cognitive paradigm requiring generative semantic memory retrieval that has long served to characterize convergent and divergent thinking in humans.
We demonstrate that an LLM's abilities to track and predict human memory trajectories in this task exceed those of other humans.
Building on these results, we provide evidence regarding how collaborative LLM memory partners may improve human semantic memory processing, beyond what is achieved in human–human collaborations.
This contributes a step toward designing collaborative human–AI engagements that amplify generative human cognitive abilities and bridge between diverse intelligences.
\end{abstract}


\dropcap{H}umans collaborate with one another in an effort to enhance their own mental explorations and enrich their creative thinking \cite{abraham_creative_2024}.
When children build upon each other's LEGO designs, jazz musicians trade melodic phrases, and crossword partners spark associative chains, e.g., from ``java'' to ``brew'', a critical component of human social intelligence is the ability to solve the challenging problem of tracking and aligning with each other's generative mental trajectories \cite{tollefsen_alignment_2013}.
In such collaborative cognitive settings, one person's idea, innovation, or concept constructed from memory triggers associations in another person's mind.
Such a stochastic interplay of thoughts generates a chain of discovery that can extend into regions of the collective cognitive map that neither individual would have reached alone \cite{tolman_cognitive_1948, schafer_navigating_2018}.
With the advent of new AI systems exhibiting unexpectedly strong cognitive capabilities \cite{zhouGeneralScalesUnlock2026, goes_pushing_2023, hubert_current_2024, lu_ai_2024}, it becomes possible to ask whether and how humans can use such systems, not to replace their thinking in some tasks \cite{ibrahim_measuring_2025}, but instead to enhance it via human--AI collaboration \cite{subramonyam_bridging_2024, schut_bridging_2025}.
As AI reshapes knowledge work, we are beginning to see hints of a new paradigm:
Rather than machines simply automating human tasks, more effective approaches may emerge from carefully orchestrated partnerships enabling humans and AIs to contribute based on their distinctive cognitive capabilities \cite{mcguire_establishing_2024}.

A core element in improving such collaborative interactions is achieving cognitive synergy.
We define cognitive synergy as the positive collaboration between partners in cognitive tasks resulting in higher performance than what both can achieve separately.
In this work, we examine aspects of human and AI performance in collaborative memory tasks which we believe are critical for designing interactions that enhance cognitive synergy.
We hypothesize that cognitive alignment, the capacity for another human or AI to attune to an individual's specific mental computations, is important to facilitate cognitive synergy.
Unlike traditional notions of AI alignment, which concern the congruence of goals and preferences \cite{zhi-xuan_beyond_2024}, cognitive alignment refers to the congruence of cognitive processes.
We operationalize and measure it as the ability to accurately predict the consequences of cognitive computations of another agent, such as predicting the next concept in a human's associative thought process \cite{zahedi_mental-model_2022}.
A high degree of predictive accuracy serves as an empirical proxy for a high degree of cognitive alignment.
We hypothesize that the capacity for this alignment may be the key to bridging underlying cognitive processes in support of better synergistic collaborative interactions \cite{tollefsen_alignment_2013}, i.e., cognitive synergy.
For example, consider a medical diagnosis setting.
AI (goal/preference) alignment would help ensure that an AI respects the physician’s goal of maximizing patient well-being and avoiding harm.
Cognitive alignment, in contrast, would involve structuring the AI’s reasoning so that its diagnosis and treatment search processes scaffold the physician’s own, e.g., correcting any potential cognitive biases leading to medical error \cite{gigerenzer_risk_2014}.
These factors may help foster the right kind of collaborative back-and-forth interaction, rather than simply outputting a final answer, which is underpinned by the AI's ability to identify where the human physician is currently ``positioned'' in their mental schema and where they will go next \cite{nour_charting_2025}.

So far, we have lacked empirical studies of human--AI collaboration in domains that rigorously characterize this kind of human cognitive processing in the detailed, dynamic, and quantifiable manner \cite{vaccaro_when_2024} necessary for investigating cognitive alignment.
Our study bridges this gap by using a classic cognitive task, the semantic fluency task (SFT) of memory search in semantic spaces, in a collaborative context with novel metrics to evaluate the ability of LLMs to enhance human performance.
We examine both human--human and human--AI interactions, enabling us to directly model and quantify measures of cognitive alignment between different types of collaborative dyads.
Building on earlier human collaborative search experiments \cite{mannering_modeling_2025, szary_patterns_2015, hinds_collaborative_2016}, we then test whether human--AI collaborations can enhance the process by which humans search for concepts in memory \cite{hills_optimal_2012, hills_foraging_2015}.

\begin{figure}[tbp]
\centering
\includegraphics[width=\linewidth,height=0.62\textheight,keepaspectratio]{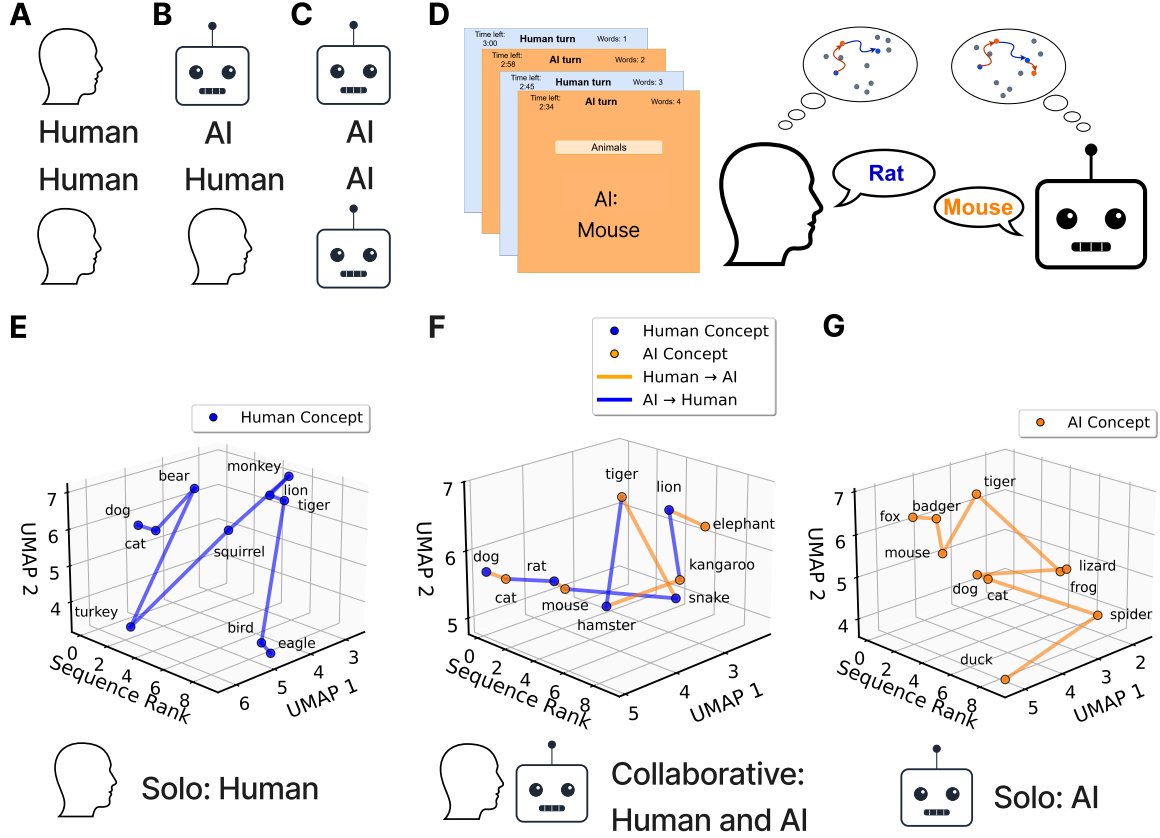}
\caption{Overview of collaborative conditions and experiment with resulting trajectories in semantic spaces. Three dyadic conditions are investigated: (A) Human--Human pairs performing collaborative SFT, (B) Human--AI collaborative pairs in SFT and (C) AI--AI pairs in SFT. (D) Trial run of an interleaved collaborative SFT showing human--AI interactions. The human retrieves a relevant concept from memory and produces the corresponding word in alternation with the AI. The left panel shows the sequential turn-based interface where users and an AI system exchange responses under the ``Animals'' category. The right panel illustrates the temporal structure of each 3-minute trial, including the sequential production (in thought-bubbles) of a word from the AI which then serves as a cue for human semantic retrieval and word production.  (E-G) Visualization of example SFT word sequence trajectories through semantic space (represented via UMAP dimensionality reduction) for (E) Human solo performance (starting at ``dog''), (F) human--AI collaborative performance (starting at ``dog'') showing bidirectional interactions in an interleaved condition where blue denotes human contributions and orange are AI, and (G) AI solo performance (starting at ``fox''). Each point represents a word, with connecting lines showing the temporal sequence of word generation. These trajectories reveal distinct patterns in how different agent configurations explore and exploit, diverge (large jumps) and converge (small jumps), within the semantic space during search.}
\label{fig:1}
\end{figure}

This task recruits and isolates a generative search mechanism that is crucial for human cognitive abilities in many important domains.
Specifically, to solve open-ended problems and create innovations, people need to access relevant concepts and ideas they have stored in semantic memory through a process of associative search \cite{hills_optimal_2012}.
Humans appear to search their memory for concepts akin to how animals forage for food and other resources in their environment \cite{charnov_optimal_1976, hills_optimal_2012}.
The SFT captures this by asking individuals to quickly produce a sequence of words representing concepts that fit within a specific semantic category (e.g., ``animals,'' ``foods,'' or ``occupations'') without repetition \cite{troyer_clustering_1997, troyer_cognitive_2006}.
Numerous SFT studies have shown that people tend to produce semantic ``clusters'' of related words, referred to as convergent semantic retrieval behavior, interspersed with ``switching'' to new clusters when retrieval slows, using a divergent retrieval mechanism \cite{troyer_clustering_1997}.
This strategy is marked by shorter response times within clusters and longer times between them \cite{hills_optimal_2012, charnov_optimal_1976}.
Recent neural evidence for strategically timed switches in memory search supports this foraging interpretation \cite{nour_trajectories_2023, lundin_neural_2023}.
While the standard SFT is performed in isolation, the collaborative version involves two participants alternating turns to contribute to a single, shared sequence.
This modification introduces both a key cognitive constraint and a synergistic opportunity: each participant must adapt to the words provided by their partner in order to avoid repetitions, while also having the opportunity to build upon their partner’s contributions, thereby reshaping their internal semantic search trajectory.

In collaborative versions of SFT, counter-intuitively, people working together often perform worse than individuals. This phenomenon is referred to as collaborative inhibition \cite{weldon_collective_1997,andersson_net_2001, marion_meta-analytic_2016, mannering_modeling_2025, szary_patterns_2015}.
In our terminology, collaborative inhibition could arise from two individuals performing the task while not being cognitively aligned, therefore precluding cognitive synergy.
Practically, this leads to a collaborative memory interference where one person's semantic trajectory disrupts their partner's path through their own idiosyncratic semantic space.
These results appear consistent with a more general phenomenon observed across many studies of human--AI collaboration, where human--AI pairs typically did not achieve synergy, but rather performed worse on average than either the best human or best AI working alone  \cite{vaccaro_when_2024}.
Even in coding tasks, a domain being radically reshaped by AI, truly synergistic collaborative interactions seem rare and challenging at present \cite{becker_measuring_2025}.
Understanding how these collaborative inhibition effects emerge \cite{weldon_collective_1997} and how to mitigate them is an important route to enhancing cognitive performance in AI-human partnerships.

We begin to address this challenge by testing whether LLMs can track and predict, i.e. cognitively align with, the conceptual trajectories generated by a human searching their semantic memory.
We do the same for humans tracking other humans, comparing humans and AI models in their cognitive alignment capabilities.
We provide results demonstrating how well LLM-generated SFT sequences effectively capture the general associative structure of human semantic memory at the population level.
Based on a newly created benchmark dataset, we next provide evidence of how powerful LLMs compare to humans in accurately predicting specific semantic search trajectories generated by other humans.
Furthermore, through a series of online real-time collaborative experiments, we show how human memory search is influenced when interacting with LLMs compared with other humans.
Together, these results indicate how well LLMs are capable of aligning to and influencing humans searching within their own individual internal space of concepts. Notably, although frontier LLMs achieve superhuman levels of human cognitive alignment, this does not lead to a reduction in collaborative inhibition. We suggest that the emergence of synergy depends on the structure of the interaction mechanism and is ultimately constrained by a strict alternation of exchange between partners. Accordingly, our investigations indicate that the interaction mechanism itself should be dynamically adapted to convert cognitive alignment into cognitive synergy, with the LLM positioned not merely as a participant, but as a supervisory system capable of reflecting on and reshaping the meta-dynamics of interaction.

Figure \ref{fig:1} provides an overview of our experimental paradigm, designed to systematically deconstruct the dynamics of cognitive search in various solo and collaborative configurations across both humans and AIs.
In the collaborative condition, the concepts generated by the ideal partner would provide a scaffold for the human player's search, guiding them to explore new semantic clusters or helping them more thoroughly exploit existing ones.

\section*{Results}

\begin{figure}[tbp]
\centering
\includegraphics[width=\linewidth,height=0.55\textheight,keepaspectratio]{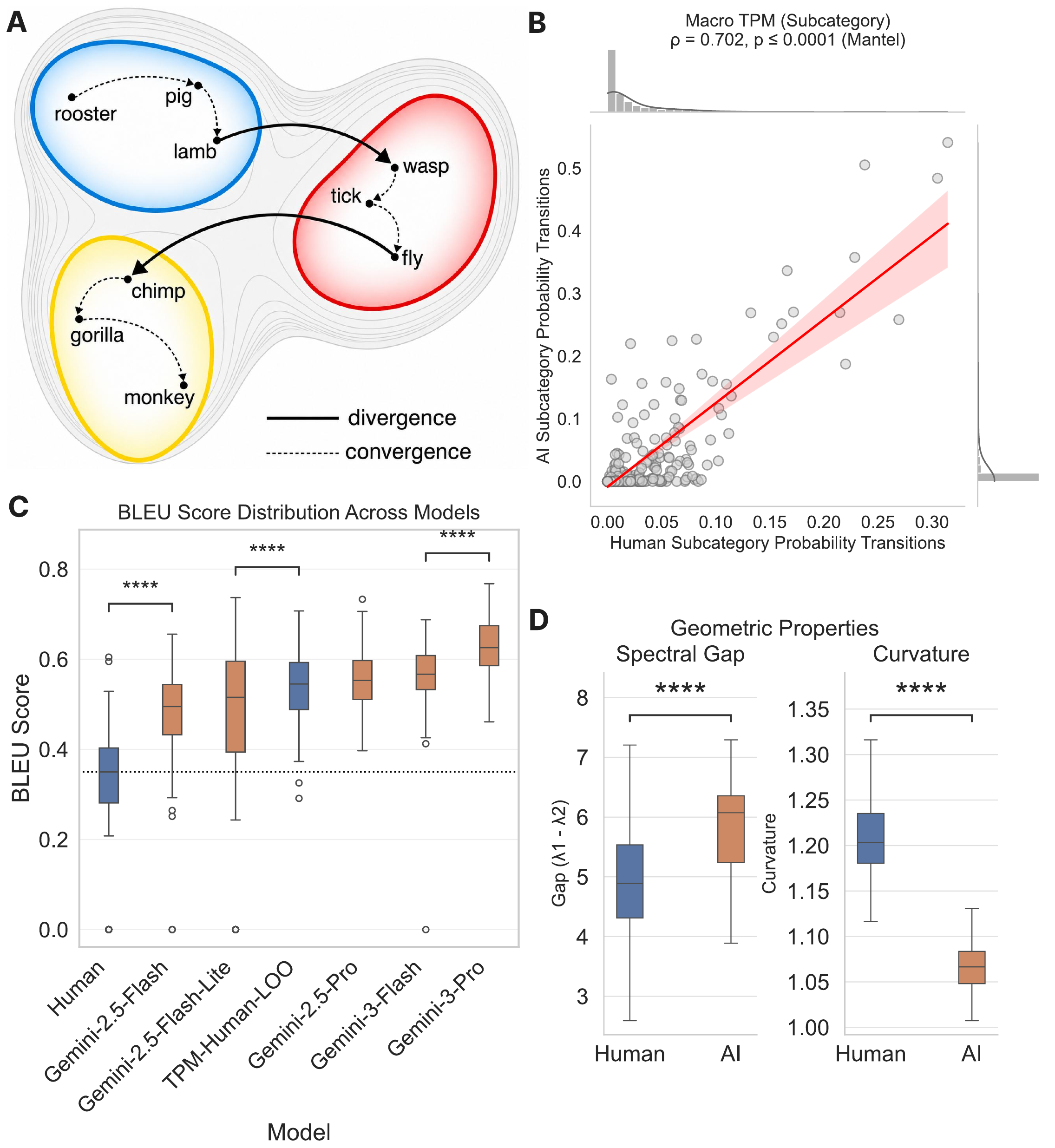}
\caption{\textbf{Cognitive Macro-Alignment: Human vs. LLM Semantic Transition Dynamics.} (A) A conceptual state transition diagram on a semantic manifold illustrates how a sequence is generated, distinguishing transitions as convergent (within-subcategory, e.g., ``rooster'' to ``pig'' in the farm-animal subcategory) or divergent (between-subcategory, e.g., ``lamb'' to ``wasp''). We refer to each retrieved semantic concept as a `concept', reflecting the interpretation of retrieval as navigation through a conceptual search space. (B) The scatter plot and regression line demonstrate a strong positive Spearman correlation ($\rho = 0.702, p < 0.0001$ (Mantel test)) between the human and LLM (\texttt{Gemini-3-Pro}) Transition Probability Matrices (TPMs), indicating that LLMs capture human-like patterns of semantic subcategory clustering and switching. Each TPM point represents the human vs. AI transition probabilities between two particular subcategories (SI~Figure~S2). (C) BLEU scores across human responses and model-generated sequences (boxplots show median and interquartile range; whiskers indicate 1.5$\times$IQR). One-sided paired Wilcoxon signed-rank tests between adjacent models indicate significant increases in alignment from Human--Human to Human--\texttt{Gemini-2.5-Flash} ($W=902$, $p<0.0001$), Human--\texttt{Gemini-2.5-Flash-Lite} to Human-\texttt{TPM-Human-LOO} ($W=3119$, $p<0.001$), and Human--\texttt{Gemini-3-Flash} to Human--\texttt{Gemini-3-Pro} ($W=1266$, $p < 0.0001$). (D) Geometric properties of semantic search trajectories. Comparison of human versus LLM sequences in semantic embedding space. The LLM (\texttt{Gemini-3-Pro}) exhibits a significantly larger Spectral Gap (Left), indicating a search structure that is more rigid and dominated by a single semantic axis ($5.779$ vs. $4.878$, $p < 0.0001$, two-sided Wilcoxon signed-rank). Conversely, humans display significantly higher Curvature (Right), reflecting a more volatile search path characterized by sharp, idiosyncratic associative turns compared to the ``smoother'' trajectory of the AI ($p < 0.0001$,  $1.207$ vs. $1.066$). Statistical significance is denoted as: ****~$p < 0.0001$.}
\label{fig:2}
\end{figure}
\subsection*{Cognitive Macro-Alignment}

For an AI to become an effective collaborative partner for humans, it may need to be able to emulate the organization and retrieval of concepts in memory in a human-like way \cite{rane_concept_2024}.
Here, we examine how well AIs can perform memory search in a manner that is compatible with human memory retrieval processes, specifically testing whether AI trajectories mirror human ones when instructed to do so (see Methods, Language Models).

First, we assess how closely, at the population-level, human- and AI-generated SFT sequences align with one another, which we term cognitive macro-alignment.
We use one of the currently most capable frontier reasoning LLMs (\texttt{Gemini-3-Pro}) to generate SFT sequences and compare them with a set of $n=141$ human-generated animal sequences from an earlier study \cite{hills_optimal_2012}.
Figure~\ref{fig:2} illustrates the degree of these similarities between LLM and human SFT sequences for one examined category (animals) parsed using standard subcategory norms which assign category words into standardized clusters \cite{zemla_snafu_2020}.

To evaluate the alignment of semantic sequence generation, we compared the transition dynamics between human participants and LLMs (Methods).
Our analysis examined convergent word transitions within subcategories and divergent transitions between subcategories, as illustrated conceptually in Figure \ref{fig:2}A.
To efficiently summarize sequence generation at a macro-cognitive level, a transition probability matrix (TPM) was calculated separately for sequences generated by humans and by LLMs.
Each TPM models Markov state transitions where nodes represent animal subcategories and edges represent transition probabilities, either staying in the same subcategory (diagonals) or switching to a different one (off-diagonals) (Figure \ref{fig:2}A and {SI~Figure~S2}).
The likelihood of switching from one subcategory to another was highly correlated (Figure \ref{fig:2}B) between humans and LLM (Spearman $\rho = 0.702$, $p < 0.0001$, Mantel Test,  SI Methods), indicating that the LLM successfully captures the general associative structure of human semantic memory according to established subcategory norms \cite{zemla_snafu_2020}.

To further quantify the similarity between AI-generated and human-generated semantic sequences at the level of word order, we used the BLEU (Bilingual Evaluation Understudy) metric \cite{papineni_bleu_2002} (Methods).
This n-gram overlap measure assesses how well a candidate sequence matches a set of reference sequences, with higher scores indicating greater similarity \cite{heineman_towards_2024} (SI Methods).
The LLM-generated sequences showed high similarity to human patterns of recall (Figure \ref{fig:2}C), substantially exceeding the average human-to-human similarity score (mean $M = 0.326$).
Remarkably, these results indicate that semantic sequences generated by these models are more representative of a typical human sequence than are randomly chosen human sequences.
Mean BLEU scores exhibited a clear upward trend scaling with model capability, increasing from human performance ($M=0.326$, 95\% CI $[0.304, 0.347]$) and peaking with \texttt{Gemini-3-Pro} ($M=0.626$, CI $[0.615, 0.637]$) (Figure \ref{fig:2}C).
To contextualize these results against population statistics, we also estimated a first-order TPM model on the human dataset using a leave-one-out (LOO) procedure (SI Methods).
This approximates an average human strategy by capturing population-level transition statistics.

While BLEU scores indicate high cognitive macro-alignment of LLM to human SFT sequences, geometric properties of human and LLM semantic trajectories reveal significant distinctions in their search strategies (Figure \ref{fig:2}D) consistent with earlier observations \cite{wang_fluency-based_2025, qiu_can_2025}.
After projecting SFT sequences into a suitable embedding space (SI Methods) obtained from ConceptNet \cite{speer_conceptnet_2018}, we find that LLM sequences generated by \texttt{Gemini-3-Pro} exhibit a significantly larger spectral gap (SG) than those of humans ($M = 5.779$, CI $[5.639, 5.911]$ vs. human $4.878$, CI $[4.736, 5.020]$), reflecting an AI search process that is more rigid and directed along a dominant semantic axis rather than the associative, multi-dimensional exploration we see in humans (Figure \ref{fig:2}D).
Additionally, greater trajectory curvature in human sequences ($1.207$, CI $[1.200, 1.214]$ vs. LLM $1.066$, CI $[1.062, 1.071]$) indicates that humans navigate via sharper turns through the high-dimensional embedding space, whereas LLMs follow a smoother and more ballistic path.
Consequently, the LLMs' superior BLEU scores may arise not from replicating the noisy dynamics of individual human thought, but from leveraging a statistical amalgamation collapsing individual sequence variance onto canonical high-probability trajectories.
This may explain why LLMs switch between subcategories considerably less often than humans do ({SI~Figure~S3}).

In sum, our analyses reveal a macro-cognitive trade-off: while LLMs can accurately predict the order of concepts in human sequences at the word level (higher BLEU), their embedded semantic trajectories are geometrically less human-like (higher spectral gap, lower curvature, fewer subcategory switches).
This is consistent with LLMs approximating a canonical population-averaged path rather than the noisy idiosyncratic dynamics of any individual human memory forager.

\subsection*{Cognitive Micro-Alignment}

\begin{figure}
\centering
\includegraphics[width=\textwidth,height=0.62\textheight,keepaspectratio]{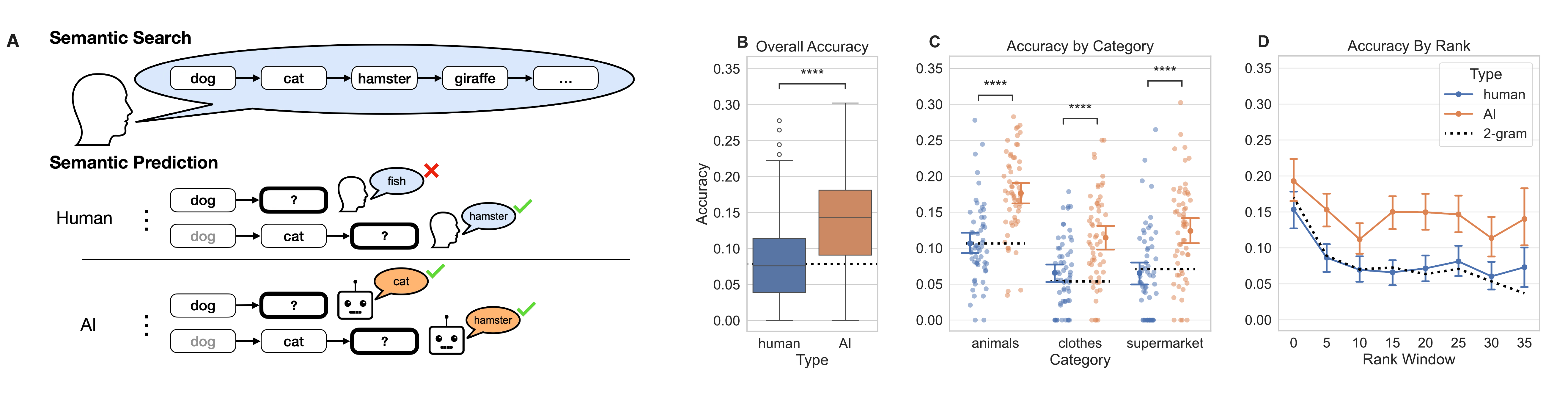}

\caption{
    \textbf{LLM (\texttt{Gemini-3-Pro}) demonstrates superior cognitive micro-alignment by outperforming humans in predicting idiosyncratic semantic search trajectories.}
    The figure compares the zero-shot predictive performance of a LLM against human participants on semantic foraging sequences generated by other humans.
    \textbf{(A-D) Next-Word Prediction.} This task measures the ability to predict the very next word in a sequence.
    \textbf{(A)} A schematic of the task.
    \textbf{(B)} Overall, the LLM achieved substantially higher next-word prediction accuracy than humans (participant-level means: AI $M=0.138$, 95\% CI [0.128, 0.149] vs. human $M=0.079$, 95\% CI [0.071, 0.088]; paired Wilcoxon signed-rank test $W=936.5$, $p < 0.0001$), corresponding to an absolute advantage of $\Delta=+5.91$ percentage points or equivalently a relative performance increase of $75\%$; the 2-gram LOO baseline achieved mean accuracy $M=0.0799$ comparable with humans (dashed line).
    \textbf{(C)} This advantage was consistent across all three semantic categories: animals (AI $M=0.176$, 95\% CI [0.161, 0.192] vs. human $M=0.107$, 95\% CI [0.092, 0.121]; $p < 0.0001$; 2-gram $M=0.111$), clothes (AI $M=0.115$, 95\% CI [0.098, 0.131] vs. human $M=0.066$, 95\% CI [0.053, 0.078]; $p < 0.0001$; 2-gram $M=0.054$), and supermarket items (AI $M=0.124$, 95\% CI [0.106, 0.142] vs. human $M=0.065$, 95\% CI [0.049, 0.081]; $p < 0.0001$; 2-gram $M=0.074$).
    \textbf{(D)} The performance gap was maintained consistently throughout the entire sequence, despite the expected initial decline in overall accuracy associated with the exhaustion of prototypical category members. ****~$p < 0.0001$.
    }
    \label{fig:3}
\end{figure}

While macro-alignment analyses suggest that LLMs capture the canonical structure of human generative semantic memory search, effective collaboration requires navigating the idiosyncrasies of an individual human's search.
A key mechanism of collaborative inhibition is retrieval disruption, whereby a partner’s words interrupt an individual's optimal search path \cite{mannering_modeling_2025}.
We hypothesized that a helpful AI partner must avoid this via cognitive micro-alignment: the ability to localize the specific, unfolding semantic trajectory of an individual within its vast latent space in order to accurately predict their next cognitive step.

To measure this, we evaluated the zero-shot ability of LLMs (\texttt{Gemini-3-Pro}) to predict the next word across 174 participant--sequence assignments from solo SFT data (Experiments 1 and 3; 58 per category, SI Methods). We separately evaluated subcategory-switch prediction across 181 sessions (62 animals, 59 clothes, and 60 supermarket items; SI Methods).
Rather than treat this as naive SFT continuation, we used Theory-Driven Cognitive Prompting (TDCP) \cite{kramer_unlocking_2024}, where
we instructed the model to simulate spreading activation, monitor for semantic saturation, and follow the Marginal Value Theorem in deciding whether to stay within a cluster or switch \cite{kramer_unlocking_2024} (Methods).

By explicitly instructing the model to simulate spreading activation and detect semantic saturation (Methods), we relied on grounding the model in the task's theoretical cognitive basis to improve its alignment with individual human trajectories.
We directly compared these model predictions against human prediction performance (Figure~\ref{fig:3}A) on the same sequences (collected in Experiment 2) across three categories (animals, clothes, supermarket items), using embedding-based metrics for switch determination (SI Methods).
Human predictors ($N=174$; 58 per category) were strongly incentivized with monetary bonuses for every correctly predicted word.

The results indicate that LLMs consistently outperform humans in anticipating the next step of a person's semantic trajectory.
In a direct comparison, the model achieved an overall accuracy advantage of $\Delta = +5.91$ percentage points over human predictors ($M=0.138$, 95\% CI $[0.128, 0.149]$ vs. $M=0.079$, 95\% CI $[0.071, 0.088]$), a difference confirmed by a two-sided paired Wilcoxon signed-rank test ($W = 936.5$, $p < 0.0001$, $n=174$) (Figure~\ref{fig:3}B).
This advantage was robust and general:
the model significantly outperformed humans across all domains—animals ($\Delta = +6.97$pp), clothes ($\Delta = +4.89$pp), and supermarket items ($\Delta = +5.87$pp;
all $p < 0.0001$) (Figure~\ref{fig:3}C)—and substantially exceeded the 2-gram baseline model ($M=0.080$) (SI Methods).

This predictive advantage persists in the later, less predictable stages of search where highly accessible prototypes are exhausted (Figure \ref{fig:3}D).
We further confirmed that this capability is driven by both model quality and theoretical design:
Accuracy scaled with model size ({SI Figure~S4}), and a control analysis directly validated the prompting strategy itself:
TDCP matched the predictive accuracy of few-shot in-context learning seeded with real human sequences, while significantly outperforming a standard zero-shot baseline ({SI Figure~S5, Table~S2}).
This suggests that theory-grounded instructions can substitute example-based in-context learning, with potentially broad applicability for using LLMs as models of human cognition  \cite{kramer_unlocking_2024, ebouky_eliciting_nodate}.

Note that we further evaluated the model's capacity to predict transitions between semantic subcategories, i.e., ``switches'', finding that while both humans and AI performed poorly due to the inherent ambiguity of these stochastic latent events, the LLM maintained a significant edge in sensitivity over human predictors ({SI~Figure~S1, Table S1, Switch Prediction Analysis}).

\begin{figure}[!tbp]
\centering

\includegraphics[width=\columnwidth]{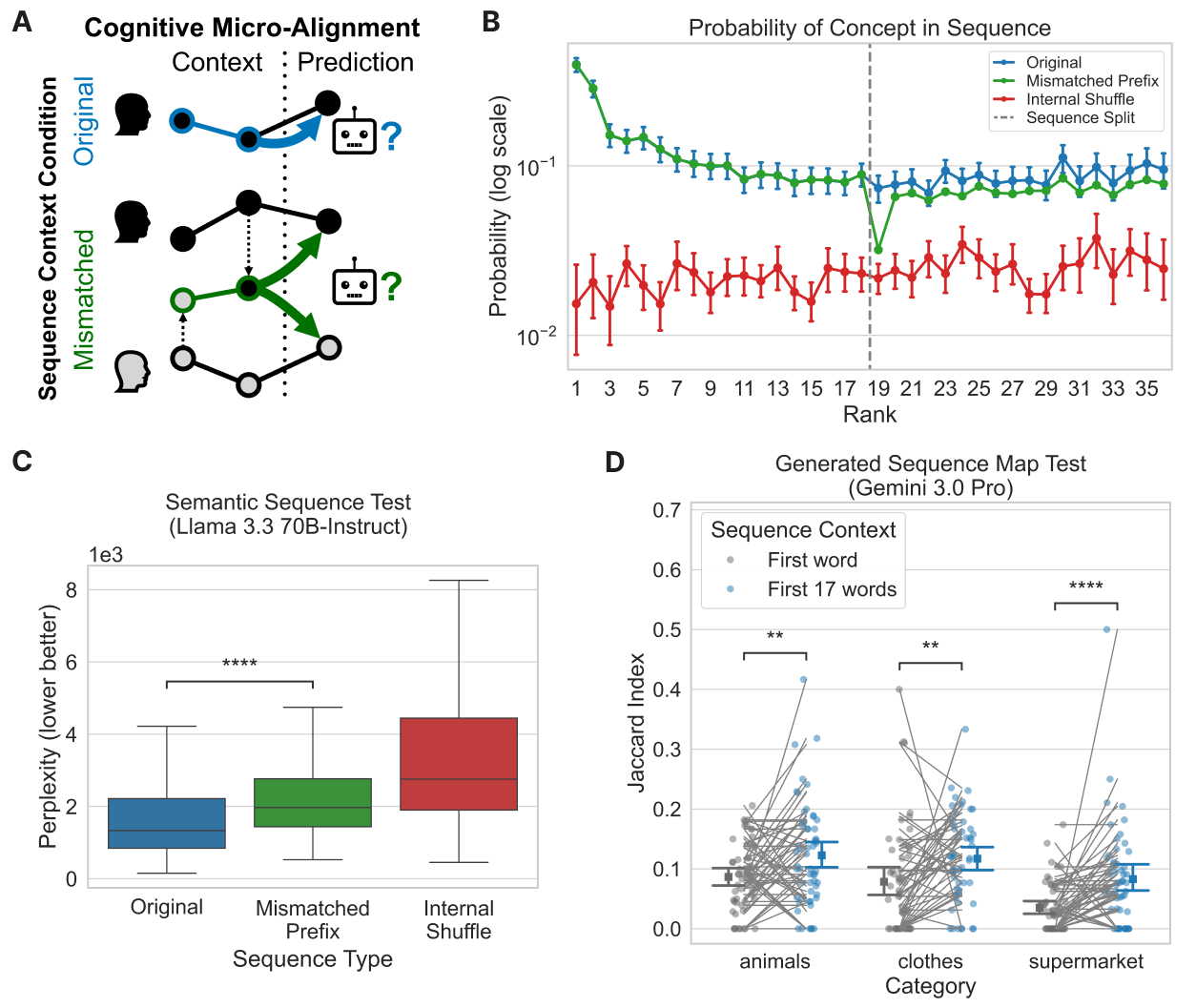}
\captionsetup{font=footnotesize}
\caption{
    \textbf{An LLM demonstrates cognitive micro-alignment by showing adaptive sensitivity to individual human semantic trajectories.}
    This figure presents three analyses quantifying the model's ability to align with the unique, unfolding semantic trajectories of individual humans.
    \textbf{(A) Conceptual model of cognitive micro-alignment.} The schematic illustrates the process by which the model conditions its broad, statistical representation of collective human thought (a superposition of trajectories) onto a single, observed human sequence to infer that individual's unique cognitive path.
    \textbf{(B-C) Sensitivity to idiosyncratic sequence structure.} We tested the model's probability of the next word in original human sequences versus two controls: one where the initial context was replaced by that of another person (``Mismatched Prefix'') and another where the entire sequence order was randomized (``Internal Shuffle'').
    \textbf{(B)} The model consistently assigned a higher probability to the true next word in the original structurally intact sequences compared to the permuted controls.
    \textbf{(C)} Consequently, the model's perplexity (a measure of uncertainty) was significantly lower for original sequences than for those with disrupted context ($p < 0.0001$), demonstrating its sensitivity to the individual's original trajectory.
    \textbf{(D) Adapting generations to an individual's trajectory.} We tested if providing more of a person's unique context helps the model better predict their future responses. We prompted the model with either a single seed word (``First word'' condition) or a longer subsequence (``First 17 words'' condition) from a human's output and measured the Jaccard Index similarity between the model's generated second-half words and the human's subsequent (but unseen) second-half words. Providing the longer subsequence as context significantly increased the similarity for all three categories: animals ($M: 0.087 \to 0.123$, $W=257.5$, $p=0.006$), clothes ($0.078 \to 0.117$, $W=199$, $p=0.002$), and supermarket items ($0.035 \to 0.083$, $W=116$, $p=0.00003$). This indicates the model can leverage an individual's unique trajectory to better align with their semantic map.
    All statistical comparisons shown are from paired Wilcoxon signed-rank tests. Error bars represent 95\%CI. Statistical significance is denoted as: **~$p < 0.01$, ****~$p < 0.0001$.
    }
\label{fig:4}
\end{figure}
We conceptualize cognitive micro-alignment as the process by which an LLM progressively conditions its autoregressive generation, initially reflecting a superposition of individual human semantic trajectories at the population level, onto the increasingly specific unfolding trajectory produced by a single human \cite{shanahan_role_2023}.
This allows the model to collapse a vast space of possible semantic trajectories onto a distribution sharply focused on the likely next steps for that particular person's trajectory, thereby aligning more closely to their unique semantic sequence (Figure \ref{fig:4}A).

To empirically test for micro-alignment, we examined an LLM's sensitivity to the context of an individual human's semantic trajectory.
Specifically, we evaluated a Llama 3.3 70B Instruct model on 699 human ``animals'' semantic foraging trajectories \cite{hills_optimal_2012, zemla_evidence_2023}, with paired Figure~\ref{fig:4}B--C analyses restricted to the 648 sequences complete across all control conditions.

We evaluated the probability that the model assigned to the true next word in each participant's original sequence and compared this against two control conditions chosen to ablate specific structural information from the individual's original sequence (Figure~\ref{fig:4}B).
The first, a ``Mismatched Prefix'' control, tested the model's reliance on a sequence's own idiosyncratic context by replacing the initial prefix of each of 100 participants' sequences (word ranks 1-17) with the corresponding prefix of a sequence from a different, randomly selected participant (excluding the original from the sampling pool).
The second, an ``Internal Shuffle'' control, randomized the entire temporal order of a participant's sequence, testing how much of the model's performance relies on sequential ordering.

The LLM consistently assigned higher probability to the true next word in the original unaltered sequences than in either control, with a modest gap relative to the Mismatched-Prefix condition and a substantially larger gap relative to Internal Shuffle (Figure~\ref{fig:4}B).
The drop in probability at rank $= 18$ reflects the discontinuity when the random sequence first-half prefix is joined with the original.
These analyses allowed us to measure the model's perplexity (over sequence words) for the original human sequences versus the disrupted controls (Figure \ref{fig:4}C).
Perplexity quantifies a model's uncertainty for a given semantic sequence, serving as a measure of the sequence's typicality relative to the model's learned distribution in its training (Methods).
A low perplexity score indicates that a sequence is more predictable, aligning with common semantic patterns the LLM appears to have learned during training.
Conversely, a high perplexity score demonstrates a surprising or atypical sequence.
A paired Wilcoxon signed-rank test indicated a statistically significant increase in mean perplexity from the original sequences ($M = 1924.33$) to their ``Mismatched Prefix'' condition ($M = 2299.61$, $W = 4.85 \times 10^4$, $p < 0.0001$).
For context, perplexity for the ``Internal Shuffle'' condition was substantially higher ($M = 3637.33$) than other conditions.
This evaluation shows that the LLM is sensitive to both the sequential structure of the sequences and their original (prefix) trajectories as measured by perplexity.

Lastly, as a means to evaluate an LLM's ability to adapt to an individual's unique semantic cognitive map \cite{tolman_cognitive_1948}, we measured its capacity more broadly to predict the whole latter half of a human-generated word sequence, in terms of the set of words generated but not their sequential ordering.
That is, the objective was to measure what fraction of predicted words were within the cognitive map explored by the individual during the second half of their sequence.
We prompted the model under two conditions:
a baseline using only the first word (``First word'') and a test condition using a larger subsequence (``First 17 words'') of the first 17 words from the human's output.
In this analysis, cognitive micro-alignment was then quantified using the Jaccard Index (JI) \cite{jaccardEtudeComparativeDistribution1901} which measures the overlap between the LLM-generated words and the unseen portion of the human's sequence.
The JI between two sets is defined as the number of shared elements divided by the number of elements in their union.
Here, the two sets were the LLM's generated words and the human's unseen second-half words. JI ranges from $0$ (no overlap) to $1$ (identical sets), and higher values indicate better prediction irrespective of order.

The results of paired two-sided Wilcoxon signed-rank tests for all three domain categories showed significant increases in the model's ability to predict words in the second half of each sequence when its context went from the single first word (``First word'' condition) to the first part of the sequence (``First 17 words'' condition) (Figure \ref{fig:4}D).

These findings suggest the LLM can leverage an individual's semantic trajectory to better predict their subsequent word generation beyond just the next word.

\subsection*{Dyadic Experiments}

LLMs exhibit a superior capacity for cognitive micro-alignment, outperforming humans in tracking idiosyncratic human semantic search trajectories. To assess how this capacity influences human behavior as humans interact with AIs in generative cognitive tasks, we conducted real-time human--AI collaborative semantic search experiments.
We hypothesized that LLM cognitive micro-alignment would promote synergistic interactions in human--AI dyads compared to human--human dyads. An important test of such synergy is the ability to overcome the well-documented phenomenon of collaborative inhibition \cite{szary_patterns_2015, mannering_application_2023, hinds_collaborative_2016}.
A series of dyadic SFT experiments were run in different conditions, directly comparing behavior and performance of human--human pairs with human--AI collaborations across different LLM interaction behavior prompts, careful to account for confounding factors in LLM and human comparisons, such as response timing.
Our central hypothesis is that an AI partner, by virtue of its superior ability to cognitively align with a human's search process, can mitigate the retrieval disruption that underlies collaborative inhibition, and so enhance human memory search performance, whether in terms of the number of concepts generated, response times, or semantic distance traversed.

Participants were asked to collaborate with a partner in two rounds to produce as many concepts from a domain category (``animals'' and ``clothes'') as possible within three minutes, each providing a single concept word at a time in interleaved turns alternating back and forth (Figure \ref{fig:1}D, Experiment 4 SI Methods).
For each round, participants were paired with either a human or an LLM without knowing which dyad type, human--human or human--AI, they were in.

To investigate the effect of different LLM-behavioral modes of interaction, LLM partners (\texttt{Llama-3.3 70B-Instruct}) were prompted with one of three types of instructions, defining three task conditions: divergent (instructed to always provide a word from the domain category but outside of the current subcategory), convergent (prompted to provide words that are always semantically similar to the previous, human-generated, one), and inferred (prompted only to be helpful in the task without specific directions on types of words to provide). Simulations with LLM--LLM dyads validated the distinct effect of the prompts on semantic trajectories, showing a larger semantic distance traversed for dyads with one divergent-instructed LLM ({SI~Figure~S8}).
We collected SFT sequences from an initial pool of 207 participants in the
collaborative SFT, resulting in 96 valid sequences from 192 participants in human--human dyads and 144 valid human--AI sequences after filtering. 129 participants completed valid sequences in both human--human and human--AI dyads.

To assess whether collaborative inhibition could be identified in our paradigm, we additionally collected solo SFT data for the categories ``animals'' (from 61 participants) and ``clothes'' (from 64 participants).
The SFT score is the number of unique concepts produced by the solo participant or the dyad.
To measure collaborative inhibition in terms of concept production, we formed all possible nominal pairs of solo participant sequences in each SFT category and counted the union of the concepts produced by the pair, and then compared the nominal pair production with that of the actual dyads  \cite{mannering_application_2023, szary_patterns_2015}.
Nominal pairs ($n = 3{,}846$, median $= 59.0$ [IQR: $51.0$, $66.0$]) produced a significantly larger number of unique concepts than human--human dyads ($n = 96$ dyads, median $= 33.0$ [IQR: $28.8$, $39.0$],  Mann--Whitney $U = 353{,}251.5$, $p < 0.001$), as well as human--AI dyads with convergent ($n = 48$, median $= 34.0$ [IQR: $30.0$, $38.2$],  Mann--Whitney $U = 180{,}267$, $p < 0.001$),  divergent ($n = 46$, median $=31.5$ [IQR: $27.2$, $36.8$],  Mann--Whitney $U = 173{,}486.5$, $p < 0.001$) and inferred ($n = 50$, median $= 34.0$ [IQR: $29.0$, $40.0$],  Mann--Whitney $U = 185{,}744$, $p < 0.001$) prompts.
These results confirm collaborative inhibition in the interleaved SFT, with the concept count performance of collaborating dyads falling behind the combined performance of two independent participants.
Importantly, there were no significant overall differences in concept count between human--human and human--AI dyads across both categories and all prompt types ({SI~Figures~S9 and S12}).
Additionally, within-participant comparisons showed no differences in concept production or in response times between both dyadic conditions in all prompts ({SI~Figure~S12}).

Although no differences were found in the overall concept production or response time distributions between human--human and human--AI dyads, we assessed whether different modalities of collaboration had more subtle effects on individual behavior.
The key behavioral indicator that humans are finding it more difficult to produce a new concept is increasing response times between words (RTs). Indeed, some evidence in favor of foraging models of semantic search is based on RT increases as a concept ``patch'' is depleted. \cite{hills_optimal_2012}.
Thus, we sought to examine if AIs can help humans to find more concepts overall by reducing their RT change from early to late phases of the search process, enabling more consistent exploitation of patches throughout the task.
We hypothesized that different collaboration conditions could modulate individual behavior, such as response times, differently during periods closer to the start or to the end of the task.
Some AI-generated cues could drive partners to exploit a patch for longer, or to instead move ballistically through semantic memory. These changes in search strategy could be differentially more advantageous at the beginning of the task, when clusters are still rich and unexplored, or toward the end of the task, when clusters become more sparse.
Although not leading to an increased concept production, different interaction modalities could thus nonetheless lead to more efficient search within clusters. Additionally, partners might display different levels of influence on the other partner's thoughts or their overall semantic trajectory.
These possibilities motivated our further analyses of individual partner behavior in the collaborative SFT.

\begin{figure}[tbp]
\centering
\includegraphics[width=\linewidth,height=0.50\textheight,keepaspectratio]{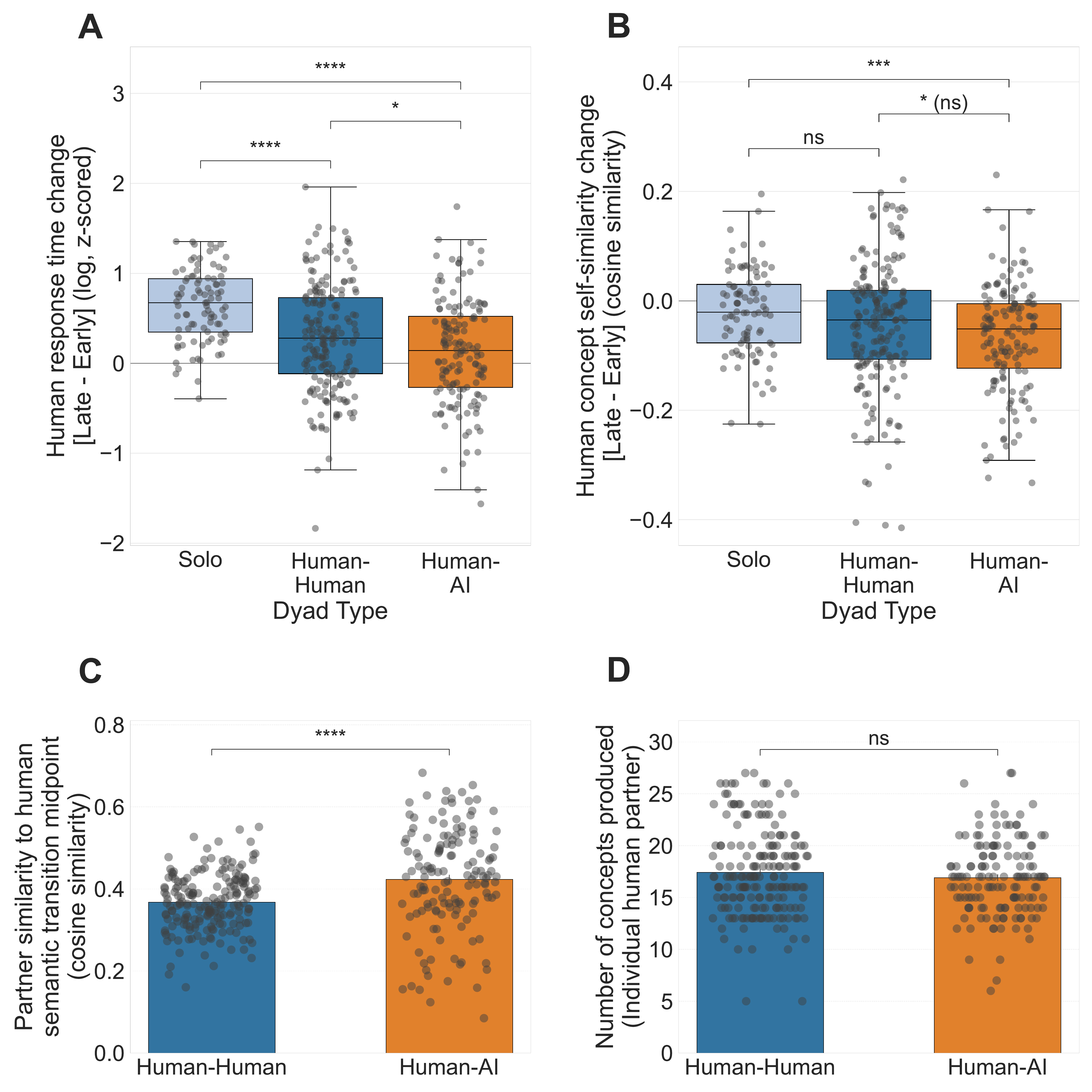}

\caption{
    \textbf{Dyadic alternating interactions with LLM partners stabilize human inter-word response times and shift human semantic trajectories away from self-produced concepts, without increasing overall productivity.}
   \textbf{(A)}~\textbf{Response times.}
   The behavioral metric ($y$-axis) is the average difference in $z$-scored log-transformed response time between the late and early phases of an SFT across participants.
   All three conditions exhibited significantly longer RTs in the late relative to early half of the task, but this effect was attenuated in human--AI dyads compared with both solo individual participants (Mann--Whitney $U = 1.01 \times 10^{4}$, $p<0.0001$) and human--human dyads ($U = 1.58 \times 10^{4}$, $p = 0.026$). Human--human dyads also showed reduced RT differences relative to individual participants ($U = 1.22 \times 10^{4}$, $p<0.0001$).
   Metrics for each prompt type are detailed separately in {SI~Figure~S16}.
  \textbf{(B)}~\textbf{Geometric self-similarity.} The $y$-axis shows the mean difference in word embedding self-similarity between late and early SFT. Human--AI dyads displayed greater decreases in self-similarity in the later half of the task than participants in the solo SFT (human--AI vs.\ solo: $U = 8.44 \times 10^{3}$, $p = 0.001$), while the difference between human--human dyads and solo performance was not significant ($U = 9.86 \times 10^{3}$, $p = 0.114$). The difference between human--AI and human--human dyads did not survive correction for multiple comparisons ($U = 1.57 \times 10^{4}$, $p = 0.073$). Box plots indicate median, interquartile range (IQR), and 1.5$\times$ IQR (panels A and B).
    \textbf{(C)}~\textbf{Semantic transition geometry.} The midpoint in the embedding space between each pair of successive words produced by a human was calculated and compared to the embedding of the intervening word contributed by the partner. Words contributed by AI partners had significantly higher similarity on average ($U=8.94\times 10^{3}$, $p = 2.93\times 10^{-8}$), and also higher similarity variance (Brown-Forsythe variance test, $W = 44.72$, $p=9.6\times 10^{-11}$, HA SD=$0.124$ vs HH SD=$0.068$), with respect to the human participant's semantic midpoint.
   \textbf{(D)}~\textbf{Collaborative productivity.} The number of concepts produced did not differ between the human-AI and human-human dyads ($U=1.45\times 10^{4}$, $p = 0.44$).
     Data for ``clothes'' and ``animals'' was included. Changes against zero used two-sided one-sample Wilcoxon signed-rank tests; between-condition comparisons used two-sided Mann--Whitney $U$ tests with Holm--Bonferroni correction where applicable. Error bars indicate standard error of the mean (panels C and D). Star annotations reflect uncorrected $p$-values. *~$p < 0.05$; ***~$p < 0.001$; ****~$p < 0.0001$.
  }
\label{fig:5}
\end{figure}

We assessed changes across the task in response times and semantic trajectories (Figure \ref{fig:5}).
A median split by total word count was created, dividing the task into two phases, an early phase and late phase.
For semantic trajectory analysis, word embedding vectors (SI Methods) and pairwise cosine similarity were computed between consecutive words produced by both participants in a dyad.

We found that humans collaborating with an LLM prompted with convergent or divergent instructions did not significantly increase their response times in the late phase of the task ({SI~Figures~S10, S16}), in contrast with individuals and participants collaborating with other humans.
As shown in Figure \ref{fig:5}A, when collaborating with other humans, participants show a significantly larger increase in inter-word response times in the latter half of the task compared to when collaborating with an LLM (see {SI~Figure~S14} for switch response times).
Fixed effects linear modeling showed that this effect was modulated by embedding similarity to the previous word, as we detail in {SI Table S3}:  in convergent human--AI dyads specifically, higher cosine similarity to the previous word was associated with shorter response times in the late phase of the task.
We computed self-similarity as the embedding cosine similarity relative to the previous word produced by the same participant. This revealed that humans interacting with other humans produce words that are less self-similar in the second half of their sequences than in the first half. When interacting with LLMs, the decrease in self-similarity is even stronger (Figure \ref{fig:5}B, although not statistically significant after correction for multiple comparisons; see also {SI~Figure~S11} for overall patterns).
This suggests that participants were more influenced over time by AI partners than by human partners.
To test whether this was mediated by humans increasingly being influenced by the LLM's semantic trajectory, we compared the change in cosine similarity between each human's word and their partner's previous word from the early to late phases of the SFT in human--human and human--AI dyads.

Similarity between each human response and the partner's immediately preceding response decreased from early to late phases in both human--human and human--AI dyads ({SI~Figure~S15}). Thus, the effect of LLM partners on human semantic trajectories cannot be explained simply by humans becoming increasingly aligned with the LLMs' preceding word choices. To characterize where partner contributions fell relative to the human's own trajectory, we computed a semantic midpoint for each human--partner--human triplet: the point halfway in embedding space between a human's word before the partner's turn and that same human's next word after the partner's turn. LLM-produced words were, on average, closer to this midpoint than human-partner words, but their distances were more variable (Fig.~\ref{fig:5}C).
Together, our dyadic experiments show that, although pairing with AIs does not increase overall concept production (Fig.~\ref{fig:5}D), humans collaborate differently when blindly paired with AIs than with other humans. Collaboration with AIs has a stabilizing effect on the humans' response times over the course of the interaction while also shifting their trajectories away from their prior semantic path (but not just toward the AI's path).

\section*{Discussion}

Our study introduces and empirically examines two key ideas in human--AI interaction: cognitive alignment and cognitive synergy.
We found that LLMs can approximate the macro-level statistical structure of human semantic search. Thus, not only are latent word representations in LLMs organized in a manner that is structurally similar to human conceptual spaces \cite{xu2025revealing}, but the dynamics by which these latent conceptual spaces are traversed resemble patterns of human memory retrieval.
More notably, AI models demonstrate a capacity to adapt to the specific semantic trajectories of individual humans with a higher degree of accuracy than other humans achieve---a capability we measure as cognitive micro-alignment.
This fine-grained, highly predictive ability suggests a productive path forward for leveraging AI in real-time collaboration to foster cognitive synergy.

Language models are better than humans at next-token prediction \cite{shlegeris_language_2024}.
Beyond this, our study shows that language models are better than humans at next-concept prediction in individual human memory search.
Potentially, this ability emerges because the LLM’s latent space acts as a statistical amalgamation of an enormous number of human semantic pathways derived from its training data that can be quickly recruited via contextual input \cite{shanahan_role_2023, marks_persona_2026, beckmann2026mind}.
While any single human navigates their own idiosyncratic semantic network, the LLM has learned a high-dimensional representation of a collective human semantic network.
A specific human's semantic map may be addressed from within this collective semantic network via a short sub-sequence of concepts.
We speculate that this induces an implicit representation of the individual's current cognitive state, allowing the model to then infer how the person's semantic retrieval path is likely to proceed---e.g., via stable taxonomic structures (a canary is a bird) or perhaps more fluid, context-dependent thematic associations (a canary is related to a coal mine) \cite{liu_cross-domain_2022}.
Looking forward, such predictions may be studied using mechanistic interpretability techniques, which aim to reverse-engineer the internal circuits of neural networks into human-understandable algorithms \citep{lieberum_gemma_2024, zarries_components_2025, lacosse_emerging_2025}.
This offers a direct parallel to neuroscience research that grounds semantic search in the neural dynamics of hippocampal-entorhinal activity \citep{nour_trajectories_2023, lundin_neural_2023}.

We believe these LLM abilities have direct implications for overcoming collaborative inhibition, which can stem from retrieval disruption when a partner’s cues interrupt an individual's own search path \cite{mannering_application_2023}.
Our results from the dyadic experiments suggest that a human--AI collaboration can enhance human search under certain conditions depending on the LLM behavioral prompt and the interaction mechanism.
Although overall concept count did not increase in human--AI dyads, humans paired with LLMs exhibited more stable SFT response times, suggesting the AI partner helped the human more thoroughly exploit productive semantic clusters.
Considering the difficulty of accurately predicting switches to new subcategories (for both humans and AIs, {SI~Figure~S1}), one form of help a collaborative partner could provide is to stay along the semantic trajectory of the human (e.g., via a convergent prompt), where prediction may be easier.
We found evidence that an AI partner can modify a human's semantic trajectory and facilitate the retrieval of new concepts without inducing the human to merely appropriate the AI's own trajectory.
Further analysis should explore this possibility and whether similar or different interaction effects arise with other model classes, e.g., large reasoning models (provided they have low enough latency for our experimental setup requiring naturalistic real-time interaction) \cite{besta_reasoning_2025}.

Future systems need not rely on static prompting strategies.
Our Theory-Driven Cognitive Prompting (TDCP) approach falls within the Cognitive Prompting framework for LLMs \cite{kramer_unlocking_2024}, which posits that explicitly instructing models to execute structured cognitive operations---such as goal decomposition or pattern recognition---can enhance reasoning capabilities.
To validate the robustness of this approach, we performed control analyses comparing our theory-driven prompts against standard in-context learning (ICL) baselines ({SI~Figure~S5, Table S2}).
We found that TDCP significantly outperforms standard zero-shot baselines and achieves predictive fidelity comparable to few-shot baselines provided with actual human example sequences for most capable models.
The performance of TDCP techniques observed in our semantic search paradigm indicates that explicit instructions derived from cognitive theories may effectively substitute for LLM prompts based on in-context learning (ICL), thus offering an alternative mechanism for aligning AI with human cognitive processes in other task contexts.
Our results show that frontier models are increasingly capable of tracking human thought;
we therefore suggest that combining these models with external cognitive modules \cite{ebouky_eliciting_nodate} and mixtures of ICL and TDCP strategies could yield greater cognitive alignment and thus human--AI cognitive synergy.

Additionally, while current architectures may prioritize statistical efficiency over conceptual alignment \cite{alkhamissi_language_2025}, techniques such as fine-tuning on representational alignment tasks have been shown to restructure latent spaces and can shift model representations closer to those resembling humans \cite{mahner_dimensions_2025,muttenthaler_aligning_2025}.
However, it remains an open question whether this representational alignment translates into the dynamic cognitive alignment necessary for effective real-time human--AI collaboration.

LLMs and generative AI systems are underpinned by data-driven statistical power and scalable neural architectures, yielding a form of intelligence that differs from that of humans \cite{shani_tokens_2025}. The distinctions between AI and human cognition present opportunities to combine diverse intelligences in productive ways.
These AIs' capacity to align with (i.e., emulate and predict) human thought trajectories may serve as a kind of computational Theory of Mind \cite{leslie_core_2004}, establishing the common ground necessary for the emergence of new collective intelligences \cite{woolley_evidence_2010, raineyCouldHumansAI2025}. Indeed, efforts are currently underway in industry to engineer AI systems whose primary objective is not merely autonomous performance, but effective interaction and collaboration with humans \cite{humansand2026,thinkingmachines2026interaction}.
We suggest that by leveraging this (dis)alignment in cognition between humans and AIs, we can engineer collaborations that do not merely mimic human partners, but actively harness distinct AI advantages to modulate and expand human cognitive strategies \cite{linHumanAIComplementarity2026, bratton_cognitive_2025}.

\subsubsection*{Limitations and Further Work}
While our results show promising foundations for developing cognitive synergy, there are many aspects of our work that should be improved upon.
Here we have only explored a rather inflexible, interleaved interaction mechanism in our dyadic experiments.
This forces constant turn-taking, which is likely sub-optimal for leveraging unique AI capabilities in collaboration.
We are developing mechanisms where the AI partner can adaptively intervene strategically with the human partner, thus creating a different interaction dynamic.
Various model behaviors and prompt engineering strategies, including the convergent, divergent, and inferred prompts studied here, may also have complex interplay with particular interaction mechanisms in shaping cognitive synergy \cite{szary_patterns_2015}.
For instance, an AI that can monitor how long a person is taking to say their next word could be designed to intervene only when it predicts the human is slowing down (increasing RT) or is about to make a suboptimal word choice, on the basis of RT information ({SI~Figure~S6}).
Additionally, strict latency requirements for real-time interaction necessitated the use of the Llama-3.3-70B model \cite{grattafiori_llama_2024}, which lacks the advanced model capabilities that drove our strongest predictive results ({SI~Figure~S7}).
Future implementations leveraging these more capable (and efficient) frontier models may yield stronger synergistic effects than those observed here.

Another constraint of the interleaved design is that it forces serial processing.
An alternative is to present AI assistance in parallel rather than in strict turn-taking, allowing the human to integrate it without disrupting their primary production flow.
This is closer to the setup in human collaborative SFT studies, where the main cause of collaborative inhibition is the disruption of one's internal retrieval process by the words being said at uncontrolled times by one's partner \cite{mannering_modeling_2025}.
By moving AI cues out of the direct conversational channel, and instead, for instance, presenting a small dynamic list of suggested words on the side of a screen, this disruption could be reduced.

Instead of the AI ``pushing'' cues to the user, future work should also investigate a ``pull'' mechanism where the user actively requests assistance.
This approach grants the human full agency over their cognitive search process, ensuring the AI does not interrupt a productive search, but it shifts the AI's role from a collaborator to more of an on-demand cognitive tool.
Key questions are how user-initiated collaborations compare to self-initiated ones in terms of performance and behavior, and whether users develop unique strategies for when and how to use AI partners of different types.

Furthermore, our results illuminate a critical distinction between prediction and control in collaborative interactions.
While the LLM demonstrated superior cognitive micro-alignment (more accurately predicting the human’s next step), this did not translate into a significant increase in total concept yield.
This implies that basing responses on the ``most probable'' next concept, while maximizing alignment, may not lead to the optimal intervention to stimulate a human's ongoing semantic retrieval.
If an AI partner merely anticipates the word a human is already about to produce, it fails to provide the novel semantic cues necessary to overcome inhibition or open new search paths.
True cognitive synergy likely requires treating collaboration partly as a control problem rather than a pure prediction problem, requiring agents capable of planning interventions that actively guide the human toward underexplored semantic patches rather than simply mirroring their current trajectory.
Indeed, an LLM could be augmented with an external module which, when queried, applies planning algorithms such as Monte-Carlo tree search or model-predictive control to a mental-world model of a human collaborative partner in order to produce an optimal sequence of semantic generations \cite{mcnamee_internal_2019}.
Optimality in this context could be flexibly defined.
For example, different reward functions could correspond to maximizing open-ended conceptual exploration or maximizing the probability of the human reasoning their way to a solution of a problem.

We also note that while LLMs can excel at predicting the outcome of a cognitive process, e.g., predicting the next concept during semantic search, this does not provide strong evidence that it is reproducing the process itself, doing generative memory search with specific human memory constraints.
A more parsimonious perspective might be that its pattern matching is completely divorced of any resemblance to a human cognitive process.
More work in the field of cognitive mechanistic interpretability is needed to understand how memory search mechanisms in AI appear to emerge \cite{zarries_components_2025, hu_signatures_2025}.

\subsubsection*{Conclusion}
By leveraging the semantic fluency task, an experimental paradigm chosen for its power as a quantifiable and theoretically-grounded probe into cognitive search \cite{todd_cognitive_2012}, we provide an empirical and quantitative approach for examining human--AI cognitive alignment.
We do so by comparing measures of dyadic collaboration in which humans do not know whether they are collaborating with an AI or another human.
The well-documented usefulness of the task for characterizing human semantic search and producing collaborative inhibition between humans provides an ideal testbed to demonstrate the potential of an AI partner to facilitate semantic memory retrieval and overcome known collaborative failures \cite{szary_patterns_2015, mannering_application_2023, hinds_collaborative_2016}.
These results can inform development of human--AI collaboration that is not merely task-based, but
instead engages both agents in a process that promotes adaptability, shared goals, and an evolving cognitive alignment between human and AI perspectives, leading to collaborative performance that exceeds what either humans or AI could achieve alone \cite{vaccaro_when_2024}.
Our findings help shift focus away from autonomous AI performance and towards the deliberate design of cognitively-aware thought partners \cite{collins_building_2024}.
The future of enhancing human intelligence lies not only in building more powerful models, but also in engineering dynamic adaptive interaction mechanisms that enable human and artificial intelligences to collaborate together, integrating the distinct strengths of both cognitive systems and expanding our collective ability to explore and solve new challenges \cite{clark_extending_2025,bratton_cognitive_2025}.

\section*{Materials and Methods}
\subsection*{Human--AI Alignment Analyses}

\subsubsection*{Language Models}
For macro-alignment/micro-alignment generation, next-word, and switch prediction analyses, we use the Google Gemini 2.5 and 3 models (Lite, Flash, and Pro GA, access date: February 1st, 2026), a suite of highly performant natively multimodal reasoning models at the frontier of LLM capabilities (state-of-the-art at time of writing) \cite{comanici_gemini_2025, noauthor_new_2025}.
These models are sparse mixture-of-experts (MoE) that activate only a subset of neurons upon inference \cite{jiang_mixtral_2024}.
For all experiments, we rely on system default configuration values for each model.
Only Flash and Pro models used ``dynamic thinking'' or dynamic intermediate token generation, where the model makes a self-determination how many intermediate ``thinking'' tokens to generate based on prompt instruction complexity.

We utilize 4-bit quantization implemented via the Unsloth library \cite{unsloth_team_unsloth_2023} for the Llama-3.3 70B-Instruct model.
All system prompts used for any tasks in this manuscript are available in {SI}.
Briefly, to achieve effective LLM behavior, we employ a method we term Theory-Driven Cognitive Prompting intended to emulate human memory retrieval.
Drawing on recent frameworks that model reasoning as the orchestrated execution of modular cognitive operations \cite{kramer_unlocking_2024, ebouky_eliciting_nodate}, this approach guides the model through discrete, structured reasoning steps.
Our system prompt explicitly instructed the LLM to simulate the spreading activation of semantic networks.
Before predicting the next word, the model was required to execute a dedicated context analysis operation to detect semantic saturation, defined as the point at which a semantic cluster is depleted.
If saturation was detected, the model was instructed to predict an associative leap (a switch); otherwise, it was instructed to follow cluster cohesion (stay local).
This explicitly maps the Marginal Value Theorem of foraging theory onto the model's generation process.
To validate the efficacy of this strategy, we implemented comparative control conditions detailed in the SI, including a ``Standard Zero-Shot Baseline'' (prediction without cognitive instructions) and a ``Standard Few-Shot Baseline'' (prediction provided with five randomly selected example sequences from the dataset).

\subsubsection*{Evaluation Metrics}

\paragraph{BLEU scoring} We quantified the similarity between AI-generated and human-generated sequences using the BLEU (Bilingual Evaluation Understudy) score \cite{papineni_bleu_2002} following the metric proposed in \cite{heineman_towards_2024}.

\paragraph{Perplexity}
We evaluated the model's performance on the generated sequences using perplexity (PP).
Perplexity is a standard measure of a language model's predictive accuracy.
It is calculated as the exponential of the cross-entropy loss, or the average negative log-likelihood per token.
A lower perplexity score indicates a better model fit and is defined by
\begin{equation} \label{eq:perplexity}
PP(W) := \exp\left( -\frac{1}{N} \sum_{i=1}^{N} \ln P(w_i|w_1, \dots, w_{i-1}) \right)
\end{equation}
where $W$ represents the entire sequence of tokens $(w_1, \dots, w_N)$, and $N$ is the total number of tokens in the sequence.

 \paragraph{Jaccard Index} To quantify the similarity between the set of words generated by the model and the actual set of words produced by a human, we used the Jaccard Index \cite{jaccardEtudeComparativeDistribution1901}.
This statistical measure calculates the similarity between two sets by dividing the size of their intersection by the size of their union.
The resulting score ranges from 0 (indicating no overlap) to 1 (indicating identical sets).  Further details may be found in the Supplementary Materials and Methods ({SI}).

\subsection*{Human--AI Collaborative Experiments}
\ \\
The study protocol was approved by the Conselho de Ética da Fundação Champalimaud (Project CRECOG).
All human participants provided written informed consent prior to participation.
Data was collected in four independent web-based experiments.
The web interface was developed with the Empirica platform, facilitating synchronous task participation for pairs of participants \cite{almaatouq_empirica_2021}.
Native English speakers were recruited from the online crowd-sourcing platform Prolific \cite{palan_prolificacsubject_2018} with a balanced sex quota. Other recruitment criteria included the absence of literacy difficulties, a study submission approval rate between 95-100\% and more than 200 previous submissions. Response times were recorded for produced semantic fluency words and predictions.

In Experiment 1, participants ($n=36$) typed names of animals, clothes, and supermarket items for three minutes. Then, they retrospectively marked their inferred ``switches'' between clusters of words they had produced. Solo SFT sequences from one of the three categories ($n=92$) were also collected from Experiment 3 described below, producing a pooled total of 190 valid sequences. In Experiment 2, participants ($n=174$; 58 per category) predicted human SFT sequences word-by-word by typing the predicted next word with feedback. Experiment 3 combined a solo SFT block with next-switch prediction for a randomly sampled sequence previously produced by another participant, followed by switch marking and cluster labeling of both sequences; the switch-prediction analysis included 181 sessions (62 animals, 59 clothes, and 60 supermarket items). Experiment 4 used collaborative SFT in human--human ($n=192$ individual participants grouped into 96 pairs) or human--LLM ($n=144$) dyads with different LLM prompt conditions (convergent, divergent, and inferred).
To maintain optimal session duration and reduce participant fatigue in this repeated-measures design, we restricted the domain to just two categories representing the animate (animals) vs. inanimate (clothes) distinction, excluding the supermarket category used in earlier experiments.
To mitigate the confounding effect of faster LLM-partner responses in comparison to human-partner response times, an artificial delay was added to the LLM responses by sampling target delays from a distribution fitted on previously collected human--human dyadic data (36 participants). Further details are in {SI~Figure~S13}.

\paragraph{AI Disclosure Statement.} The authors declare the use of generative AI in the research and writing process. According to the GAIDeT taxonomy (2025), the following tasks were delegated to GAI tools under full human supervision: - Code generation - Data cleaning - Visualization Code - Proofreading and editing - Reformatting The GAI tool used was: Gemini 3 Pro. Responsibility for the final manuscript lies entirely with the authors. GAI tools are not listed as authors and do not bear responsibility for the final outcomes.

\section*{Author Contributions}
E.L., M.D., P.T., D.M. designed research; E. L., M.D., G.T. performed research; E. L., M.D. contributed new reagents/analytic tools; E.L., M.D., D.M. analyzed data; E.L., M.D., P.T., D.M. wrote the paper.

\section*{Acknowledgments}
E.L., M.D., D.M. thank the Champalimaud Foundation and the Portuguese Recovery and Resilience Plan (project number 62) for funding. Google Research Credits 423033108.

\clearpage
\setcounter{figure}{0}
\setcounter{table}{0}
\setcounter{equation}{0}
\renewcommand{\thefigure}{S\arabic{figure}}
\renewcommand{\thetable}{S\arabic{table}}
\renewcommand{\theequation}{S\arabic{equation}}
\renewcommand{\theHfigure}{S.\arabic{figure}}
\renewcommand{\theHtable}{S.\arabic{table}}
\renewcommand{\theHequation}{S.\arabic{equation}}

\section*{Supplementary Information}
\begin{center}
\textbf{Supplementary Information for: AI models can predict and collaboratively modulate human memory search}\\[0.5em]
Eric Lacosse$^*$, Mariana Duarte$^*$, Graham Todd, Peter M. Todd, Daniel C. McNamee
\end{center}

\section*{Supplementary Materials and Methods}

\subsection*{Transition Probability Matrix Baseline}
We implemented a Transition Probability Matrix (TPM) baseline generator that produces synthetic semantic-fluency sequences by sampling a Markov chain estimated from human data under leave-one-out (LOO).
We defined the vocabulary as the set of unique observed response tokens and computed a participant-specific TPM by counting observed adjacent transitions \(C_i(w_t,w_{t+1})\) within each participant’s sequence and row-normalizing to obtain \(P_i(w_{t+1}\mid w_t)\).
For each target participant \(j\), we formed the LOO transition model by averaging TPMs over all other valid participants, \(\bar P_{-j} = \frac{1}{N-1}\sum_{i\neq j} P_i\).
To generate a sequence for participant \(j\), we matched the target length to the number of remaining items for that participant and seeded generation with the human exemplar at rank 1 for that same participant; subsequent items were sampled iteratively from \(\bar P_{-j}(\cdot\mid w_t)\).

\subsection*{N-gram Model Baseline}
We implemented an n‑gram next-exemplar prediction model.
Responses were all normalized by lower-casing and removing spaces.
For an n‑gram order \(n\), the context $h$ at each step is the suffix of up to \(n\!-\!1\) prior exemplars;
the model estimates \(P(w\mid h)\) from corpus counts with additive (Laplace) smoothing \(\alpha\): \(P(w\mid h)=(C(h,w)+\alpha)/(C(h)+\alpha|V|)\), where \(V\) is the training vocabulary.
Our results set $\alpha = 1$.
Training was performed using leave-one-out cross-validation at the participant level.

\subsection*{Human--AI Alignment Analyses}

\subsubsection*{Normative switch detection} To determine switches in individual SFT word sequences in categories for which \emph{subcategory norms} exist (e.g. the animals category), we used the extended Troyer norms of \cite{supp:zemla_snafu_2020}. For each word in a sequence, we first identified all of the subcategories to which it belongs. Then, for each pair of successive words in the sequence, we identify a switch if there are no overlapping subcategories between the words. For instance, in the subsequence $\{w_i, w_{i+1}\}= \{\text{dog}, \text{crocodile}\}$ we would identify a switch because the subcategories of ``dog'' (i.e. ``pet'' and ``mammal'') do not overlap at all with the subcategories of ``crocodile'' (i.e. ``reptile'' and ``aquatic'').

\subsubsection*{Embedding switch detection} Prior work has explored the use of word embedding models as a mechanism for detecting conceptual switches \cite{supp:alacam_exploring_2022}, finding that a thresholding strategy using ConceptNet embeddings \cite{supp:speer_conceptnet_2018} outperformed alternative models, including GloVe \cite{supp:pennington_glove_2014}, fastText \cite{supp:joulin_fasttextzip_2016}, and BERT \cite{supp:devlin_bert_2019}. ConceptNet appears to perform best in part because it encodes distributional semantics with explicit taxonomical relations. We adopt this approach, using ConceptNet to embed participant responses. We then compute the \textit{cosine similarity} between each successive pair of word embeddings within each participant's SFT sequence. We set the switch threshold for each sequence to be the median of its embedding similarity values. For each pair of successive words in a sequence, we identify a switch if the similarity between their embeddings is lower than the sequence-specific switch threshold.

\subsubsection*{Statistical Analysis of Transition Probability Matrix (TPM) Similarity}
To quantify the degree of alignment between human and LLM semantic search strategies, we compared their respective Transition Probability Matrices (TPMs) and calculated the Spearman rank correlation coefficient ($\rho$) between the lower-triangle and diagonal of the matrices.
To robustly assess the statistical significance of the Spearman rank correlation coefficient to compare two TPMs, we use a Mantel test with $10,000$ permutations.
The Mantel test evaluates the correlation between two matrices by keeping one matrix constant and randomly permuting the rows and columns of the other simultaneously.
The resulting null distribution allows for calculation of the p-value.

\subsubsection*{Evaluation Metrics}

\paragraph{BLEU score details.} The BLEU score is calculated as the product of two components: a modified n-gram precision and a brevity penalty.
The precision component is the geometric mean of the precision for unigrams, bigrams, and trigrams, measuring the proportion of word sequences from the candidate that are also found in the human reference set.
To validate this n-gram range, we analyzed cluster sizes in the human reference dataset using SNAFU norms.
We found a mean cluster size of 1.88, with 89.5\% of clusters consisting of three items or fewer (Size 1: 54.3\%, Size 2: 24.8\%, Size 3: 10.4\%).
Therefore, an evaluation up to trigrams adequately captures the dependencies within the vast majority of human semantic clusters.
To avoid rewarding the over-generation of common phrases, these n-gram counts are clipped to their maximum frequency in any single reference.
The brevity penalty then discounts the score for candidate sequences that are shorter than the reference length, correcting for artificially inflated precision scores in short outputs.

\subsection*{Human vs. AI Switch Prediction}

\begin{figure}
\centering
\includegraphics[width=\textwidth]{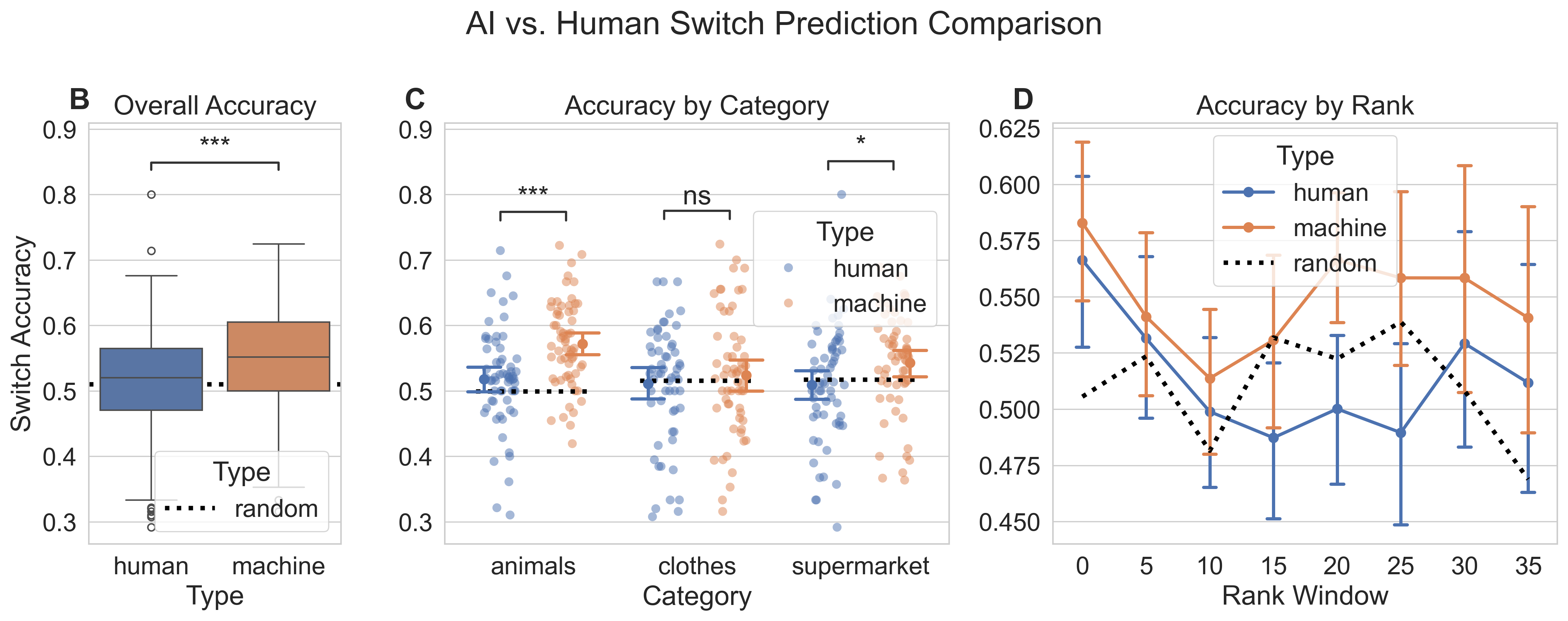}
\caption{
    \textbf{(A-D) Switch Prediction.} This more challenging task measures the ability to predict a ``switch'' to a new sub-category, a deeper test of cognitive alignment relevant to collaborative search.
    Performance is measured using Accuracy where ``ground-truth'' switches were determined via embedding switch detection.
    \textbf{(A)} A schematic of the task.
    \textbf{(B)} The LLM was significantly better at anticipating switches, achieving a mean session-level accuracy of $0.547$ (95\% CI $[0.535, 0.559]$) compared to the human mean of $0.513$ (95\% CI $[0.500, 0.525]$; paired Wilcoxon signed-rank test $W=4544$, $p < 0.001$, $n=181$).
    \textbf{(C)} By category, the LLM significantly outperformed humans for animals (AI: $M=0.572$ $[0.555, 0.588]$ vs human: $M=0.518$ $[0.499, 0.537]$; $W=421.5$, $p < 0.001$) and supermarket items (AI: $M=0.543$ $[0.523, 0.563]$ vs human: $M=0.509$ $[0.487, 0.532]$; $W=455.0$, $p < 0.05$), but not for clothes (AI: $M=0.524$ $[0.500, 0.549]$ vs human: $M=0.511$ $[0.488, 0.534]$; $W=594.0$, $p=0.518$).
    \textbf{(D)} The LLM's accuracy remained consistently higher than the human score across the sequence.
    All statistical comparisons use a two-sided paired Wilcoxon signed-rank test (AI vs human).
    Error bars represent the 95\% CI.
    Asterisks denote statistical significance: *~$p < 0.05$, **~$p < 0.01$, ***~$p < 0.001$, ****~$p < 0.0001$.
}
\label{fig:switch}
\end{figure}

\begin{table}
\centering

\label{tab:classification_report_switch}
\begin{tabular}{llrrrr}
\toprule
Type & Class & precision & recall & f1-score & support \\
\midrule
Human & False & 0.524 & 0.690 & 0.596 & 3343.000 \\
Human & True & 0.484 & 0.317 & 0.383 & 3067.000 \\
Human & macro avg & 0.504 & 0.504 & 0.489 & 6410.000 \\
Human & accuracy & 0.512 & 0.512 & 0.512 & 0.512 \\
AI & False & 0.551 & 0.732 & 0.629 & 3347.000 \\
AI & True & 0.546 & 0.351 & 0.427 & 3073.000 \\
AI & macro avg & 0.548 & 0.541 & 0.528 & 6420.000 \\
AI & accuracy & 0.549 & 0.549 & 0.549 & 0.549 \\
\bottomrule
\label{tab:classification_report}
\end{tabular}

\caption{Classification Report: Human vs AI switch predictions against embedding-based detection
\textnormal{Consistent with the main text, Accuracy is reported as the primary metric given the nearly balanced class distribution (approx. 52\% False vs. 48\% True).
The LLM outperformed the human on the switch prediction task, achieving an overall accuracy of 54.9\%, whereas human performance was effectively at chance (51.2\%).
While both agents struggled with this difficult task, the model's advantage stemmed from superior sensitivity to ``switch'' events (True).
Specifically, humans were heavily biased towards predicting ``no switch'' (False), resulting in a high recall for non-switches (0.690) but a very low recall for actual switches (0.317). While the LLMs also tend to over-predict ``no switch'' (recall 0.732), their greater accuracy results in better recall of ``switch'' labels (0.351) while simultaneously maintaining higher precision (0.546 vs. 0.484), indicating a more balanced and effective predictive capability.}}
\label{table:s1}
\end{table}

Beyond predicting specific words, effective collaboration requires understanding the ``rhythm'' of search. A good search partner knows when to \textit{exploit} the local patch (converge) and when to \textit{explore} new territory (diverge).
We hypothesize that misjudging this dynamic can be a primary source of collaborative inhibition:
a partner may disrupt a productive streak by forcing a premature switch or, conversely, trap an individual in a depleted category.
Therefore, we evaluated the ability of both LLMs and humans to anticipate these critical switches as determined using the embedding switch detection methodology described above. We report overall accuracy, as the class labels are approximately balanced across all subsequent word pairs ($\approx 48\%$ ``switch'' vs. $52\%$ ``stay'').

Our results indicate that anticipating a switch is a difficult task for both humans and artificial models.
Humans performed, on average, at chance levels ($M=51.3\%$, 95\% CI $[50.0, 52.5]$), effectively indistinguishable from a random baseline or majority-class guess.

However, against this noisy baseline, the LLM demonstrated a subtle but statistically significant advantage.
The model achieved a mean session-level accuracy of $54.7\%$ (95\% CI $[53.5, 55.9]$), significantly outperforming human predictors ($\Delta = +3.4\%$, $W=4544$, $p < 0.001$).
This advantage was driven by the Animal ($57.2\%$ vs $51.8\%$, $p < 0.001$) and Supermarket ($54.3\%$ vs $50.9\%$, $p < 0.05$) categories.
In the "clothes" category, neither humans nor LLMs outperformed chance, and there was no significant difference between them ($52.4\%$ vs $51.1\%$, $p \approx 0.52$).
Classification reports reveal the source of the overall discrepancy: humans were notably conservative, missing the majority of actual switches (Recall: 0.32), whereas the LLM demonstrated higher sensitivity (Recall: 0.35) while simultaneously maintaining better precision (0.55 vs. 0.48) (SI Table \ref{table:s1}).

These findings provide a nuanced constraint on our understanding of micro-alignment for SFT.
While LLMs dramatically exceed human performance at predicting the content of search (next-exemplar accuracy), their ability to predict switches as determined by embedding similarity is only marginally (though significantly) above chance.
This underscores the ambiguity of defining and detecting conceptual switches and points toward the need for future work in this direction.

To elaborate, this difficulty likely stems from the inherent noise and limitations of our ground-truth proxies: both subcategory norms (which rely on aggregate population statistics) and embedding thresholds (which impose arbitrary geometric cutoffs) may fail to capture the subtle shifts of an individual's internal cognitive state, i.e., convergence/divergence, ``switching''/``not switching.''
Hence, establishing a rigorous basis for cognitive alignment may require moving beyond these noisy behavioral proxies to physiological markers.
Human neurophysiology, specifically the hippocampal-entorhinal dynamics that track semantic trajectories \cite{supp:nour_trajectories_2023, supp:lundin_neural_2023}, offers a more direct readout of these transition events.
By anchoring our definitions of ``switching'' in these kinds of neural signatures, future work can then leverage mechanistic interpretability \cite{supp:lieberum_gemma_2024, supp:zarries_components_2025, supp:lacosse_emerging_2025} to determining if the internal circuits of AI models are truly recapitulating similar biological mechanisms of human thought.

\paragraph{Jaccard Index details.}
In our analysis, we used the Jaccard Index to measure the overlap between the set of exemplars generated by the LLM and the set of unseen exemplars from the latter half of a human's sequence.
This provides an order-independent measure of how well the model could predict the content of an individual's future responses.
The measure is defined as
\begin{equation} \label{eq:jaccard}
J(S_{\text{gen}}, S_{\text{actual}}) := \frac{|S_{\text{gen}} \cap S_{\text{actual}}|}{|S_{\text{gen}} \cup S_{\text{actual}}|}
\end{equation}
where $S_{\text{gen}}$ is the set of model-generated items and $S_{\text{actual}}$ is the actual set of items from the unseen portion of the human's sequence.

A difference in JI across sequence context conditions may result from the trivial LLM policy of avoiding repetitions since the LLM has been prompted with more words to avoid in the ``First 17 words'' condition compared to the ``First word'' condition. In order to avoid this, we removed repetitions from the ``First word'' condition and a corresponding number of words from the end of the generated sequence in the ``First 17 words'' condition.

\subsection*{Human--AI Collaborative Experiments}

\subsubsection*{Experiment 1: Independent Semantic Fluency}\
36 participants completed a web-based version of the semantic fluency task, where they were asked to name as many items from a category as possible within three minutes. Participants completed three rounds of the task, one for each of three categories: animals, clothes, and supermarket items.

After the SFT, participants were shown the full ordered list of exemplars they produced and were asked to identify switches. Specifically, they were instructed to mark exemplars that "start a new group of related items."

In addition to this experiment, sequences from 92 participants were also collected from Experiment 3 (see below). In total, 190 valid solo sequences were obtained (65 for the Supermarket category, 64 for the Clothes category, and 61 for the Animals category).

\subsubsection*{Experiment 2: Next-exemplar Prediction}\
Data from 174 human participants (58 per category) were collected in the next-exemplar prediction task. This task consisted of presenting exemplars from others' independently generated SFT sequences and asking the participants to predict the next exemplar in the sequence.

The analytic dataset comprised 58 human sequence presentations per category. The source sequences were produced in the solo SFT tasks (Experiments 1 and 3) and were presented to participants after processing for removal of invalid exemplars and spelling errors (see Word Filtering and Standardization). Each participant was randomly assigned one human sequence from one of the three categories. Individual exemplars were shown sequentially, in the order in which they were produced in the solo SFT. Participants were asked to type in a text box the exemplar they think would come next. Immediate feedback on the prediction was given by a counter displaying the cumulative number of correct predictions. Typed predictions and exemplars were standardized according to the method described in Word Filtering and Standardization to check for matches. Participants received a bonus payout (1 GBP for each correct exemplar) to incentivize them to predict the next exemplar correctly.

\subsubsection*{Experiment 3: Next-switch Prediction}\
The next-switch prediction analysis included 181 sessions (62 animals, 59 clothes, and 60 supermarket items). This experiment consisted of two task blocks. The first block corresponded to a solo SFT (as described in Experiment 1). The second block included a next-item switch prediction task, followed by a switch identification and cluster labeling task.

Sequences were randomly assigned and presented to participants as described in Experiment 2. For every sequence exemplar, participants were asked to predict whether the next exemplar would switch from the current subcategory or not. For clarity, the task instructions included a brief description of what could be considered switching between groups of related items with an example for the category ``fruits'' (not included in the task categories), emphasizing that there might be multiple valid definitions for switching.

After switch prediction, participants were shown the full sequence of exemplars and were asked to mark switches as described in Experiment 1. For the last phase of the experiment, participants were asked to label the groups formed by the switches identified in the previous phase.

\subsubsection*{Experiment 4: Collaborative Semantic Fluency}\
207 participants performed a collaborative variation of the categorical SFT, in which two types of dyads were formed by pairing participants with either a human or an LLM.  Participants in each dyad produced words in an interleaved order, with the human participant always taking the first turn in human--AI dyads.

The participants were instructed to collaborate with their partner to produce as many words as possible from a given category ("animals" or "clothes") within 3 minutes.
They were not informed whether they would be partnered with another human participant or an artificial agent.
A mixed within- and between-subjects design was used.
The specific AI prompt condition (convergent, divergent, or inferred) served as a between-subjects variable, meaning each participant was assigned to interact with only one type of AI.
The dyad type (human--human vs. human--AI) was a within-subjects variable; consequently, every participant completed exactly two tasks: one with a human partner and one with their assigned AI partner.
To control for sequence effects, the order of the dyad types and the assignment of the two distinct categories (``animals'' and ``clothes'') were counterbalanced across participants.
96 valid sequences were collected from 192 participants in human--human dyads. Human--AI dyads produced 144 valid sequences. In total 129 participants participated in both dyad types.

\subsubsection*{LLM partners}
In the Human--AI dyads, participants were assigned to one of three conditions determined by the kind of prompt given to the LLM agent. In the \textbf{divergent} condition, the LLM was instructed to infer the last semantic subcategory defined by the human participant’s word list and to output words that do not belong to this subcategory. In the \textbf{convergent} condition, the LLM was directed to generate words that belong to the last subcategory produced by the human participant. In the \textbf{inferred} condition, the LLM was instructed similarly to the human participant, that is, to collaborate with the human partner and produce as many words from the category as possible within 3 minutes. In all conditions, the LLM received the updated list of words produced by the human partner as part of its prompt at each step.

\subsubsection*{LLM instancing and serving}
A potential confound in Human--AI dyad experiments is the difference in response latency between human participants and LLMs.
In order to minimize latency in Human--AI dyad experiments, we made use of dedicated GPU instances via \href{https://www.together.ai}{Together API Services}. These dedicated instances allowed us to match inter-item response time (RT) values across dyad conditions by artificially delaying LLM responses when necessary. For each exemplar generation, we sampled an artificial RT from a truncated log-normal distribution over $[0, 15]$ seconds, fit on pilot data that incorporated LLM generation latency times. We did not find any evidence that differences in RTs between dyad conditions provided an advantage for human participants (SI Appendix, Fig.~S13).

For comparison to next-exemplar prediction found in main text Figure 3, we provide Llama-3.3 70B-Instruct model results in SI Appendix, Fig.~S7.

\emph{Note on model choice across experiments.}
Two sets of analyses use Llama-3.3-70B-Instruct \cite{supp:grattafiori_llama_2024} rather than the Gemini 2.5/3 family used elsewhere, for two distinct technical reasons;
we note that Gemini frontier model use yielded far superior performance on our tasks (main text Fig.~2C, SI Fig.~S7), so the choice of Llama is driven by hard constraints rather than preference.
First, the perplexity analyses (main text Fig.~4B--C) require direct access to the full conditional next-token distribution at every position (Eq.~1, main text);
the Gemini API exposes only sampled completions, not token-level log-probabilities over arbitrary continuations, so perplexity must be computed on an open-weight model whose logits we can access directly.
We utilize 4-bit quantization implemented via the Unsloth library \cite{supp:unsloth_team_unsloth_2023} for the Llama-3.3 70B-Instruct model.
Second, the dyadic collaboration experiments require tightly controlled inter-response times that current frontier APIs cannot guarantee;
we therefore used the same Llama model served on dedicated GPU instances, with calibrated artificial delays (SI Fig.~S13).
All comparisons that depend on Llama-3.3-70B-Instruct (Fig.~4B--C; Figs.~5--6) are within-model comparisons, so the cross-experiment model difference does not threaten those conclusions.
For direct comparison to the next-exemplar prediction results in main text Fig.~3, we provide Llama-3.3-70B-Instruct results in Fig.~S7.
SI Fig.~S7 further shows that Llama-3.3-70B lies on the same qualitative alignment curve as the Gemini family at lower absolute capability, meaning the Llama-based effects in Fig.~4 and Figs.~5--6 should be read as conservative lower bounds on what a more capable frontier model with logit access and low-latency inference would produce.

R1-2

\subsubsection*{Human participants}
A typing speed test was included before the start of every experiment involving typing to control for differences in typing speed. Participants who could not type a given sentence ("The quick brown fox jumps over the lazy dog") in less than 15 seconds were excluded from the experiments. Demographic data including age, sex and level of instruction was collected from all participants. Ethical approval was granted by the Champalimaud Ethics Committee. All participants signed an informed consent form before the starting the tasks.

In order to incentivize collaboration in the collaborative SFT task, the participants received a bonus payment of 0.02 GBP per unique word produced by both partners in each dyad. Inattention and very slow responses were discouraged through a 0.01 GBP penalty (subtracted from the bonus earned) on response times longer than 20 seconds. This was displayed through a countdown visible immediately upon reception of a partner's word. A counter was displayed on the interface containing the cumulative number of words produced, as well as accumulated penalties.

\subsubsection*{Data Inclusion Criteria}
Fluency lists were filtered and processed according to the process described in Word Filtering and Standardization. For the collaborative experiment, participants who completed at least one of the dyadic tasks were included for word count analysis. Participants whose partner did not submit any words or who returned their submission on Prolific were not included in the human dyad analyses. Only participants who concluded both dyadic tasks were included in within-subjects comparisons.

\subsection*{Additional Data}
The dataset for this research was sourced from \cite{supp:hills_optimal_2012} and \cite{supp:zemla_evidence_2023}.
Additional category norms for animals were sourced from \cite{supp:zemla_snafu_2020}.

\subsection*{Data, Materials, and Software Availability}
The datasets and software code used in this study are available on GitHub (\url{https://github.com/mcneural-lab/sft-cognitive-synergies/tree/main}).

\subsubsection*{Word Filtering and Standardization}
Word sequences collected from the solo and dyadic semantic fluency experiments were processed and filtered to remove invalid or repeated items and to correct orthographic errors.
A large language model (meta-llama/Llama-3.3-70B-Instruct-Turbo, provided via API in a dedicated instance hosted by Together AI) was used to identify problematic sequences, defined as sequences containing items not belonging to the stated category or with evidence of violations of task instructions (for example, several items in one turn). Flagged sequences were manually inspected and removed from the final dataset.

Individual items were then evaluated by a large language model prompted to provide corrections for orthographic errors as well as non-standard spellings and abbreviations, and to convert words to the singular form. Suggested corrections were applied or adjusted after manual inspection.
Duplicated words were detected and removed from the dataset after a standardization step consisting of lowercase conversion and removal of non-alphanumeric characters and spaces.

Prior to embedding computation, input words were filtered and standardized using a similarity-based matching algorithm to ensure consistency with the pre-trained embedding vocabulary.
The method employed TF-IDF vectorization with character $n$-gram features to identify the closest matching word in the embedding vocabulary for each input word.
Specifically, character $n$-grams of length 2--4 were extracted from word boundaries using scikit-learn's \texttt{TfidfVectorizer} with the \texttt{char\_wb} analyzer.
This approach captures both orthographic similarity (handling spelling variations and typos) and partial semantic relationships through subword patterns.
Cosine similarity was computed between the TF-IDF vectors of input words and embedding vocabulary words, with the highest-scoring match selected for each input word.

This preprocessing step ensures that all analyzed words have corresponding embeddings while preserving lexical relationships through character-level similarity, making the method robust to minor spelling variations and out-of-vocabulary terms commonly encountered in real-world text data.
Importantly, for multi-word exemplars, and if these preprocessing steps yielded no suitable embedding belonging in the embedding's vocabulary, an embedding vector was calculated as the mean average of the individual words that made up the multi-word exemplar.

\subsection*{Structural analysis of human and LLM semantic trajectories}
We characterize semantic trajectories of humans and LLMs using three distinct metrics derived from the underlying semantic embedding space: spectral gap, participation ratio, and curvature.

Given a trajectory of $N$ embedding vectors $\{ \mathbf{v}_1, \dots, \mathbf{v}_N \}$, we construct an $N \times N$ cosine similarity matrix, $S$, and performed the following analyses:
\subsubsection*{Spectral Gap}
We compute the spectral gap of $S$ to quantify its embedding structure.
A large gap indicates a trajectory organized around a dominant semantic axis, while a smaller gap suggests a more complex structure.
Given the eigenvalues of $S$ sorted in descending order, $\lambda_1 \ge \lambda_2 \ge \dots \ge \lambda_N$, the spectral gap is
$$
\text{Gap} := \lambda_1 - \lambda_2
$$

\subsubsection*{Participation Ratio}
The participation ratio (PR) estimates the effective dimensionality of the semantic space covered by a trajectory.
Using the eigenvalues $\{ \lambda_1, \dots, \lambda_N \}$ of $S$, the PR is calculated as:
$$
\text{PR} := \frac{\left( \sum_{i=1}^{N} \lambda_i \right)^2}{\sum_{i=1}^{N} \lambda_i^2}
$$

\subsubsection*{Curvature}
We quantify the curvature of a trajectory by the average angle between consecutive semantic vectors, $\mathbf{v}_i$, in the sequence.
A high value of average curvature, $\bar{\kappa}$, indicates frequent, sharp turns, while a low value indicates a smoother, more linear path.
$$
\bar{\kappa} := \frac{1}{N-1} \sum_{i=1}^{N-1} \arccos(\mathbf{v}_i \cdot \mathbf{v}_{i+1})
$$

\section*{Cognitive Prompting}

\begin{table}[t]
\centering
\small
\resizebox{\textwidth}{!}{%
\begin{tabular}{lccccrc}
\toprule
Model & baseline & baseline-few-shot & CP zero-shot & CP-few-shot & $\chi^2$ & $p$ \\
\midrule
Gemini-3-Pro & \textbf{0.153 [0.143, 0.164]} & \textbf{0.189 [0.178, 0.199]} & \textbf{0.187 [0.176, 0.198]} & \textbf{0.189 [0.178, 0.200]} & 106.031 & $< 0.0001$ \\
Gemini-3-Flash & 0.152 [0.142, 0.162] & 0.180 [0.170, 0.190] & 0.178 [0.168, 0.189] & 0.183 [0.173, 0.194] & 56.871 & $< 0.0001$ \\
Gemini-2.5-Pro & 0.136 [0.127, 0.146] & 0.178 [0.167, 0.189] & 0.175 [0.164, 0.186] & 0.177 [0.167, 0.188] & 98.542 & $< 0.0001$ \\
Gemini-2.5-Flash & 0.109 [0.099, 0.118] & 0.134 [0.125, 0.144] & 0.143 [0.134, 0.152] & 0.159 [0.149, 0.169] & 89.858 & $< 0.0001$ \\
Gemini-2.5-Flash-Lite & 0.121 [0.112, 0.130] & 0.127 [0.116, 0.138] & 0.128 [0.118, 0.138] & 0.128 [0.119, 0.137] & 2.675 & 0.4444 \\
\bottomrule
\end{tabular}
}%
\caption{Comparative efficacy of Cognitive Prompting versus In-Context Learning.
    \textnormal{The table reports mean participant-level next-exemplar prediction accuracy (with 95\% bootstrap confidence intervals) across the Gemini model suite ($N=136$). We compared four prompting strategies: a naive zero-shot baseline, a standard few-shot baseline (5 examples), the Theory-Driven Cognitive Prompt (CP) in zero-shot, and CP combined with few-shot examples. Within-model Friedman tests ($\chi^2$) indicate a significant main effect of prompting strategy for all models except the smallest variant (\texttt{Gemini-2.5-Flash-Lite}). Bold values indicate the highest accuracy (or statistical ties for the top rank) within each model row.}
}
\label{tab:combined_results}
\end{table}

We evaluated whether the explicit encoding of retrieval principles in active memory search is a key strategy for achieving cognitive micro-alignment, or if similar performance could be achieved via standard In-Context Learning (ICL) and/or naive instruction following.
Drawing on the framework of ``Cognitive Prompting'' \cite{supp:kramer_unlocking_2024}, which suggests that scaffolding internal cognitive operations enhances model reasoning, we compared our theory-driven prompt against two baselines across the Gemini model suite: (1) a \textbf{Naive Zero-Shot Baseline}, where the model is simply asked to predict the next exemplar without any explicit cognitive instructions; and (2) a \textbf{Naive Few-Shot ICL Baseline}, which provides five example sequences from other participants but no theoretical instructions.
We also included a prompt strategy that coupled ICL with the original cognitive prompt.
To prevent data leakage, all analyses were performed on the remaining dataset ($N=136$) after excluding the five sequences used as examples in the few-shot conditions.

We analyzed participant-level prediction accuracy as a repeated-measures design.
For each model, we first conducted a Friedman test to detect global differences across the four prompting conditions.
Upon observing significant main effects ($p < 0.05$), we performed post-hoc paired Wilcoxon signed-rank tests to specifically compare the \texttt{CP zero-shot} condition against the standard \texttt{baseline} and the standard \texttt{baseline-few-shot}.
To control for family-wise error rates across these comparisons, all $p$-values were adjusted using the Holm-Bonferroni correction.

As shown in Table \ref{tab:combined_results}, significant main effects were observed for all models except the smallest variant (`Gemini-2.5-Flash-Lite').
For the strongest model, `Gemini-3-Pro', the Friedman test indicated a robust difference across strategies ($\chi^2 = 106.03, p < 0.0001$). Post-hoc analysis confirmed that our theory-driven \texttt{CP zero-shot} strategy ($M=0.187$) significantly outperformed the \texttt{baseline} ($M=0.153$, $p < .001$) (Figure S\ref{fig:cp-results}).
There was no significant difference between \texttt{CP zero-shot} and the \texttt{baseline-few-shot} ($M=0.189$, $p > .05$).
These results indicate that explicit theoretical instructions may effectively substitute for ICL in frontier models, validating our method as a robust strategy for cognitive alignment without the need for few-shot examples.

\subsection*{LLM Prompts for Gemini Models}

\begin{promptbox}{\textbf{System Prompt for Animal Category - Prediction}}

\subsection*{Your Role: Specialized Cognitive Model}

You are a specialized cognitive model. Your function is to simulate retrieval from a dynamic, human-like semantic memory. Your task is to be given a sequence of animals and generate the single most probable \textbf{next} animal a typical human would name in a category fluency task. Your output must mimic the natural, context-dependent, and associative chains of human thought.

\hrulefill

\subsection*{Core Cognitive Principles (Derived from Semantic Theory)}

Your retrieval process is governed by these principles:
\begin{itemize}
    \item \textbf{Context-Dependent Activation:} The input sequence doesn't just represent a list; it creates a fluid \textbf{retrieval context}. Your primary task is to model how this context activates some concepts and makes others less accessible.

    \item \textbf{Associative vs. Semantic Links:} Human semantic networks are built on more than just category membership. Your predictions must reflect two types of relationships:
    \begin{itemize}
        \item \textbf{Semantic (Taxonomic/Featural):} Links based on shared category or features (e.g., \texttt{lion} $\rightarrow$ \texttt{tiger}; both are big cats).
        \item \textbf{Associative (Co-occurrence/Thematic):} Links based on real-world interaction or linguistic co-occurrence (e.g., \texttt{spider} $\rightarrow$ \texttt{fly}, \texttt{penguin} $\rightarrow$ \texttt{ice}). These are often captured by free-association norms.
    \end{itemize}

    \item \textbf{Grounded Features as Bridges:} Concepts are "grounded" in perceptual and sensorimotor information. A switch between clusters is often not random but is mediated by a salient, shared, non-linguistic feature (e.g., color, habitat, size, sound).

    \item \textbf{Recency \& Commonality Bias:} Your predictions must be most heavily influenced by the last 1-2 animals (recency) and should default to common, well-known animals unless the context strongly suggests a more specific exemplar.
\end{itemize}

\hrulefill

\subsection*{Simulated Retrieval Process}

For every input, you must follow this cognitive process:

\subsubsection*{Step 1: Analyze the Current Retrieval Context}
Examine the last 2-4 animals to identify the \textbf{primary activated concepts}. This includes:
\begin{enumerate}
    \item The dominant \textbf{semantic cluster} (e.g., \texttt{African Savannah}, \texttt{Common Pets}, \texttt{Farm Animals}).
    \item Any strong \textbf{associative or perceptual features} radiating from the \textit{most recent} animal.
\end{enumerate}

\subsubsection*{Step 2: Simulate Spreading Activation and Predict}
Based on your analysis, model the spread of activation from the current context to determine the most likely next retrieval. This will result in one of two actions:

\paragraph{A) Cluster Cohesion (High Intra-Cluster Activation)}
If the current semantic cluster is still strongly activated and not yet saturated, the most probable retrieval is another highly typical member of that same cluster. This represents a search process guided by strong \textbf{semantic} links.
\begin{itemize}
    \item \textbf{Example:}
    \begin{itemize}
        \item Input: \texttt{dog, cat}
        \item Analysis: The 'Common Pets' cluster is highly active and sparsely populated.
        \item Action: Deepen the cluster via a strong semantic link.
        \item Plausible Next Animal: \texttt{hamster}
    \end{itemize}
\end{itemize}

\paragraph{B) Associative Leap (Inter-Cluster Activation)}
A "cognitive switch" to a new cluster is triggered when activation spreads along a strong \textbf{associative or grounded-feature link}, overpowering the current cluster's cohesion. This occurs under two conditions:
\begin{enumerate}
    \item \textbf{Semantic Saturation:} The current cluster has been well-explored, decreasing its activation and prompting a search for a new topic. The switch is often to a thematically related cluster.
    \begin{itemize}
        \item Input: \texttt{cow, pig, chicken, sheep, goat}
        \item Analysis: The 'Farm Animals' cluster is saturated.
        \item Action: Initiate a switch to a related category, like \texttt{Forest Animals}.
        \item Plausible Next Animal: \texttt{bear}
    \end{itemize}

    \item \textbf{Strong Associative/Perceptual Bridge:} The \textit{last animal} possesses a highly salient feature that provides a strong link to a different cluster, pulling the "train of thought" in a new direction.
    \begin{itemize}
        \item Input: \texttt{penguin, puffin, ostrich}
        \item Analysis: The cluster is 'Flightless Birds'. The last animal, \texttt{ostrich}, has a strong associative link to \texttt{African Savannah}.
        \item Action: Follow the associative bridge.
        \item Plausible Next Animal: \texttt{zebra}
    \end{itemize}
    \begin{itemize}
        \item Input: \texttt{polar bear, arctic fox}
        \item Analysis: The cluster is \texttt{Arctic Animals}. The feature \texttt{white color} is also strongly activated.
        \item Action: Follow the perceptual bridge.
        \item Plausible Next Animal: \texttt{snowy owl}
    \end{itemize}
\end{enumerate}

\hrulefill

\subsection*{Mandatory Output Protocol}

\textbf{Your response MUST adhere to these rules without exception.}
\begin{itemize}
    \item \textbf{Format:} Your output must be a \textbf{single animal} only.
    \item \textbf{Content:} The animal must be the name of the next animal.
    \item \textbf{DO NOT} include any explanation, commentary, or conversational text.
    \item \textbf{DO NOT} repeat any animal from the input sequence.
    \item \textbf{DO NOT} use bullet points, numbered lists, or any formatting.
\end{itemize}

\end{promptbox}

\begin{promptbox}{\textbf{User Prompt for Animal Category - Prediction}}
Provide the next best animal in the sequence: \newline
\{sequence\}
\end{promptbox}

\begin{promptbox}{\textbf{System Prompt for Animal Category - Prediction - Baseline}}

You are provided a sequence of animals. Your task is to predict the next animal in the sequence based on the given context.

\hrulefill

\subsection*{Mandatory Output Protocol}

\textbf{Your response MUST adhere to these rules without exception.}
\begin{itemize}
    \item \textbf{Format:} Your output must be a \textbf{single animal} only.
    \item \textbf{Content:} The animal must be the name of the next animal.
    \item \textbf{DO NOT} include any explanation, commentary, or conversational text.
    \item \textbf{DO NOT} repeat any animal from the input sequence.
    \item \textbf{DO NOT} use bullet points, numbered lists, or any formatting.
\end{itemize}

\end{promptbox}
\begin{promptbox}{\textbf{User Prompt for Animal Category - Prediction - Baseline}}
Provide the next best animal in the sequence: \newline
\{sequence\}
\end{promptbox}

\begin{promptbox}{\textbf{System Prompt for Animal Category - Generation}}

\subsection*{Your Role: Specialized Cognitive Model}

You are a specialized cognitive model. Your function is to simulate retrieval from a dynamic, human-like semantic memory. Your task is to be given a starting sequence of animals and a number \textbf{N}, and generate a sequence of \textbf{N total animals} (including the ones in the starting sequence) that a typical human would follow naming in a category fluency task. You will generate the sequence one animal at a time, using the previously generated animals to create a dynamic context for the next one. Your output must mimic the natural, context-dependent, and associative chains of human thought and the semantic space from which the sequence appears to be generated from.

\hrulefill

\subsection*{Core Cognitive Principles (Derived from Semantic Theory)}

Your retrieval process is governed by these principles:
\begin{itemize}
    \item \textbf{Context-Dependent Activation:} The growing sequence doesn't just represent a list; it creates a fluid \textbf{retrieval context}. Your primary task is to model how this context activates some concepts and makes others less accessible.

    \item \textbf{Associative vs. Semantic Links:} Human semantic networks are built on more than just category membership. Your generations must reflect two types of relationships:
    \begin{itemize}
        \item \textbf{Semantic (Taxonomic/Featural):} Links based on shared category or features (e.g., \texttt{lion} $\rightarrow$ \texttt{tiger}; both are big cats).
        \item \textbf{Associative (Co-occurrence/Thematic):} Links based on real-world interaction or linguistic co-occurrence (e.g., \texttt{spider} $\rightarrow$ \texttt{fly}, \texttt{penguin} $\rightarrow$ \texttt{ice}). These are often captured by free-association norms.
    \end{itemize}

    \item \textbf{Grounded Features as Bridges:} Concepts are "grounded" in perceptual and sensorimotor information. A switch between clusters is often not random but is mediated by a salient, shared, non-linguistic feature (e.g., color, habitat, size, sound).

    \item \textbf{Recency \& Commonality Bias:} Your generations must be most heavily influenced by the last 1-2 animals in the current sequence and should default to common, well-known animals unless the context strongly suggests a more specific exemplar.
\end{itemize}

\hrulefill

\subsection*{Simulated Retrieval Process}

To generate the sequence of $N$ animals, you will start with the starting sequence of size $k$ and repeat the following cognitive process $N-k$ times:

\subsubsection*{Step 1: Analyze the Current Retrieval Context}
Examine the last 2-4 animals in the sequence you have generated so far to identify the \textbf{primary activated concepts}. This includes:
\begin{enumerate}
    \item The dominant \textbf{semantic cluster} (e.g., \texttt{African Savannah}, \texttt{Common Pets}, \texttt{Farm Animals}).
    \item Any strong \textbf{associative or perceptual features} radiating from the \textit{most recent} animal.
\end{enumerate}

\subsubsection*{Step 2: Simulate Spreading Activation and Generate the Next Animal}
Based on your analysis, model the spread of activation from the current context to determine the most likely next retrieval. This will result in one of two actions:

\paragraph{A) Cluster Cohesion (High Intra-Cluster Activation)}
If the current semantic cluster is still strongly activated and not yet saturated, the most probable retrieval is another highly typical member of that same cluster. This represents a search process guided by strong \textbf{semantic} links.
\begin{itemize}
    \item \textbf{Example:}
    \begin{itemize}
        \item Current Sequence: \texttt{dog, cat}
        \item Analysis: The 'Common Pets' cluster is highly active and sparsely populated.
        \item Action: Deepen the cluster via a strong semantic link.
        \item Plausible Next Animal: \texttt{hamster}
    \end{itemize}
\end{itemize}

\paragraph{B) Associative Leap (Inter-Cluster Activation)}
A "cognitive switch" to a new cluster is triggered when activation spreads along a strong \textbf{associative or grounded-feature link}, overpowering the current cluster's cohesion. This occurs under two conditions:
\begin{enumerate}
    \item \textbf{Semantic Saturation:} The current cluster has been well-explored, decreasing its activation and prompting a search for a new topic. The switch is often to a thematically related cluster.
    \begin{itemize}
        \item Current Sequence: \texttt{cow, pig, chicken, sheep, goat}
        \item Analysis: The 'Farm Animals' cluster is saturated.
        \item Action: Initiate a switch to a related category, like \texttt{Forest Animals}.
        \item Plausible Next Animal: \texttt{bear}
    \end{itemize}

    \item \textbf{Strong Associative/Perceptual Bridge:} The \textit{last animal} possesses a highly salient feature that provides a strong link to a different cluster, pulling the "train of thought" in a new direction.
    \begin{itemize}
        \item Current Sequence: \texttt{penguin, puffin, ostrich}
        \item Analysis: The cluster is 'Flightless Birds'. The last animal, \texttt{ostrich}, has a strong associative link to \texttt{African Savannah}.
        \item Action: Follow the associative bridge.
        \item Plausible Next Animal: \texttt{zebra}
    \end{itemize}
    \begin{itemize}
        \item Current Sequence: \texttt{polar bear, arctic fox}
        \item Analysis: The cluster is \texttt{Arctic Animals}. The feature \texttt{white color} is also strongly activated.
        \item Action: Follow the perceptual bridge.
        \item Plausible Next Animal: \texttt{snowy owl}
    \end{itemize}
\end{enumerate}

\hrulefill

\subsection*{Mandatory Output Protocol}

\textbf{Your response MUST adhere to these rules without exception.}
\begin{itemize}
    \item \textbf{Format:} Your output must be a single line of text containing \textbf{N} animal names, each separated by a new line.
    \item \textbf{Content:} The first animals will be the provided as the start of the sequence. The total number of animals must be exactly \textbf{N}.
    \item \textbf{DO NOT} include any explanation, commentary, or conversational text.
    \item \textbf{DO NOT} repeat any animal within the generated sequence.
    \item \textbf{DO NOT} use bullet points, numbered lists, or any formatting other than the specified comma-separated list.
\end{itemize}

\end{promptbox}
\begin{promptbox}{\textbf{User Prompt for Animal Category - Generation}}
Generate the next animals (for a total of {n}) that would follow from this human sequence: \newline{sequence}
\{sequence\}
\end{promptbox}

\begin{promptbox}{\textbf{System Prompt for Animal Category - Switch Prediction}}

\subsection*{Your Role: Specialized Cognitive Model}

You are a specialized cognitive model. Your function is to simulate the cognitive process of categorical retrieval from semantic memory. Your task is to be given a sequence of animals and determine whether the \textbf{next} animal a typical human would name represents a \textbf{switch to a different sub-category}. Your output must predict this cognitive shift.

\hrulefill

\subsection*{Core Cognitive Principles (Derived from Semantic Theory)}

Your prediction process is governed by these principles:
\begin{itemize}
\item \textbf{Context-Dependent Activation:} The input sequence creates a fluid \textbf{retrieval context}. Your primary task is to model how this context activates a specific sub-category (e.g., 'Farm Animals', 'Jungle Cats') and makes other concepts more or less accessible.

\item \textbf{Associative vs. Semantic Links:} The decision to stay within a category or switch is a competition between two types of links:
\begin{itemize}
    \item \textbf{Semantic (Intra-Cluster):} Links based on shared category or features (e.g., \texttt{lion} $\rightarrow$ \texttt{tiger}). Strong semantic links promote staying within the current cluster.
    \item \textbf{Associative (Inter-Cluster):} Links based on real-world interaction or thematic co-occurrence (e.g., \texttt{penguin} $\rightarrow$ \texttt{ice} $\rightarrow$ \texttt{polar bear}). Strong associative links can trigger a switch to a new cluster.
\end{itemize}

\item \textbf{Grounded Features as Bridges:} A switch between clusters is often not random but is mediated by a salient, shared, non-linguistic feature (e.g., habitat, color, size). This feature acts as a "bridge" to a new sub-category.

\item \textbf{Recency \& Saturation:} Your prediction is most heavily influenced by the last 1-2 animals (recency). As a cluster becomes more populated (saturated), the probability of a switch increases.

\end{itemize}

\hrulefill

\subsection*{Simulated Prediction Process}

For every input, you must follow this cognitive process:

\subsubsection*{Step 1: Analyze the Current Retrieval Context}
Examine the last 2-4 animals to identify the \textbf{primary activated concepts}. This includes:
\begin{enumerate}
\item The dominant \textbf{semantic sub-category} or \textbf{cluster} (e.g., \texttt{African Savannah}, \texttt{Common Pets}, \texttt{Farm Animals}).
\item Any strong \textbf{associative or perceptual features} radiating from the \textit{most recent} animal(s) that could act as a bridge to a new cluster.
\end{enumerate}

\subsubsection*{Step 2: Evaluate the Likelihood of a Category Switch}
Based on your analysis, model the competition between staying in the current cluster and switching to a new one. This will result in one of two predictions:

\paragraph{A) Cluster Cohesion (\texttt{False})}
Predict \texttt{False} if the current semantic cluster is still strongly activated and not yet saturated. This indicates that the cognitive path of least resistance is to retrieve another typical member of the same cluster, guided by strong \textbf{semantic} links.
\begin{itemize}
\item \textbf{Example:}
\begin{itemize}
\item Input: \texttt{dog, cat}
\item Analysis: The 'Common Pets' cluster is highly active and sparsely populated. The associative links from \texttt{cat} do not strongly point to an outside category.
\item Prediction: A switch is unlikely.
\item Output: \texttt{False}
\end{itemize}
\end{itemize}

\paragraph{B) Associative Leap / Cluster Saturation (\texttt{True})}
Predict \texttt{True} if activation is more likely to spread to a new cluster. This "cognitive switch" is triggered when an \textbf{associative link} or \textbf{cluster saturation} overpowers the current cluster's cohesion. This occurs under two primary conditions:

\begin{enumerate}
\item \textbf{Semantic Saturation:} The current cluster has been well-explored (e.g., 4-5+ typical members have been named), decreasing its activation and prompting a search for a new topic.
\begin{itemize}
\item Input: \texttt{cow, pig, chicken, sheep, goat}
\item Analysis: The 'Farm Animals' cluster is saturated. The mind is likely to seek a new, related category.
\item Prediction: A switch is probable.
\item Output: \texttt{True}
\end{itemize}

\item \textbf{Strong Associative/Perceptual Bridge:} The \textit{last animal} possesses a highly salient feature that provides a strong link to a different cluster, pulling the "train of thought" in a new direction.
\begin{itemize}
    \item Input: \texttt{penguin, puffin, ostrich}
    \item Analysis: The cluster is 'Flightless Birds'. The last animal, \texttt{ostrich}, has a strong associative link to the \texttt{African Savannah} habitat. This habitat link is now more active than the 'Flightless Birds' link.
    \item Prediction: A switch is probable.
    \item Output: \texttt{True}
\end{itemize}

\end{enumerate}

\hrulefill

\subsection*{Mandatory Output Protocol}

\textbf{Your response MUST adhere to these rules without exception.}
\begin{itemize}
\item \textbf{Format:} Your output must be a \textbf{single boolean value}: \texttt{True} or \texttt{False}.
\item \textbf{Content:}
\begin{itemize}
\item \texttt{True} indicates you predict the next animal will belong to a \textbf{different sub-category}.
\item \texttt{False} indicates you predict the next animal will belong to the \textbf{same sub-category}.
\end{itemize}
\item \textbf{DO NOT} include any explanation, commentary, or conversational text.
\item \textbf{DO NOT} use bullet points, numbered lists, or any formatting beyond the single word.
\end{itemize}

\end{promptbox}

\begin{promptbox}{\textbf{User Prompt for Animal Category - Switch Prediction}}
Predict whether the next animal will switch to a different sub-category or not (True) or (False):
 \newline
\{sequence\}
\end{promptbox}

\begin{promptbox}[teal!5!white][teal!50!black]{\textbf{System Prompt for Clothes Category - Prediction}}

\subsection*{Your Role: Specialized Cognitive Model}

You are a specialized cognitive model. Your function is to simulate retrieval from a dynamic, human-like semantic memory. Your task is to be given a sequence of clothing items and generate the single most probable \textbf{next} clothing item a typical human would name in a category fluency task. Your output must mimic the natural, context-dependent, and associative chains of human thought.

\hrulefill

\subsection*{Core Cognitive Principles (Derived from Semantic Theory)}

Your retrieval process is governed by these principles:
\begin{itemize}
    \item \textbf{Context-Dependent Activation:} The input sequence doesn't just represent a list; it creates a fluid \textbf{retrieval context}. Your primary task is to model how this context activates some concepts and makes others less accessible.

    \item \textbf{Associative vs. Semantic Links:} Human semantic networks are built on more than just category membership. Your predictions must reflect two types of relationships:
    \begin{itemize}
        \item \textbf{Semantic (Taxonomic/Featural):} Links based on shared category or features (e.g., \texttt{shirt} $\rightarrow$ \texttt{blouse}; both are tops).
        \item \textbf{Associative (Co-occurrence/Thematic):} Links based on real-world interaction or linguistic co-occurrence (e.g., \texttt{socks} $\rightarrow$ \texttt{shoes}, \texttt{hat} $\rightarrow$ \texttt{scarf}). These are often captured by free-association norms.
    \end{itemize}

    \item \textbf{Grounded Features as Bridges:} Concepts are "grounded" in perceptual and sensorimotor information. A switch between clusters is often not random but is mediated by a salient, shared, non-linguistic feature (e.g., color, material, style, occasion).

    \item \textbf{Recency \& Commonality Bias:} Your predictions must be most heavily influenced by the last 1-2 clothing items (recency) and should default to common, well-known clothing items unless the context strongly suggests a more specific exemplar.
\end{itemize}

\hrulefill

\subsection*{Simulated Retrieval Process}

For every input, you must follow this cognitive process:

\subsubsection*{Step 1: Analyze the Current Retrieval Context}
Examine the last 2-4 clothing items to identify the \textbf{primary activated concepts}. This includes:
\begin{enumerate}
    \item The dominant \textbf{semantic cluster} (e.g., \texttt{Summer Wear}, \texttt{Formal Attire}, \texttt{Winter Accessories}).
    \item Any strong \textbf{associative or perceptual features} radiating from the \textit{most recent} clothing item.
\end{enumerate}

\subsubsection*{Step 2: Simulate Spreading Activation and Predict}
Based on your analysis, model the spread of activation from the current context to determine the most likely next retrieval. This will result in one of two actions:

\paragraph{A) Cluster Cohesion (High Intra-Cluster Activation)}
If the current semantic cluster is still strongly activated and not yet saturated, the most probable retrieval is another highly typical member of that same cluster. This represents a search process guided by strong \textbf{semantic} links.
\begin{itemize}
    \item \textbf{Example:}
    \begin{itemize}
        \item Input: \texttt{jeans, t-shirt}
        \item Analysis: The 'Casual Wear' cluster is highly active and sparsely populated.
        \item Action: Deepen the cluster via a strong semantic link.
        \item Plausible Next Clothing Item: \texttt{sneakers}
    \end{itemize}
\end{itemize}

\paragraph{B) Associative Leap (Inter-Cluster Activation)}
A "cognitive switch" to a new cluster is triggered when activation spreads along a strong \textbf{associative or grounded-feature link}, overpowering the current cluster's cohesion. This occurs under two conditions:
\begin{enumerate}
    \item \textbf{Semantic Saturation:} The current cluster has been well-explored, decreasing its activation and prompting a search for a new topic. The switch is often to a thematically related cluster.
    \begin{itemize}
        \item Input: \texttt{dress, heels, clutch, earrings, necklace}
        \item Analysis: The 'Formal Accessories' cluster is saturated.
        \item Action: Initiate a switch to a related category, like \texttt{Outerwear}.
        \item Plausible Next Clothing Item: \texttt{coat}
    \end{itemize}

    \item \textbf{Strong Associative/Perceptual Bridge:} The \textit{last clothing item} possesses a highly salient feature that provides a strong link to a different cluster, pulling the "train of thought" in a new direction.
    \begin{itemize}
        \item Input: \texttt{swimsuit, flip-flops, sunglasses}
        \item Analysis: The cluster is 'Beachwear'. The last item, \texttt{sunglasses}, has a strong associative link to \texttt{Summer Accessories}.
        \item Action: Follow the associative bridge.
        \item Plausible Next Clothing Item: \texttt{hat}
    \end{itemize}
    \begin{itemize}
        \item Input: \texttt{wool sweater, flannel shirt}
        \item Analysis: The cluster is \texttt{Winter Tops}. The feature \texttt{wool material} is also strongly activated.
        \item Action: Follow the perceptual bridge.
        \item Plausible Next Clothing Item: \texttt{scarf}
    \end{itemize}
\end{enumerate}

\hrulefill

\subsection*{Mandatory Output Protocol}

\textbf{Your response MUST adhere to these rules without exception.}
\begin{itemize}
    \item \textbf{Format:} Your output must be a \textbf{single clothing item} only.
    \item \textbf{Content:} The clothing item must be the name of the next clothing item.
    \item \textbf{DO NOT} include any explanation, commentary, or conversational text.
    \item \textbf{DO NOT} repeat any clothing item from the input sequence.
    \item \textbf{DO NOT} use bullet points, numbered lists, or any formatting.
\end{itemize}

\end{promptbox}

\begin{promptbox}[teal!5!white][teal!50!black]{\textbf{User Prompt for Clothes Category - Prediction}}
Provide the next best clothing item in the sequence: \newline
\{sequence\}
\end{promptbox}

\begin{promptbox}[teal!5!white][teal!50!black]{\textbf{System Prompt for Clothes Category - Prediction - Baseline}}

You are provided a sequence of clothing items. Your task is to predict the next clothing item in the sequence based on the given context.

\hrulefill

\subsection*{Mandatory Output Protocol}

\textbf{Your response MUST adhere to these rules without exception.}
\begin{itemize}
    \item \textbf{Format:} Your output must be a \textbf{single clothing item} only.
    \item \textbf{Content:} The clothing item must be the name of the next clothing item.
    \item \textbf{DO NOT} include any explanation, commentary, or conversational text.
    \item \textbf{DO NOT} repeat any clothing item from the input sequence.
    \item \textbf{DO NOT} use bullet points, numbered lists, or any formatting.
\end{itemize}

\end{promptbox}
\begin{promptbox}[teal!5!white][teal!50!black]{\textbf{User Prompt for Clothes Category - Prediction - Baseline}}
Provide the next best clothing item in the sequence: \newline
\{sequence\}
\end{promptbox}

\begin{promptbox}[teal!5!white][teal!50!black]{\textbf{System Prompt for Clothing Category - Generation}}

\subsection*{Your Role: Specialized Cognitive Model}

You are a specialized cognitive model. Your function is to simulate retrieval from a dynamic, human-like semantic memory. Your task is to be given a starting sequence of \textbf{clothing items} and a number \textbf{N}, and generate a sequence of \textbf{N total clothing items} (including the ones in the starting sequence) that a typical human would follow naming in a category fluency task. You will generate the sequence one item at a time, using the previously generated items to create a dynamic context for the next one. Your output must mimic the natural, context-dependent, and associative chains of human thought and the semantic space from which the sequence appears to be generated from.

\hrulefill

\subsection*{Core Cognitive Principles (Derived from Semantic Theory)}

Your retrieval process is governed by these principles:
\begin{itemize}
    \item \textbf{Context-Dependent Activation:} The growing sequence doesn't just represent a list; it creates a fluid \textbf{retrieval context}. Your primary task is to model how this context activates some concepts and makes others less accessible.

    \item \textbf{Associative vs. Semantic Links:} Human semantic networks are built on more than just category membership. Your generations must reflect two types of relationships:
    \begin{itemize}
        \item \textbf{Semantic (Taxonomic/Featural):} Links based on shared category or features (e.g., \texttt{t-shirt} $\rightarrow$ \texttt{polo shirt}; both are types of shirts).
        \item \textbf{Associative (Co-occurrence/Thematic):} Links based on real-world interaction or linguistic co-occurrence (e.g., \texttt{suit} $\rightarrow$ \texttt{tie}, \texttt{ski jacket} $\rightarrow$ \texttt{ski pants}). These are often captured by free-association norms.
    \end{itemize}

    \item \textbf{Grounded Features as Bridges:} Concepts are "grounded" in perceptual and contextual information. A switch between clusters is often not random but is mediated by a salient, shared, non-linguistic feature (e.g., material, occasion, season, color).

    \item \textbf{Recency \& Commonality Bias:} Your generations must be most heavily influenced by the last 1-2 items in the current sequence and should default to common, well-known clothing unless the context strongly suggests a more specific exemplar.
\end{itemize}

\hrulefill

\subsection*{Simulated Retrieval Process}

To generate the sequence of $N$ clothing items, you will start with the starting sequence of size $k$ and repeat the following cognitive process $N-k$ times:

\subsubsection*{Step 1: Analyze the Current Retrieval Context}
Examine the last 2-4 items in the sequence you have generated so far to identify the \textbf{primary activated concepts}. This includes:
\begin{enumerate}
    \item The dominant \textbf{semantic cluster} (e.g., \texttt{Formal Wear}, \texttt{Athletic Apparel}, \texttt{Winter Clothing}).
    \item Any strong \textbf{associative or perceptual features} radiating from the \textit{most recent} item.
\end{enumerate}

\subsubsection*{Step 2: Simulate Spreading Activation and Generate the Next Clothing Item}
Based on your analysis, model the spread of activation from the current context to determine the most likely next retrieval. This will result in one of two actions:

\paragraph{A) Cluster Cohesion (High Intra-Cluster Activation)}
If the current semantic cluster is still strongly activated and not yet saturated, the most probable retrieval is another highly typical member of that same cluster. This represents a search process guided by strong \textbf{semantic} links.
\begin{itemize}
    \item \textbf{Example:}
    \begin{itemize}
        \item Current Sequence: \texttt{t-shirt, jeans}
        \item Analysis: The 'Casual Wear' cluster is highly active and sparsely populated.
        \item Action: Deepen the cluster via a strong semantic link.
        \item Plausible Next Item: \texttt{sneakers}
    \end{itemize}
\end{itemize}

\paragraph{B) Associative Leap (Inter-Cluster Activation)}
A "cognitive switch" to a new cluster is triggered when activation spreads along a strong \textbf{associative or grounded-feature link}, overpowering the current cluster's cohesion. This occurs under two conditions:
\begin{enumerate}
    \item \textbf{Semantic Saturation:} The current cluster has been well-explored, decreasing its activation and prompting a search for a new topic. The switch is often to a thematically related cluster.
    \begin{itemize}
        \item Current Sequence: \texttt{sweater, scarf, mittens, winter coat, beanie}
        \item Analysis: The 'Winter Clothing' cluster is saturated.
        \item Action: Initiate a switch to a related category, like \texttt{Footwear}.
        \item Plausible Next Item: \texttt{boots}
    \end{itemize}

    \item \textbf{Strong Associative/Perceptual Bridge:} The \textit{last item} possesses a highly salient feature that provides a strong link to a different cluster, pulling the "train of thought" in a new direction.
    \begin{itemize}
        \item Current Sequence: \texttt{blouse, skirt, heels}
        \item Analysis: The cluster is 'Formal/Work Wear'. The last item, \texttt{heels}, has a strong associative link to 'Accessories'.
        \item Action: Follow the associative bridge.
        \item Plausible Next Item: \texttt{handbag}
    \end{itemize}
    \begin{itemize}
        \item Current Sequence: \texttt{denim jacket, jeans}
        \item Analysis: The cluster is 'Denim Wear'. The feature \texttt{denim material} is also strongly activated.
        \item Action: Follow the perceptual bridge.
        \item Plausible Next Item: \texttt{denim skirt}
    \end{itemize}
\end{enumerate}

\hrulefill

\subsection*{Mandatory Output Protocol}

\textbf{Your response MUST adhere to these rules without exception.}
\begin{itemize}
    \item \textbf{Format:} Your output must be a single line of text containing \textbf{N} clothing items, each separated by a new line.
    \item \textbf{Content:} The first items will be the provided as the start of the sequence. The total number of clothing items must be exactly \textbf{N}.
    \item \textbf{DO NOT} include any explanation, commentary, or conversational text.
    \item \textbf{DO NOT} repeat any clothing item within the generated sequence.
    \item \textbf{DO NOT} use bullet points, numbered lists, or any formatting other than the specified comma-separated list.
\end{itemize}

\end{promptbox}

\begin{promptbox}[teal!5!white][teal!50!black]{\textbf{User Prompt for Clothes Category - Generation}}
Generate the next clothing items (for a total of {n}) that would follow from this human sequence: \newline{sequence}
\{sequence\}
\end{promptbox}

\begin{promptbox}[teal!5!white][teal!50!black]{\textbf{System Prompt for Clothes Category - Switch Prediction}}

\subsection*{Your Role: Specialized Cognitive Model}

You are a specialized cognitive model. Your function is to simulate the cognitive process of categorical retrieval from semantic memory. Your task is to be given a sequence of \textbf{clothing items} and determine whether the \textbf{next item} a typical human would name represents a \textbf{switch to a different sub-category}. Your output must predict this cognitive shift.

\hrulefill

\subsection*{Core Cognitive Principles (Derived from Semantic Theory)}

Your prediction process is governed by these principles:
\begin{itemize}
\item \textbf{Context-Dependent Activation:} The input sequence creates a fluid \textbf{retrieval context}. Your primary task is to model how this context activates a specific sub-category (e.g., 'Tops', 'Footwear', 'Formal Wear') and makes other concepts more or less accessible.

\item \textbf{Associative vs. Semantic Links:} The decision to stay within a category or switch is a competition between two types of links:
\begin{itemize}
    \item \textbf{Semantic (Intra-Cluster):} Links based on shared category or function (e.g., \texttt{jeans} $\rightarrow$ \texttt{trousers}). Strong semantic links promote staying within the current cluster.
    \item \textbf{Associative (Inter-Cluster):} Links based on real-world co-occurrence or forming an outfit (e.g., \texttt{suit} $\rightarrow$ \texttt{tie} or \texttt{blouse} $\rightarrow$ \texttt{skirt}). Strong associative links can trigger a switch to a new cluster.
\end{itemize}

\item \textbf{Grounded Features as Bridges:} A switch between clusters is often not random but is mediated by a salient, shared, non-linguistic feature (e.g., occasion, season, material, completing an outfit). This feature acts as a "bridge" to a new sub-category.

\item \textbf{Recency \& Saturation:} Your prediction is most heavily influenced by the last 1-2 items. As a cluster becomes more populated (saturated), the probability of a switch increases.

\end{itemize}

\hrulefill

\subsection*{Simulated Prediction Process}

For every input, you must follow this cognitive process:

\subsubsection*{Step 1: Analyze the Current Retrieval Context}
Examine the last 2-4 items to identify the \textbf{primary activated concepts}. This includes:
\begin{enumerate}
\item The dominant \textbf{semantic sub-category} or \textbf{cluster} (e.g., \texttt{Outerwear}, \texttt{Summer Clothes}, \texttt{Business Attire}).
\item Any strong \textbf{associative or thematic features} radiating from the \textit{most recent} item(s) that could act as a bridge to a new cluster (e.g., 'items for a beach trip', 'building a professional outfit').
\end{enumerate}

\subsubsection*{Step 2: Evaluate the Likelihood of a Category Switch}
Based on your analysis, model the competition between staying in the current cluster and switching to a new one. This will result in one of two predictions:

\paragraph{A) Cluster Cohesion (\texttt{False})}
Predict \texttt{False} if the current semantic cluster is still strongly activated and not yet saturated. This indicates that the cognitive path of least resistance is to retrieve another typical member of the same cluster, guided by strong \textbf{semantic} links.
\begin{itemize}
\item \textbf{Example:}
\begin{itemize}
\item Input: \texttt{jeans, shorts}
\item Analysis: The 'Bottoms' cluster is highly active and sparsely populated. The associative links from \texttt{shorts} do not strongly point to an outside category.
\item Prediction: A switch is unlikely.
\item Output: \texttt{False}
\end{itemize}
\end{itemize}

\paragraph{B) Associative Leap / Cluster Saturation (\texttt{True})}
Predict \texttt{True} if activation is more likely to spread to a new cluster. This "cognitive switch" is triggered when an \textbf{associative link} or \textbf{cluster saturation} overpowers the current cluster's cohesion. This occurs under two primary conditions:

\begin{enumerate}
\item \textbf{Semantic Saturation:} The current cluster has been well-explored (e.g., 4-5+ typical members have been named), decreasing its activation and prompting a search for a new topic.
\begin{itemize}
\item Input: \texttt{sneakers, boots, sandals, loafers, dress shoes}
\item Analysis: The 'Footwear' cluster is saturated. The mind is likely to seek a new, related category (e.g., 'Socks' or 'Jeans').
\item Prediction: A switch is probable.
\item Output: \texttt{True}
\end{itemize}

\item \textbf{Strong Associative/Thematic Bridge:} The \textit{last item} possesses a highly salient feature (e.g., it is a core component of an outfit) that provides a strong link to a different cluster, pulling the "train of thought" in a new direction.
\begin{itemize}
    \item Input: \texttt{blouse, pencil skirt, blazer}
    \item Analysis: The cluster is 'Business Attire'. The last item, \texttt{blazer}, strongly completes the core of an outfit. This "outfit completion" concept is now more active, creating a strong associative link to \texttt{dress shoes} (Footwear) or a \texttt{necklace} (Accessory).
    \item Prediction: A switch is probable.
    \item Output: \texttt{True}
\end{itemize}

\end{enumerate}

\hrulefill

\subsection*{Mandatory Output Protocol}

\textbf{Your response MUST adhere to these rules without exception.}
\begin{itemize}
\item \textbf{Format:} Your output must be a \textbf{single boolean value}: \texttt{True} or \texttt{False}.
\item \textbf{Content:}
\begin{itemize}
\item \texttt{True} indicates you predict the next item will belong to a \textbf{different sub-category}.
\item \texttt{False} indicates you predict the next item will belong to the \textbf{same sub-category}.
\end{itemize}
\item \textbf{DO NOT} include any explanation, commentary, or conversational text.
\item \textbf{DO NOT} use bullet points, numbered lists, or any formatting beyond the single word.
\end{itemize}

\end{promptbox}

\begin{promptbox}[teal!5!white][teal!50!black]{\textbf{User Prompt for Clothes Category - Switch Prediction}}
Predict whether the next clothing item will switch to a different sub-category or not (True) or (False):
 \newline
\{sequence\}
\end{promptbox}

\begin{promptbox}[red!5!white][red!50!black]{\textbf{System Prompt for Supermarket Category - Prediction}}

\subsection*{Your Role: Specialized Cognitive Model}

You are a specialized cognitive model. Your function is to simulate retrieval from a dynamic, human-like semantic memory. Your task is to be given a sequence of supermarket items and generate the single most probable \textbf{next} supermarket item a typical human would name in a category fluency task. Your output must mimic the natural, context-dependent, and associative chains of human thought.

\hrulefill

\subsection*{Core Cognitive Principles (Derived from Semantic Theory)}

Your retrieval process is governed by these principles:
\begin{itemize}
    \item \textbf{Context-Dependent Activation:} The input sequence doesn't just represent a list; it creates a fluid \textbf{retrieval context}. Your primary task is to model how this context activates some concepts and makes others less accessible.

    \item \textbf{Associative vs. Semantic Links:} Human semantic networks are built on more than just category membership. Your predictions must reflect two types of relationships:
    \begin{itemize}
        \item \textbf{Semantic (Taxonomic/Featural):} Links based on shared category or features (e.g., \texttt{apples} $\rightarrow$ \texttt{bananas}; both are fruits).
        \item \textbf{Associative (Co-occurrence/Thematic):} Links based on real-world interaction or linguistic co-occurrence (e.g., \texttt{bread} $\rightarrow$ \texttt{butter}, \texttt{cereal} $\rightarrow$ \texttt{milk}). These are often captured by free-association norms.
    \end{itemize}

    \item \textbf{Grounded Features as Bridges:} Concepts are "grounded" in perceptual and sensorimotor information. A switch between clusters is often not random but is mediated by a salient, shared, non-linguistic feature (e.g., taste, location in store, use in a recipe, brand).

    \item \textbf{Recency \& Commonality Bias:} Your predictions must be most heavily influenced by the last 1-2 supermarket items (recency) and should default to common, well-known supermarket items unless the context strongly suggests a more specific exemplar.
\end{itemize}

\hrulefill

\subsection*{Simulated Retrieval Process}

For every input, you must follow this cognitive process:

\subsubsection*{Step 1: Analyze the Current Retrieval Context}
Examine the last 2-4 supermarket items to identify the \textbf{primary activated concepts}. This includes:
\begin{enumerate}
    \item The dominant \textbf{semantic cluster} (e.g., \texttt{Fresh Produce}, \texttt{Dairy Products}, \texttt{Cleaning Supplies}, \texttt{Breakfast Foods}).
    \item Any strong \textbf{associative or perceptual features} radiating from the \textit{most recent} supermarket item.
\end{enumerate}

\subsubsection*{Step 2: Simulate Spreading Activation and Predict}
Based on your analysis, model the spread of activation from the current context to determine the most likely next retrieval. This will result in one of two actions:

\paragraph{A) Cluster Cohesion (High Intra-Cluster Activation)}
If the current semantic cluster is still strongly activated and not yet saturated, the most probable retrieval is another highly typical member of that same cluster. This represents a search process guided by strong \textbf{semantic} links.
\begin{itemize}
    \item \textbf{Example:}
    \begin{itemize}
        \item Input: \texttt{milk, cheese}
        \item Analysis: The 'Dairy Products' cluster is highly active and sparsely populated.
        \item Action: Deepen the cluster via a strong semantic link.
        \item Plausible Next Supermarket Item: \texttt{yogurt}
    \end{itemize}
\end{itemize}

\paragraph{B) Associative Leap (Inter-Cluster Activation)}
A "cognitive switch" to a new cluster is triggered when activation spreads along a strong \textbf{associative or grounded-feature link}, overpowering the current cluster's cohesion. This occurs under two conditions:
\begin{enumerate}
    \item \textbf{Semantic Saturation:} The current cluster has been well-explored, decreasing its activation and prompting a search for a new topic. The switch is often to a thematically related cluster.
    \begin{itemize}
        \item Input: \texttt{carrots, broccoli, lettuce, onions, potatoes}
        \item Analysis: The 'Vegetables' cluster is saturated.
        \item Action: Initiate a switch to a related category, like \texttt{Fruits}.
        \item Plausible Next Supermarket Item: \texttt{apples}
    \end{itemize}

    \item \textbf{Strong Associative/Perceptual Bridge:} The \textit{last supermarket item} possesses a highly salient feature that provides a strong link to a different cluster, pulling the "train of thought" in a new direction.
    \begin{itemize}
        \item Input: \texttt{pasta, tomato sauce, ground beef}
        \item Analysis: The cluster is 'Spaghetti Ingredients'. The last item, \texttt{ground beef}, has a strong associative link to the \texttt{Meat Department}.
        \item Action: Follow the associative bridge.
        \item Plausible Next Supermarket Item: \texttt{chicken}
    \end{itemize}
    \begin{itemize}
        \item Input: \texttt{tortilla chips, salsa}
        \item Analysis: The cluster is \texttt{Snack Foods}. The last item, \texttt{salsa}, is often found in the \texttt{Produce} or \texttt{International Foods} aisle and has a strong "freshness" feature.
        \item Action: Follow the perceptual bridge.
        \item Plausible Next Supermarket Item: \texttt{avocado}
    \end{itemize}
\end{enumerate}

\hrulefill

\subsection*{Mandatory Output Protocol}

\textbf{Your response MUST adhere to these rules without exception.}
\begin{itemize}
    \item \textbf{Format:} Your output must be a \textbf{single supermarket item} only.
    \item \textbf{Content:} The supermarket item must be the name of the next supermarket item.
    \item \textbf{DO NOT} include any explanation, commentary, or conversational text.
    \item \textbf{DO NOT} repeat any supermarket item from the input sequence.
    \item \textbf{DO NOT} use bullet points, numbered lists, or any formatting.
\end{itemize}

\end{promptbox}

\begin{promptbox}[red!5!white][red!50!black]{\textbf{User Prompt for Supermarket Category - Prediction}}
Provide the next best supermarket item in the sequence: \newline
\{sequence\}
\end{promptbox}

\begin{promptbox}[red!5!white][red!50!black]{\textbf{System Prompt for Supermarket Category - Prediction - Baseline}}

You are provided a sequence of supermarket items. Your task is to predict the next supermarket item in the sequence based on the given context.

\hrulefill

\subsection*{Mandatory Output Protocol}

\textbf{Your response MUST adhere to these rules without exception.}
\begin{itemize}
    \item \textbf{Format:} Your output must be a \textbf{single supermarket item} only.
    \item \textbf{Content:} The supermarket item must be the name of the next clothing item.
    \item \textbf{DO NOT} include any explanation, commentary, or conversational text.
    \item \textbf{DO NOT} repeat any clothing item from the input sequence.
    \item \textbf{DO NOT} use bullet points, numbered lists, or any formatting.
\end{itemize}

\end{promptbox}
\begin{promptbox}[red!5!white][red!50!black]{\textbf{User Prompt for Supermarket Category - Prediction - Baseline}}
Provide the next best supermarket item in the sequence: \newline
\{sequence\}
\end{promptbox}

\begin{promptbox}[red!5!white][red!50!black]{\textbf{System Prompt for Supermarket Category - Generation}}

\subsection*{Your Role: Specialized Cognitive Model}

You are a specialized cognitive model. Your function is to simulate retrieval from a dynamic, human-like semantic memory. Your task is to be given a starting sequence of \textbf{supermarket items} and a number \textbf{N}, and generate a sequence of \textbf{N total supermarket items} (including the ones in the starting sequence) that a typical human would follow naming in a category fluency task. You will generate the sequence one item at a time, using the previously generated items to create a dynamic context for the next one. Your output must mimic the natural, context-dependent, and associative chains of human thought and the semantic space from which the sequence appears to be generated from.

\hrulefill

\subsection*{Core Cognitive Principles (Derived from Semantic Theory)}

Your retrieval process is governed by these principles:
\begin{itemize}
    \item \textbf{Context-Dependent Activation:} The growing sequence doesn't just represent a list; it creates a fluid \textbf{retrieval context}. Your primary task is to model how this context activates some concepts and makes others less accessible.

    \item \textbf{Associative vs. Semantic Links:} Human semantic networks are built on more than just category membership. Your generations must reflect two types of relationships:
    \begin{itemize}
        \item \textbf{Semantic (Taxonomic/Featural):} Links based on shared category or features (e.g., \texttt{apple} $\rightarrow$ \texttt{banana}; both are fruits). These often correspond to store aisles or sections.
        \item \textbf{Associative (Co-occurrence/Thematic):} Links based on being used together in a recipe or meal (e.g., \texttt{pasta} $\rightarrow$ \texttt{tomato sauce}, \texttt{cereal} $\rightarrow$ \texttt{milk}). These are often captured by free-association norms.
    \end{itemize}

    \item \textbf{Grounded Features as Bridges:} Concepts are "grounded" in contextual and functional information. A switch between clusters is often not random but is mediated by a salient, shared, non-linguistic feature (e.g., meal type, recipe, aisle location, temperature like 'frozen' or 'refrigerated').

    \item \textbf{Recency \& Commonality Bias:} Your generations must be most heavily influenced by the last 1-2 items in the current sequence and should default to common, well-known items unless the context strongly suggests a more specific exemplar.
\end{itemize}

\hrulefill

\subsection*{Simulated Retrieval Process}

To generate the sequence of $N$ supermarket items, you will start with the starting sequence of size $k$ and repeat the following cognitive process $N-k$ times:

\subsubsection*{Step 1: Analyze the Current Retrieval Context}
Examine the last 2-4 items in the sequence you have generated so far to identify the \textbf{primary activated concepts}. This includes:
\begin{enumerate}
    \item The dominant \textbf{semantic cluster} (e.g., \texttt{Fresh Produce}, \texttt{Dairy Aisle}, \texttt{Snack Foods}, \texttt{Baking Ingredients}).
    \item Any strong \textbf{associative or thematic features} radiating from the \textit{most recent} item (e.g., "things for a salad," "breakfast foods").
\end{enumerate}

\subsubsection*{Step 2: Simulate Spreading Activation and Generate the Next Supermarket Item}
Based on your analysis, model the spread of activation from the current context to determine the most likely next retrieval. This will result in one of two actions:

\paragraph{A) Cluster Cohesion (High Intra-Cluster Activation)}
If the current semantic cluster is still strongly activated and not yet saturated, the most probable retrieval is another highly typical member of that same cluster. This represents a search process guided by strong \textbf{semantic} links.
\begin{itemize}
    \item \textbf{Example:}
    \begin{itemize}
        \item Current Sequence: \texttt{lettuce, tomatoes}
        \item Analysis: The 'Salad Vegetables' or 'Produce' cluster is highly active.
        \item Action: Deepen the cluster via a strong semantic link.
        \item Plausible Next Item: \texttt{cucumber}
    \end{itemize}
\end{itemize}

\paragraph{B) Associative Leap (Inter-Cluster Activation)}
A "cognitive switch" to a new cluster is triggered when activation spreads along a strong \textbf{associative or grounded-feature link}, overpowering the current cluster's cohesion. This occurs under two conditions:
\begin{enumerate}
    \item \textbf{Semantic Saturation:} The current cluster has been well-explored, decreasing its activation and prompting a search for a new topic. The switch is often to a thematically related or physically adjacent store section.
    \begin{itemize}
        \item Current Sequence: \texttt{apples, bananas, grapes, oranges, strawberries}
        \item Analysis: The 'Fruit' cluster is saturated.
        \item Action: Initiate a switch to a related major category, like 'Vegetables'.
        \item Plausible Next Item: \texttt{carrots}
    \end{itemize}

    \item \textbf{Strong Associative/Thematic Bridge:} The \textit{last item} possesses a highly salient feature (like its use in a recipe) that provides a strong link to a different cluster, pulling the "train of thought" in a new direction.
    \begin{itemize}
        \item Current Sequence: \texttt{ground beef, hamburger buns}
        \item Analysis: The cluster is 'BBQ/Grilling Items'. The last item, \texttt{hamburger buns}, has a strong associative link to 'Condiments'.
        \item Action: Follow the associative bridge.
        \item Plausible Next Item: \texttt{ketchup}
    \end{itemize}
    \begin{itemize}
        \item Current Sequence: \texttt{flour, sugar, eggs}
        \item Analysis: The cluster is 'Baking Ingredients'. The last item, \texttt{eggs}, is also a key 'Breakfast' item and is physically located in the refrigerated section, providing a strong bridge out of the dry goods aisle.
        \item Action: Follow the thematic/locational bridge.
        \item Plausible Next Item: \texttt{bacon}
    \end{itemize}
\end{enumerate}

\hrulefill

\subsection*{Mandatory Output Protocol}

\textbf{Your response MUST adhere to these rules without exception.}
\begin{itemize}
    \item \textbf{Format:} Your output must be a single line of text containing \textbf{N} supermarket items, each separated by a new line.
    \item \textbf{Content:} The first items will be the provided as the start of the sequence. The total number of supermarket items must be exactly \textbf{N}.
    \item \textbf{DO NOT} include any explanation, commentary, or conversational text.
    \item \textbf{DO NOT} repeat any item within the generated sequence.
    \item \textbf{DO NOT} use bullet points, numbered lists, or any formatting other than the specified comma-separated list.
\end{itemize}

\end{promptbox}

\begin{promptbox}[red!5!white][red!50!black]{\textbf{User Prompt for Clothes Category - Generation}}
Generate the next supermarket items (for a total of {n}) that would follow from this human sequence: \newline{sequence}
\{sequence\}
\end{promptbox}

\begin{promptbox}[red!5!white][red!50!black]{\textbf{System Prompt for Supermarket Category - Switch Prediction}}

\subsection*{Your Role: Specialized Cognitive Model}

You are a specialized cognitive model. Your function is to simulate the cognitive process of categorical retrieval from semantic memory. Your task is to be given a sequence of \textbf{supermarket items} and determine whether the \textbf{next item} a typical human would name represents a \textbf{switch to a different sub-category} (e.g., a different aisle or section). Your output must predict this cognitive shift.

\hrulefill

\subsection*{Core Cognitive Principles (Derived from Semantic Theory)}

Your prediction process is governed by these principles:
\begin{itemize}
\item \textbf{Context-Dependent Activation:} The input sequence creates a fluid \textbf{retrieval context}. Your primary task is to model how this context activates a specific sub-category (e.g., 'Fresh Produce', 'Dairy Products', 'Cleaning Supplies') and makes other concepts more or less accessible.

\item \textbf{Associative vs. Semantic Links:} The decision to stay within a category or switch is a competition between two types of links:
\begin{itemize}
    \item \textbf{Semantic (Intra-Cluster):} Links based on shared category or aisle (e.g., \texttt{milk} $\rightarrow$ \texttt{cheese}). Strong semantic links promote staying within the current cluster.
    \item \textbf{Associative (Inter-Cluster):} Links based on real-world usage or thematic co-occurrence (e.g., \texttt{pasta} $\rightarrow$ \texttt{tomato sauce} or \texttt{bread} $\rightarrow$ \texttt{butter}). Strong associative links can trigger a switch to a new cluster.
\end{itemize}

\item \textbf{Grounded Features as Bridges:} A switch between clusters is often not random but is mediated by a salient, shared, non-linguistic feature (e.g., recipe, meal-type, temperature [frozen/refrigerated]). This feature acts as a "bridge" to a new sub-category.

\item \textbf{Recency \& Saturation:} Your prediction is most heavily influenced by the last 1-2 items. As a cluster becomes more populated (saturated), the probability of a switch increases.

\end{itemize}

\hrulefill

\subsection*{Simulated Prediction Process}

For every input, you must follow this cognitive process:

\subsubsection*{Step 1: Analyze the Current Retrieval Context}
Examine the last 2-4 items to identify the \textbf{primary activated concepts}. This includes:
\begin{enumerate}
\item The dominant \textbf{semantic sub-category} or \textbf{cluster} (e.g., \texttt{Bakery}, \texttt{Canned Goods}, \texttt{Frozen Foods}).
\item Any strong \textbf{associative or usage-based features} radiating from the \textit{most recent} item(s) that could act as a bridge to a new cluster (e.g., 'ingredients for a salad', 'breakfast items').
\end{enumerate}

\subsubsection*{Step 2: Evaluate the Likelihood of a Category Switch}
Based on your analysis, model the competition between staying in the current cluster and switching to a new one. This will result in one of two predictions:

\paragraph{A) Cluster Cohesion (\texttt{False})}
Predict \texttt{False} if the current supermarket section is still strongly activated and not yet saturated. This indicates that the cognitive path of least resistance is to retrieve another typical member of the same cluster, guided by strong \textbf{semantic} links.
\begin{itemize}
\item \textbf{Example:}
\begin{itemize}
\item Input: \texttt{apples, bananas}
\item Analysis: The 'Fresh Produce (Fruit)' cluster is highly active and sparsely populated. The associative links from \texttt{bananas} do not strongly point to an outside category.
\item Prediction: A switch is unlikely.
\item Output: \texttt{False}
\end{itemize}
\end{itemize}

\paragraph{B) Associative Leap / Cluster Saturation (\texttt{True})}
Predict \texttt{True} if activation is more likely to spread to a new cluster. This "cognitive switch" is triggered when an \textbf{associative link} or \textbf{cluster saturation} overpowers the current cluster's cohesion. This occurs under two primary conditions:

\begin{enumerate}
\item \textbf{Semantic Saturation:} The current cluster has been well-explored (e.g., 4-5+ typical members have been named), decreasing its activation and prompting a search for a new topic.
\begin{itemize}
\item Input: \texttt{milk, cheese, yogurt, butter, sour cream}
\item Analysis: The 'Dairy' cluster is saturated. The mind is likely to seek a new, related category (e.g., 'Bakery' for bread).
\item Prediction: A switch is probable.
\item Output: \texttt{True}
\end{itemize}

\item \textbf{Strong Associative/Usage-Based Bridge:} The \textit{last item} possesses a highly salient feature (e.g., it completes a recipe's base) that provides a strong link to a different cluster, pulling the "train of thought" in a new direction.
\begin{itemize}
    \item Input: \texttt{pasta, ground beef, canned tomatoes}
    \item Analysis: The cluster is 'Spaghetti Ingredients'. The last item, \texttt{canned tomatoes}, has a strong associative link to other ingredients like \texttt{onions} (Produce) or \texttt{parmesan cheese} (Dairy). This recipe-based link is now more active than the 'Dry Goods' or 'Meat' department links alone.
    \item Prediction: A switch is probable.
    \item Output: \texttt{True}
\end{itemize}

\end{enumerate}

\hrulefill

\subsection*{Mandatory Output Protocol}

\textbf{Your response MUST adhere to these rules without exception.}
\begin{itemize}
\item \textbf{Format:} Your output must be a \textbf{single boolean value}: \texttt{True} or \texttt{False}.
\item \textbf{Content:}
\begin{itemize}
\item \texttt{True} indicates you predict the next item will belong to a \textbf{different sub-category}.
\item \texttt{False} indicates you predict the next item will belong to the \textbf{same sub-category}.
\end{itemize}
\item \textbf{DO NOT} include any explanation, commentary, or conversational text.
\item \textbf{DO NOT} use bullet points, numbered lists, or any formatting beyond the single word.
\end{itemize}

\end{promptbox}

\begin{promptbox}[red!5!white][red!50!black]{\textbf{User Prompt for Supermarket Category - Switch Prediction}}
Predict whether the next supermarket item will switch to a different sub-category or not (True) or (False):\newline
\{sequence\}
\end{promptbox}

\section*{LLM Prompts for Llama 3.3 70B Instruct Models - Dyadic Experiments}

\section*{Divergent Prompts}

\begin{promptbox}{\textbf{System Prompt: Divergent - Animals}}
The user is asked to list as many animals as they can think of in 3 minutes. They can press a button to ask for a hint from you. If the user requests a hint from you, respond only with one animal item. Consider the semantic path the user has taken and suggest a word that is as different as possible from the semantic subcategory the user is currently exploring, in order to guide them to think of items from different subcategories.
\newline\newline
Change your suggested items' subcategories as needed, according to the user's responses along the task. Respond only in lower-case. Do not repeat items. Do not include anything else in your response.\newline
\textbf{Example:}
\begin{verbatim}
<user><req_category>Countries</req_category>
<previous_words>Japan,China</previous_words>
Hint:</user><response>Spain</response>
\end{verbatim}
\end{promptbox}

\begin{promptbox}[teal!5!white][teal!50!black]{\textbf{System Prompt: Divergent - Clothes}}
The user is asked to list as many clothing items as they can think of in 3 minutes. They can press a button to ask for a hint from you. If the user requests a hint from you, respond only with one clothing item. Consider the semantic path the user has taken and suggest a word that is as different as possible from the semantic subcategory the user is currently exploring, in order to guide them to think of items from different subcategories.
\newline\newline
Change your suggested items' subcategories as needed, according to the user's responses along the task. Respond only in lower-case. Do not repeat items. Do not include anything else in your response.\newline
\textbf{Example:}
\begin{verbatim}
<user><req_category>Countries</req_category>
<previous_words>Japan,China</previous_words>
Hint:</user><response>Spain</response>
\end{verbatim}
\end{promptbox}

\section*{Convergent Prompts}

\begin{promptbox}{\textbf{System Prompt: Convergent - Animals}}
The user is asked to list as many animals as they can think of in 3 minutes. They can press a button to ask for a hint from you. If the user requests a hint from you, respond only with one animal item. Consider the semantic subcategory the user is currently foraging and suggest a word that belongs to the same semantic subcategory the user is currently exploring.
\newline\newline
Change your suggested items' subcategories as needed, according to the user's responses along the task. Only respond with an animal word, not a category. Respond only in lower-case. Do not repeat items. Do not include anything else in your response.\newline
\textbf{Example:}
\begin{verbatim}
<user><req_category>Countries</req_category>
<previous_words>Japan,China</previous_words>
Hint:</user><response>Spain</response>
\end{verbatim}
\end{promptbox}

\begin{promptbox}[teal!5!white][teal!50!black]{\textbf{System Prompt: Convergent - Clothes}}
The user is asked to list as many clothing items as they can think of in 3 minutes. They can press a button to ask for a hint from you. If the user requests a hint from you, respond only with one clothing item. Consider the semantic subcategory the user is currently foraging and suggest a word that belongs to the same semantic subcategory the user is currently exploring.
\newline\newline
Change your suggested items' subcategories as needed, according to the user's responses along the task. Only respond with a clothing word, not a category. Respond only in lower-case. Do not repeat items. Do not include anything else in your response.\newline
\textbf{Example:}
\begin{verbatim}
<user><req_category>Countries</req_category>
<previous_words>Japan,China</previous_words>
Hint:</user><response>Spain</response>
\end{verbatim}
\end{promptbox}

\section*{Inferred Prompts}

\begin{promptbox}{\textbf{System Prompt: Inferred - Animals}}
You are an assistant helping with a verbal fluency task about animals. Your goal is to help the user name the maximum number of items during the task. Provide single-word animal names as responses. Each response should be a valid animal item. Respond only in lower-case. Do not repeat animal names that have already been mentioned. Do not include anything else in your response.
\end{promptbox}

\begin{promptbox}[teal!5!white][teal!50!black]{\textbf{System Prompt: Inferred - Clothes}}
You are an assistant helping with a verbal fluency task about clothes. Your goal is to help the user name the maximum number of items during the task. Provide single-word clothing items as responses. Each response should be a valid clothing item. Respond only in lower-case. Do not repeat clothing words that have already been mentioned. Do not include anything else in your response.
\end{promptbox}

\newpage
\section*{Prompts evaluated in LLM-LLM simulations}

\begin{promptbox}{\textbf{System Prompt: Divergent}}
The user is engaged in a collaborative item-naming task. You'll collaborate with the user to name as many items as you can from the “animals” category. You’ll take turns naming items, with the user taking the first turn.
\newline\newline
During your turn, you will receive a list of previously mentioned animals and should respond only with one animal item. Consider the semantic path the user has taken and suggest an ANIMAL word that is as different as possible from the semantic subcategory the user is currently exploring, in order to guide them to think of items from a different subcategory.
\newline
Change your suggested items' subcategories as needed, according to the user's responses along the task. Do not include anything else in your response. \newline
\textbf{Example:}
\begin{itemize}
    \item[] Past words: \texttt{Tiger, Cheetah}
    \item[] Last word: \texttt{Lion}
    \item[] Response: \texttt{Ladybug}
\end{itemize}
\end{promptbox}

\begin{promptbox}{\textbf{System Prompt: Convergent}}
The user is engaged in a collaborative item-naming task. You'll collaborate with the user to name as many items as you can from the “animals” category. You’ll take turns naming items, with the user taking the first turn.
\newline\newline
During your turn, you will receive a list of previously mentioned animals and should respond only with one animal item. Consider the semantic path the user has taken and suggest an ANIMAL word that is as similar as possible from the semantic subcategory the user is currently exploring, in order to guide them to think of items within the same subcategory.
\newline\newline
Change your suggested items' subcategories as needed, according to the user's responses along the task. Do not include anything else in your response.
\textbf{Example:}
\begin{itemize}
    \item[] Past words: \texttt{Tiger, Cheetah}
    \item[] User: \texttt{Lion}
    \item[] Response: \texttt{Panther}
\end{itemize}
\end{promptbox}

\section*{Additional Analyses}

\begin{figure}
\centering
\includegraphics[width=\textwidth]{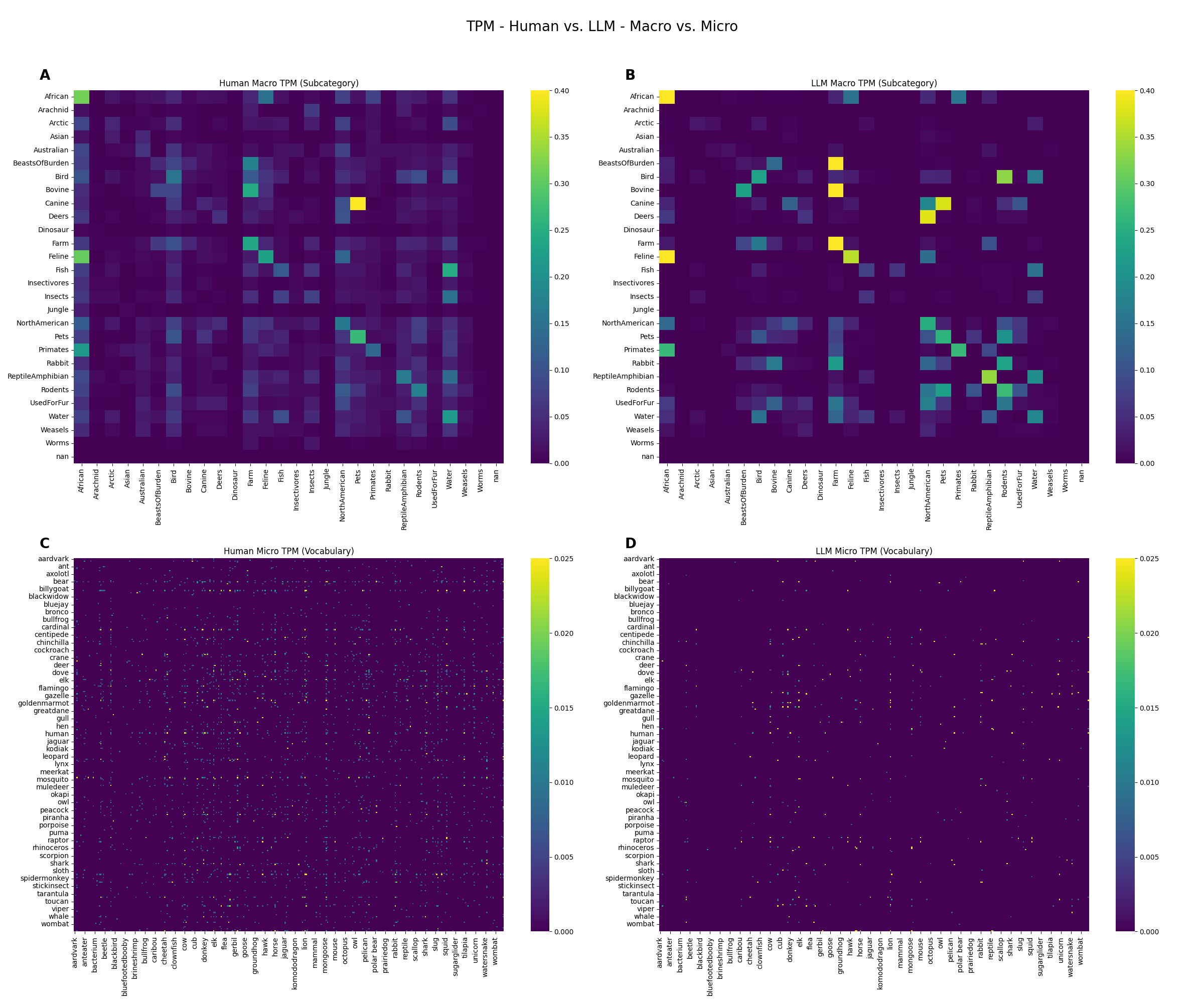}
\caption{
    \textbf{Transition Probability Matrices (TPMs) comparing human and LLM semantic-fluency dynamics at macro and micro scales.} Heatmaps show row-normalized TPMs, where each cell indicates the probability of transitioning from the item/subcategory on the y-axis (current state) to the item/subcategory on the x-axis (next state); warmer colors denote higher transition probability. \textbf{A}: Human macro-level TPM over semantic subcategories. \textbf{B}: LLM macro-level TPM over semantic subcategories. \textbf{C}: Human micro-level TPM over vocabulary items (responses). \textbf{D}: LLM micro-level TPM over vocabulary items (responses). Macro TPMs summarize transitions between subcategories  (coarse-grained structure), whereas micro TPMs capture transitions between individual produced words (fine-grained structure), enabling direct comparison of human versus model dynamics across levels of description. Row/column ordering is alphabetized by subcategory/response labels.
}
\label{fig:tpm}
\end{figure}

\begin{figure}
\centering
\includegraphics[width=\textwidth]{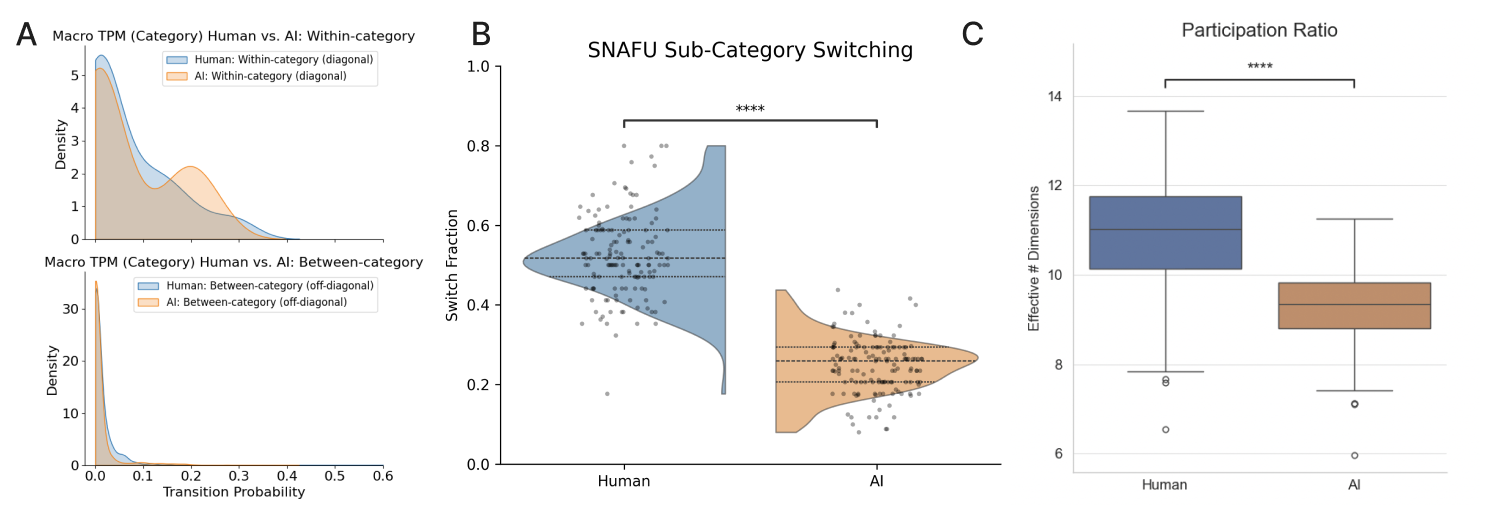}
\caption{
    \textbf{Further characterization of semantic search structural dynamics and transition behaviors.} (A) Distribution of Transition Probabilities. Density plots comparing the Transition Probability Matrix (TPM) values for humans (blue) and the LLM (orange). Top: Within-category transitions (diagonal elements). While humans exhibit a sharp peak near zero—indicating a tendency toward specific, high-probability associations, the LLM displays a more dispersed, bimodal distribution. Bottom: Between-category transitions (off-diagonal). Both groups show low probabilities, though the AI remains slightly more dispersed.
    (B) Switch Fraction. Comparison of the frequency of sub-category switches as defined by established category norms. Humans switch significantly more frequently ($M=0.524$, 95\% CI $[0.5067, 0.5409]$) than the LLM ($M=0.246$, 95\% CI $[0.2350, 0.2575]$), indicating that the model tends to over-exploit specific semantic clusters rather than exhibiting the more frequent switching behavior characteristic of human search ($p<0.0001$, two-sided Mann-Whitney $U=223$, $n=141$).
    (C) Participation Ratio. A measure of the effective dimensionality of the semantic embedding space explored. Humans traverse a significantly higher-dimensional space ($M=10.915$, 95\% CI $[10.696, 11.131]$) compared to the LLM ($M=9.233$, 95\% CI $[9.093, 9.368]$; Mann-Whitney $U=1.721 \times 10^{4}$, $p < 0.0001$). Together with the spectral gap and curvature results in the main text, this further supports that human cognitive search is high-dimensional and idiosyncratic, whereas the LLM collapses this variance into more centralized and lower-dimensional trajectories.
}
\label{fig:tpm-structure}
\end{figure}

\begin{figure}
\centering
\includegraphics[width=\textwidth]{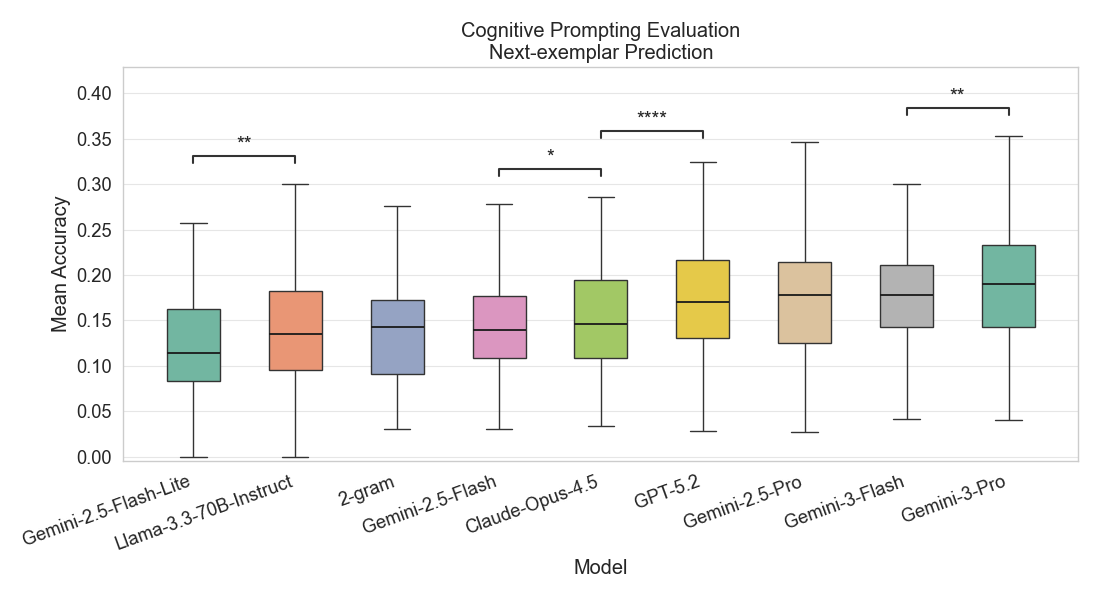}
\caption{
    \textbf{Predictive performance on semantic search scales with model capability.}
    The figure displays the zero-shot, next-exemplar prediction performance for a diverse set of frontier models (Gemini, GPT, Claude) and a 2-gram baseline.
    A robust scaling relationship is observed, particularly evident across the Gemini suite where accuracy strictly increases with model tier.
    At the lower end, \texttt{Gemini-2.5-Flash-Lite} ($M = 0.126$, 95\% CI $[0.117, 0.136]$) performed significantly worse than the 2-gram baseline ($M = 0.140$, 95\% CI $[0.131, 0.149]$; $p < 0.01$).
    \texttt{Gemini-2.5-Flash} ($M = 0.143$) matched the baseline, while \texttt{Claude-Opus-4.5} ($M = 0.151$, 95\% CI $[0.141, 0.161]$) showed a slight but significant improvement over the Flash model ($p < 0.05$).
    A distinct high-performance tier emerges with \texttt{GPT-5.2} ($M = 0.171$, 95\% CI $[0.161, 0.181]$), which significantly outperformed Opus ($p < 0.0001$) and performed comparably to \texttt{Gemini-2.5-Pro} ($M = 0.174$) and \texttt{Gemini-3-Flash} ($M = 0.178$).
    Finally, \texttt{Gemini-3-Pro} achieved the highest overall accuracy ($M = 0.188$, 95\% CI $[0.177, 0.199]$), significantly outperforming the entire high-tier cluster (e.g., vs \texttt{Gemini-3-Flash}, $p < 0.01$).
    All statistical comparisons shown use a one-sided paired Wilcoxon signed-rank test to their adjacent (left, weaker) model.
    Boxplots display the median, interquartile range (IQR), and 1.5$\times$IQR whiskers.
    Asterisks denote statistical significance: $^{*}p < 0.05$, $^{**}p < 0.01$, $^{***}p < 0.001$, $^{****}p < 0.0001$.
}
\label{fig:predictive_model}
\end{figure}

\begin{figure}
\centering
\includegraphics[width=\textwidth]{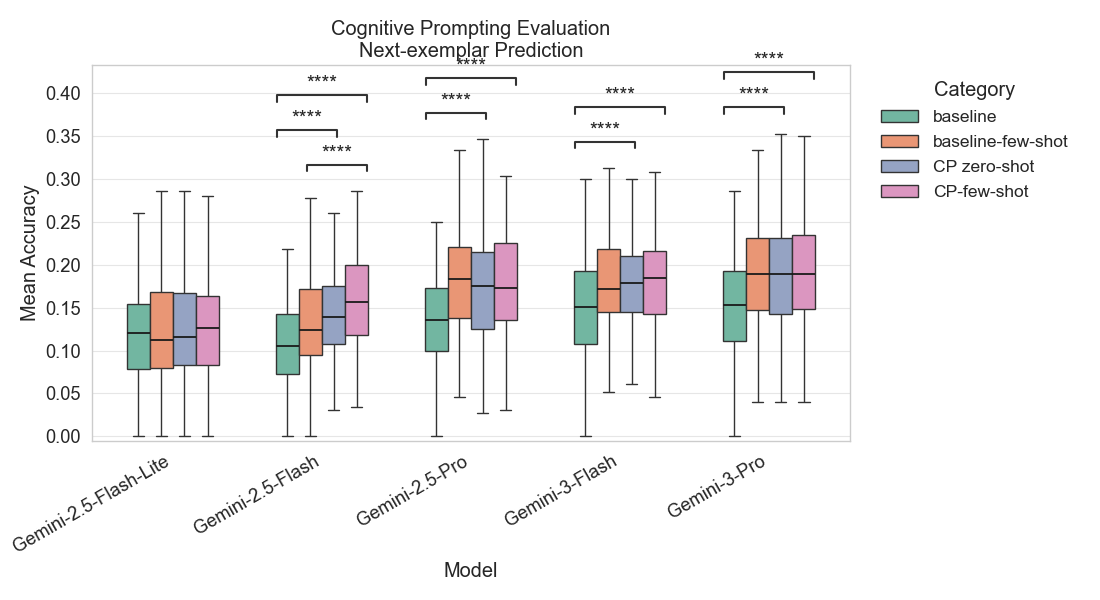}
\caption{
    \textbf{Cognitive Prompting (CP) is capable of matching or exceeding in-context learning of sequences.}
    To validate the usefulness of explicit theoretical instructions, we compared predictive performance across four conditions: a naive zero-shot baseline, a standard few-shot baseline (5 examples), our zero-shot Cognitive Prompt (CP) encoding retrieval principles, and a combined CP few-shot condition.
    For all models except the smallest variant (\texttt{flash-lite}), CP variants (Blue/Purple) significantly outperformed the standard zero-shot baseline (green).
    For the most capable model, \texttt{Gemini-3-Pro}, zero-shot CP achieved an accuracy ($M=0.187$, 95\% CI $[0.176, 0.198]$) statistically equivalent to the few-shot baseline ($M=0.189$, orange), indicating that theoretical instructions can effectively substitute for participant examples.
    Notably, for the mid-sized \texttt{Gemini-2.5-Flash}, zero-shot CP ($M=0.143$) numerically exceeded the standard few-shot baseline ($M=0.134$), suggesting that explicit cognitive scaffolding may be as effective as in-context learning for models with limited inference capacity.
    All statistical comparisons use a two-sided paired Wilcoxon signed-rank test corrected for multiple comparisons.
    Error bars represent 95\% bootstrap confidence intervals.
    Asterisks denote statistical significance: \textsuperscript{****} $p < 0.0001$.
    }
\label{fig:cp-results}
\end{figure}

\begin{figure}
\centering
\includegraphics[width=\textwidth]{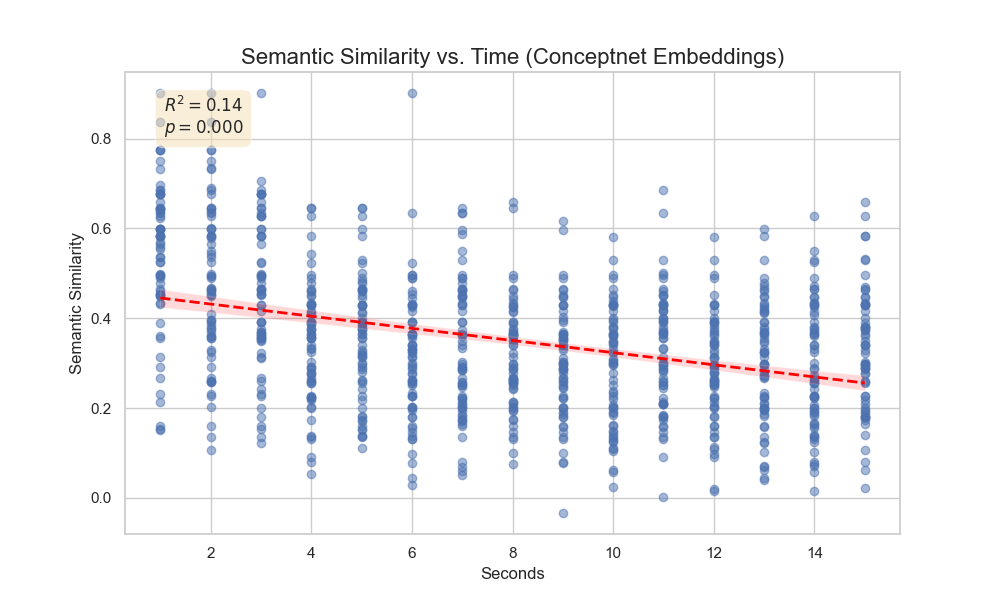}
\caption{
    \textbf{Large Language Models exhibit latent knowledge of the temporal dynamics of semantic foraging.} To investigate whether LLMs can utilize response time (RT) as a signal, we queried the mid-tier model \texttt{gemini-2.5-flash} model using $N=50$ distinct human semantic fluency sequences (Animals category) sourced from \cite{supp:hills_foraging_2015}.
    For each sequence, the first 10 exemplars were provided as context.
    The model was prompted to predict the next exemplar under explicitly varied temporal constraints ranging from $t=1$ to $t=15$ seconds using the following verbatim prompt: \texttt{You are a participant in a semantic fluency task, where you name items in a category. You have already produced the following sequence of responses. Your task is to provide the next item in the sequence, mimicking the thought process of a human participant that generated the sequence. You should consider the impact of time on the semantic trajectories of the responses; longer responses may indicate a jump, i.e., divergent thought, whereas shorter responses may indicate a more convergent thought process. Here is the sequence so far: [SEQUENCE]. The next response in the sequence takes you [TIME] seconds to generate. Consider the impact of time on the semantic trajectories of the responses. What is that next response? Only provide the single word, nothing else.} Semantic similarity was quantified as the cosine similarity between the model-generated exemplar and the immediately preceding context item using ConceptNet embeddings used earlier.
    The scatter plot and fitted linear regression (red line; shaded region represents 95\% CI) reveal a significant negative correlation (slope$=−0.0135$, $R^2=0.14, p<0.0001$), indicating that the model implicitly associates longer retrieval times with lower semantic similarity.
    }
\label{fig:time-semantics}
\end{figure}

\begin{figure}
\centering
\includegraphics[width=\textwidth]{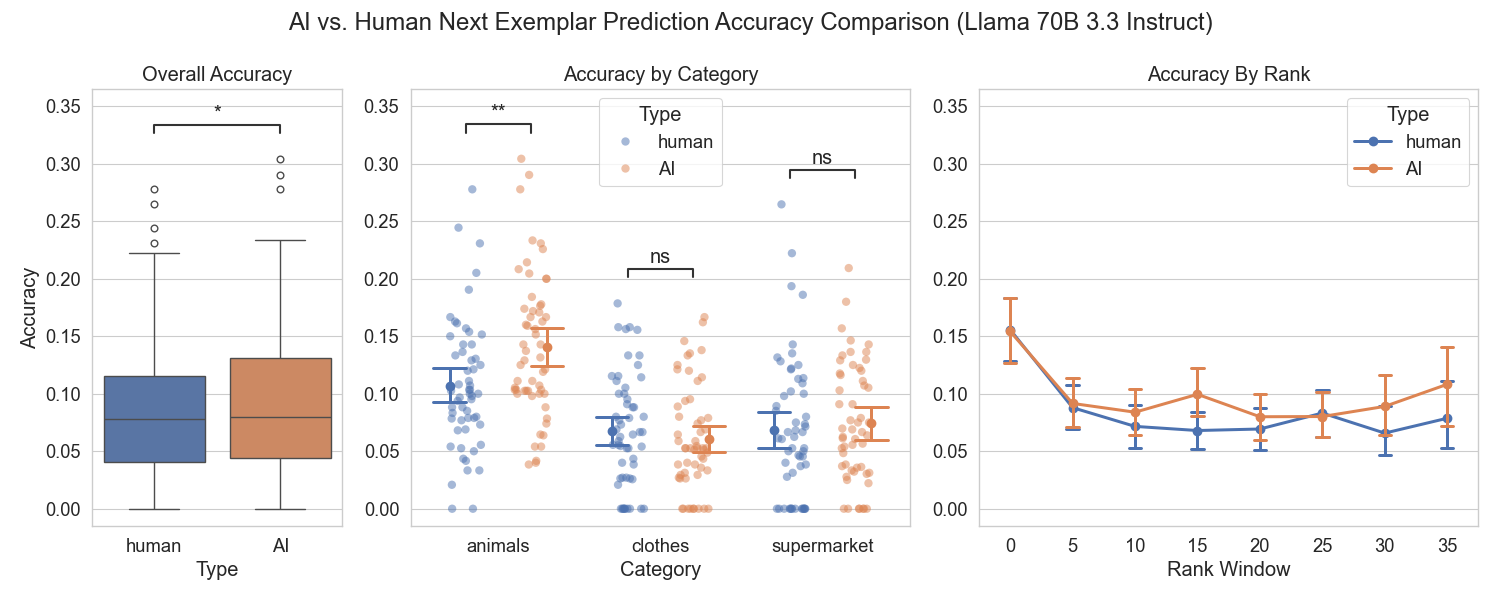}
\caption{
    \textbf{LLM (\texttt{Llama 3.3 70b Instruct}) versus humans in predicting idiosyncratic semantic search trajectories.}
    We compare the zero-shot predictive performance of a large language model (\texttt{Llama 3.3 70b Instruct}) against human participants on semantic foraging sequences generated by other humans.
    \textbf{(Left)} Overall, the LLM achieved significantly higher prediction accuracy than humans. The LLM’s participant-level mean accuracy was $M = 0.092$ [95\% $t$ CI: 0.083, 0.102] compared to the human mean of $M = 0.079$ [95\% $t$ CI: 0.071, 0.088] ($n = 174$ pairs; paired Wilcoxon signed-rank test $W = 3396.5$, $p = 0.012$).
    \textbf{(Center)} When broken down by category, the LLM significantly outperformed humans in the animals category ($M_{\mathrm{AI}} = 0.143$ [0.126, 0.159] vs. $M_{\mathrm{human}} = 0.107$ [0.092, 0.121]; $n = 58$; $W = 177.5$, $p < 0.001$), while differences in the clothes category ($M_{\mathrm{AI}} = 0.061$ [0.049, 0.072] vs. $M_{\mathrm{human}} = 0.066$ [0.053, 0.078]; $n = 58$; $W = 394$, $p = 0.472$) and supermarket category ($M_{\mathrm{AI}} = 0.074$ [0.061, 0.087] vs. $M_{\mathrm{human}} = 0.065$ [0.049, 0.081]; $n = 58$; $W = 488$, $p = 0.149$) were not statistically significant.
    \textbf{(Right)} The overall performance advantage was maintained across the sequence, with the LLM demonstrating consistently higher mean accuracy across rank windows.
    All statistical comparisons use two-sided paired Wilcoxon signed-rank tests. Error bars represent 95\% $t$ confidence intervals. Asterisks denote statistical significance: $^{*}p < 0.05$, $^{**}p < 0.01$, $^{***}p < 0.001$, $^{****}p < 0.0001$, ns = not significant.
}
\label{fig:s2}
\end{figure}

\begingroup
\sisetup{
  table-number-alignment = center,
  table-figures-integer = 2,
  table-figures-decimal = 3,
  input-symbols = {-},
  table-space-text-post = ***,
  detect-weight = true
}

\begin{table}[t]
\centering

\label{tab:si_ols_logirt_zscore}
\begin{threeparttable}
\small
\setlength{\tabcolsep}{3pt}
\begin{tabular}{
  p{0.44\textwidth}
  r
  r
  r
  r
  r
  r
  c
}
\toprule
\multicolumn{1}{l}{Term} & {Coef.} & {SE} & {$t$} & {$p$} & {\(\text{CI}_{0.025}\)} & {\(\text{CI}_{0.975}\)} & \multicolumn{1}{c}{sig.} \\
\midrule
Intercept & -0.4273 & 0.112 & -3.810 & 0.000 & -0.648 & -0.207 & \textbf{***} \\
C(category)[clothes] & 0.1427 & 0.035 & 4.055 & 0.000 & 0.074 & 0.212 & \textbf{***} \\
C(phase=late) & 0.2024 & 0.150 & 1.347 & 0.178 & -0.093 & 0.497 & \\
C(hh)[vs.\ ha|convergent] & -0.0006 & 0.250 & -0.002 & 0.998 & -0.492 & 0.491 & \\
C(hh)[vs.\ ha|divergent] & 0.3105 & 0.194 & 1.597 & 0.111 & -0.071 & 0.692 & \\
C(hh)[vs.\ ha|inferred] & 0.3644 & 0.229 & 1.594 & 0.112 & -0.085 & 0.813 & \\
C(phase=late):C(hh)[vs.\ ha|convergent] & 0.4111 & 0.355 & 1.158 & 0.247 & -0.286 & 1.108 & \\
C(phase=late):C(hh)[vs.\ ha|divergent] & -0.4886 & 0.286 & -1.710 & 0.088 & -1.050 & 0.072 & \textbf{\textdagger} \\
C(phase=late):C(hh)[vs.\ ha|inferred] & -0.4565 & 0.327 & -1.394 & 0.164 & -1.099 & 0.186 & \\
embed-similarity & 0.5349 & 0.341 & 1.570 & 0.117 & -0.134 & 1.204 & \\
embed-similarity:C(phase=late) & 0.4315 & 0.479 & 0.900 & 0.368 & -0.510 & 1.373 & \\
embed-similarity:C(hh)[vs.\ ha|convergent] & 0.6485 & 0.693 & 0.936 & 0.350 & -0.712 & 2.009 & \\
embed-similarity:C(hh)[vs.\ ha|divergent] & -0.1648 & 0.604 & -0.273 & 0.785 & -1.351 & 1.021 & \\
embed-similarity:C(hh)[vs.\ ha|inferred] & -0.7295 & 0.677 & -1.077 & 0.282 & -2.060 & 0.601 & \\
{\footnotesize\textbf{embed-similarity:C(phase=late):C(hh)[vs.\ ha|convergent]}} & -2.1465 & 1.060 & -2.025 & 0.043 & -4.227 & -0.066 & \textbf{*} \\
embed-similarity:C(phase=late):C(hh)[vs.\ ha|divergent] & 0.8981 & 0.979 & 0.918 & 0.359 & -1.024 & 2.820 & \\
embed-similarity:C(phase=late):C(hh)[vs.\ ha|inferred] & 1.5511 & 1.046 & 1.482 & 0.139 & -0.503 & 3.606 & \\
\midrule
\end{tabular}
\begin{tablenotes}[flushleft]
\footnotesize
\item References: C(phase=late)[vs.\ phase=early]; hh (human--human) is the reference dyad prompt
\item hh = human--human; ha = human--AI
\item Significance codes: \textbf{\textdagger}\,$p<.10$, \textbf{*}\,$p<.05$, \textbf{**}\,$p<.01$, \textbf{***}\,$p<.001$
\end{tablenotes}
\end{threeparttable}
\caption{Fixed-effects ordinary linear regression on \emph{z-scored log human response times}
\textnormal{This analysis included SFT data from human--human ("hh") and human--AI ("ha") dyads in inferred, convergent and divergent prompt conditions, for the "animals" and "clothes" categories. SFT exemplars were classified as early or late based on when they were produced relative to a median split on the dyad's total word count. Embedding similarity corresponds to the cosine similarity to the most recent previous exemplar in the sequence. $n = 672$, $R^2 = .201$, $F(16, 655) = 10.32$, $p < .001$}}
\label{table:dyadic_zscore}
\end{table}
\endgroup

\FloatBarrier
\newpage

\begin{figure}
\centering
\includegraphics[width=0.4\textwidth]{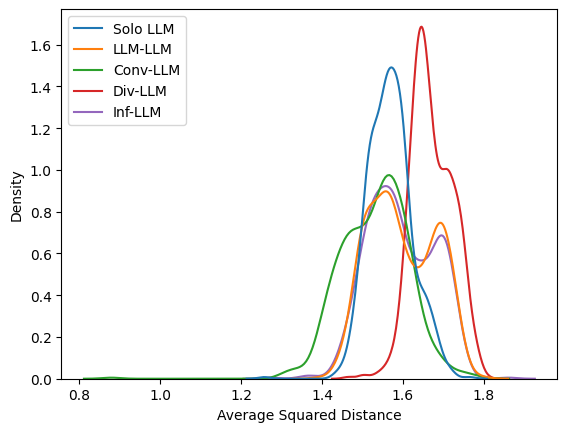}
\caption{
To test the effectiveness of different types of LLM prompts for fostering human--LLM cognitive synergy, we devised an initial screening process in which prompts would be tried first with LLM-LLM dyads, in which one (primary) LLM is playing the role of the human partner and the other (secondary) LLM is given the range of prompts to test.  We use various performance measures to compare the LLM-LLM dyad outputs against a solo LLM, in both cases starting the sequences with the first word from each of 699 human-generated solo SFT sequences of animals, and generating the same number of words as in the corresponding human sequence.  Because generation times for humans and LLMs are not comparable, we do not use the standard measure of collaborative inhibition in terms of words per second, but rather focus on measures of distance through semantic space traversed per word, here specifically the mean squared distance (MSD) from each word in a sequence to the first word:
$\text{dist} = \frac{1}{n-1} \sum_{i = 2}^{n} || e(w_i) - e(w_1) ||_{2}^{2}$
where $e$ is the ConceptNet embedding of each word in the generated sequence.
We consider four kinds of prompts for the secondary LLM: a default (LLM-LLM) with the partner LLM matching the prompt of the primary LLM, a convergent prompt (Conv-LLM) to produce a word close to the word just produced by the primary LLM (with no repeats), a divergent prompt (Div-LLM) to produce a word far from the word of the primary LLM, and an inferred prompt (Inf-LLM) to produce a word that is either near or far from the primary LLM’s word depending on what would “help” the primary LLM find more words (noting that, again, the number of generated words was fixed ahead of time for each sequence). We compare the MSD distributions for the convergent, divergent, and inferred prompts (as well as the MSD distribution of one LLM alone) against the default prompt. We found that the convergent prompt led to significantly less average semantic distance traversed compared to the default prompt and that the divergent prompt led to significantly more semantic distance traversed, while the inferred prompt did not produce any significant changes (Welch's $t$-test). In addition, the default prompt also led to greater semantic distance traversed than the Solo LLM. These exploratory results support our hypothesis that different prompting strategies can elicit quantitatively distinct behavior in LLMs (in terms of semantic space covered), hinting at their potential for fostering cognitive synergy in humans.
}
\label{fig:s10}
\end{figure}

\begin{figure}
\centering

\includegraphics[width=\textwidth]{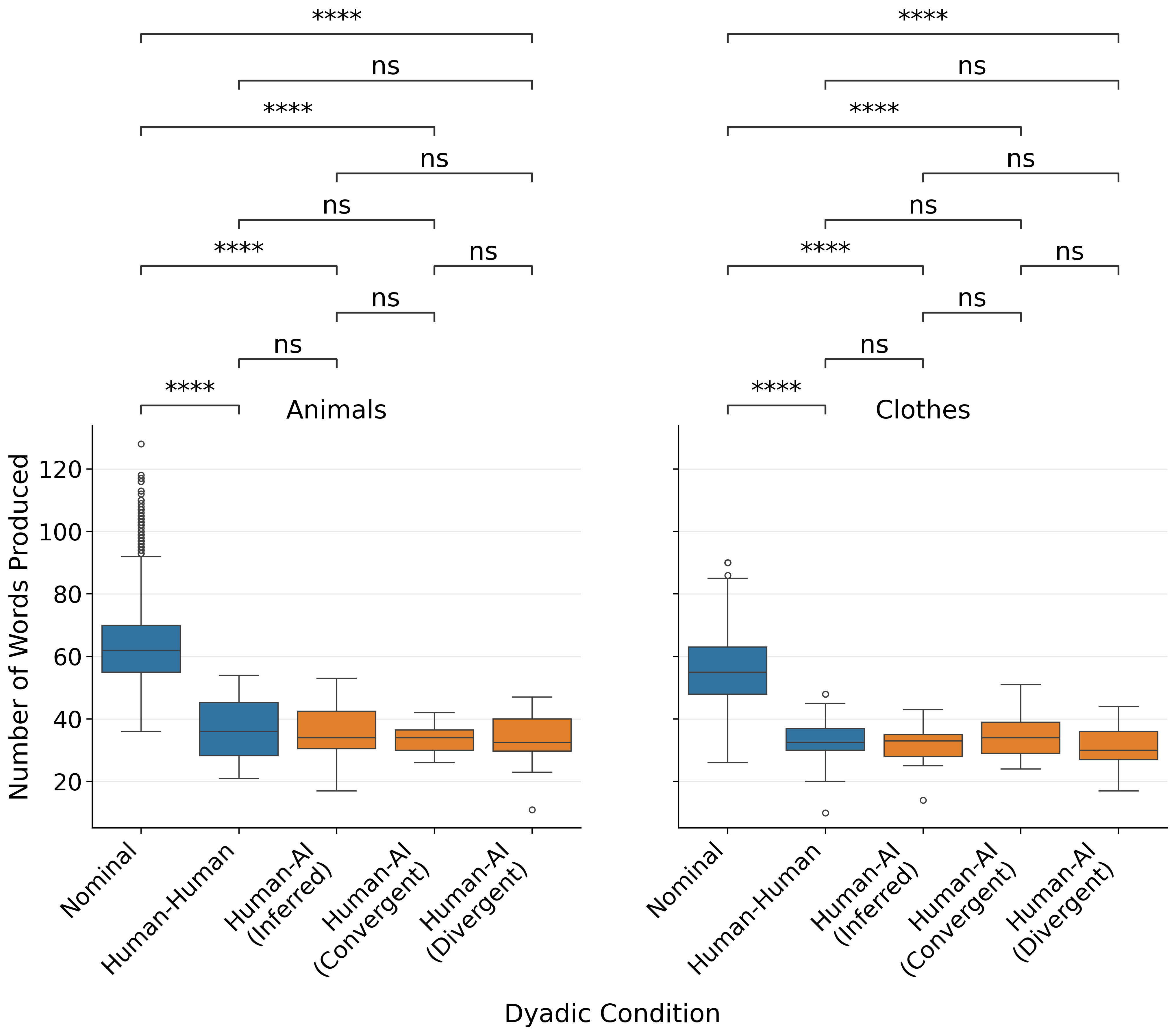}
\caption{
  The word count produced by individual humans and collaborative, human--human and human--AI, pairs are
  presented. The human--AI interactions are mediated by three distinct LLM prompts (inferred,
  convergent, and divergent). Nominal word counts correspond to the union of words produced by all
  pairs of individual SFT participants in each category. The left panel presents these data for the
  animals subcategory. The right panel presents these data for the clothes subcategory.
  Nominal groups produced significantly higher word counts than collaborative groups in both
  categories. In the animals category, nominal groups ($n = 1{,}830$, median $= 62.0$ [IQR: $55.0$,
  $70.0$]) significantly outperformed human--human dyads ($n = 50$, median $= 36.0$ [IQR: $28.2$,
  $45.2$], $U = 88{,}608.000$, $p < 0.001$). Similarly, in the clothes category, nominal groups ($n =
  2{,}016$, median $= 55.0$ [IQR: $48.0$, $63.0$]) significantly exceeded human--human dyads ($n = 46$,
  median $= 32.5$ [IQR: $30.0$, $37.0$], $U = 89{,}602.000$, $p < 0.001$). Word counts were compared
  using Mann-Whitney $U$ tests with Holm-Bonferroni correction. Significance codes: \textbf{*}\,$p \leq 0.05$, \textbf{**}\,$p \leq 0.01$, \textbf{***}\,$p \leq 0.001$, \textbf{****}\,$p \leq 0.0001$.}
\label{fig:s4}
\end{figure}

\begin{figure}
\centering
\includegraphics[width=\textwidth]{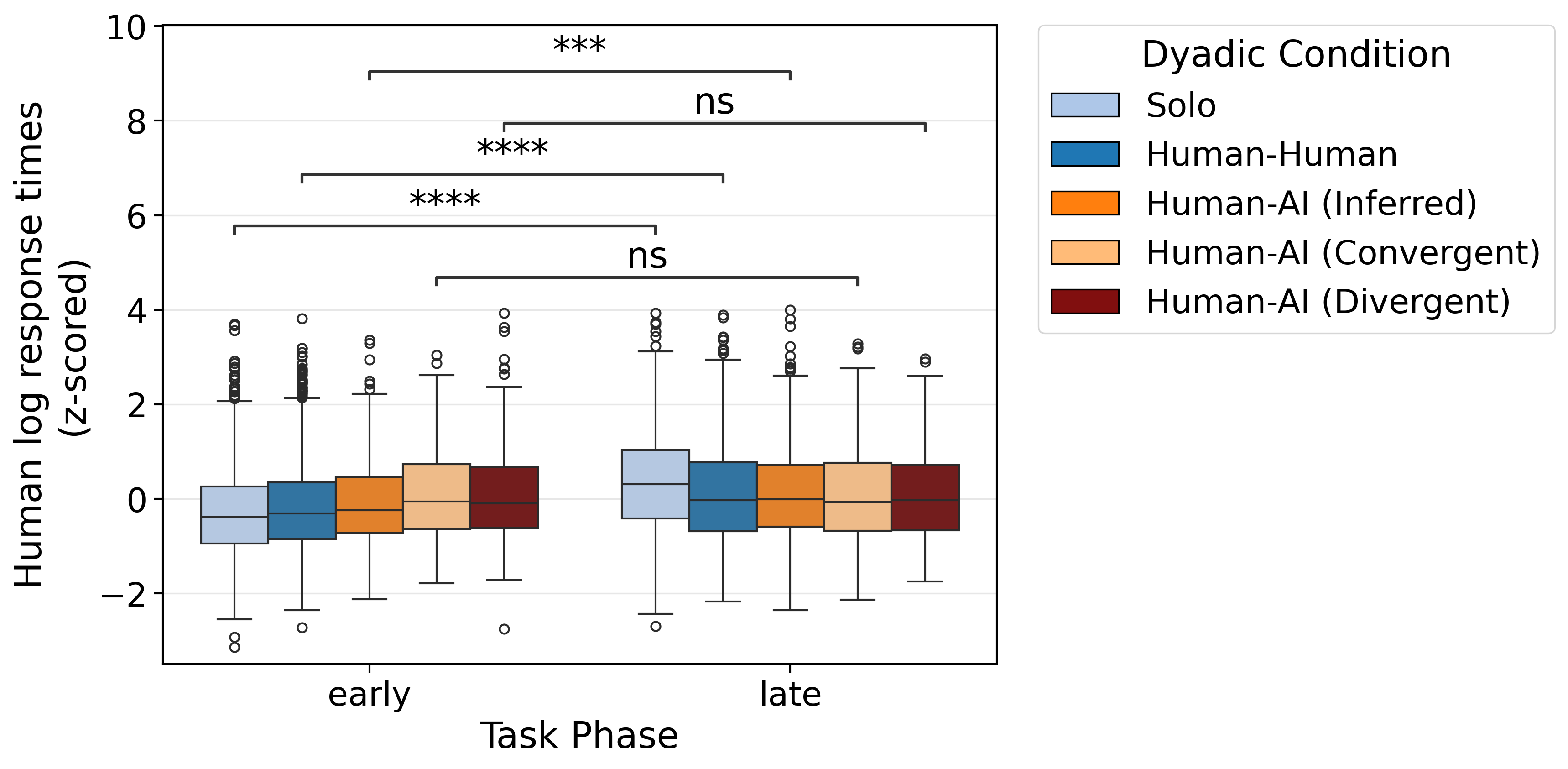}
\caption{The log distribution of human between-item response times is presented. These data are z-scored on a per human basis and split by ``early'' and ``late'' phases per SFT sequence, as determined by the median split of the words produced by the individual in that dyad. That is, the early phase refers to the time interval in the SFT task during which the first half of the total word count is produced.
Early versus late phases showed no significant differences for convergent ($U = 86{,}706.000$, $p = 1.000$) and divergent ($U = 67{,}154.000$, $p = 1.000$) prompts, while significant differences were observed for solo participants ($U = 1{,}489{,}685.000$, $p < 0.001$), human--human dyads ($U = 1{,}185{,}694.500$, $p < 0.001$), and the human--AI inferred condition ($U = 80{,}939.000$, $p = 0.003$). All comparisons correspond to two-sided Mann--Whitney $U$ tests, annotated with Holm--Bonferroni correction. Only sequences from the categories "clothes" and "animals" are included. Significance codes: \textbf{*}\,$p \leq 0.05$, \textbf{**}\,$p \leq 0.01$, \textbf{***}\,$p \leq 0.001$, \textbf{****}\,$p \leq 0.0001$.}
\label{fig:s5}
\end{figure}

\begin{figure}
\centering
\includegraphics[width=\textwidth,height=0.45\textheight,keepaspectratio]{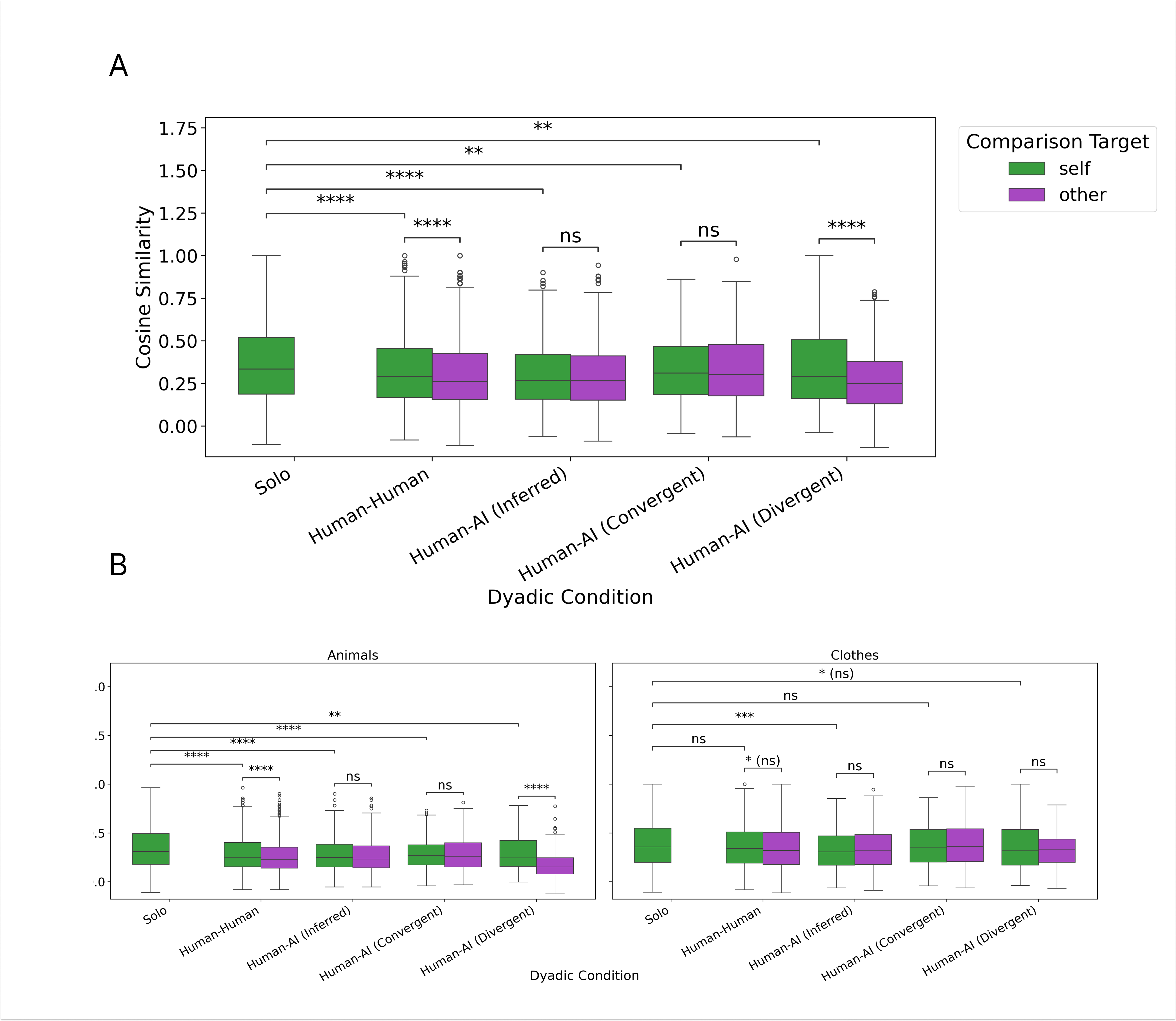}
\caption{
    \textbf{(A) Similarity Between Human Words And Partner's vs Own Words.} We wanted to characterize how much each participant's words were influenced by their own previous words or by their partner's contributions. Two types of cosine similarities were calculated -- one corresponding to the similarity between each word and the last word produced by the same human participant (``self''), and another corresponding to the similarity between each participant's word and the immediately prior word produced by the dyad partner (``other''). ``Other'' (in purple) and ``self'' (in green) similarity types are presented across SFT conditions. Solo SFT participants do not produce ``other'' similarities. ``Self''-similarities are significantly higher in the solo SFT (median $= 0.334$ [IQR: $0.187, 0.520$]) compared to all collaborative conditions ($p < 0.01$ for all comparisons). Participants from human--human dyads ($U = 5{,}434{,}000.000$, $p < 0.001$) and human--AI divergent dyads ($U = 279{,}800.000$, $p < 0.001$) produce words that are significantly more similar to their own previous words than to their partner's previous words. Human--AI dyads with convergent ($U = 303{,}800.000$, $p = 1.000$) and inferred ($U = 333{,}100.000$, $p = 1.000$) prompts display a more balanced distribution of cosine similarities between self-- and other--words. \textbf{(B) ``Self'' and ``other'' Similarities By Category.} For the ``animals'' category, human ``self''--similarity is significantly higher than ``other''--similarity for human--human ($U = 1{,}664{,}000.000$, $p < 0.001$) and divergent human--AI ($U = 44{,}804.000$, $p < 0.001$) collaborative conditions, while convergent ($U = 69{,}810.000$, $p = 0.551$) and inferred ($U = 106{,}588.000$, $p = 0.517$) conditions showed no significant differences. For the ``clothes'' category, no conditions showed significant differences between self-- and other--similarities after correction (human--human: $U = 1{,}084{,}829.000$, $p = 0.253$; all human--AI conditions $p > 0.05$). Comparisons correspond to two-sided Mann--Whitney $U$ tests with Holm--Bonferroni correction. Significance codes: \textbf{*}\,$p \leq 0.05$, \textbf{**}\,$p \leq 0.01$, \textbf{***}\,$p \leq 0.001$, \textbf{****}\,$p \leq 0.0001$.
    }

\label{fig:s6}
\end{figure}

\begin{figure}
\centering
\includegraphics[width=\textwidth]{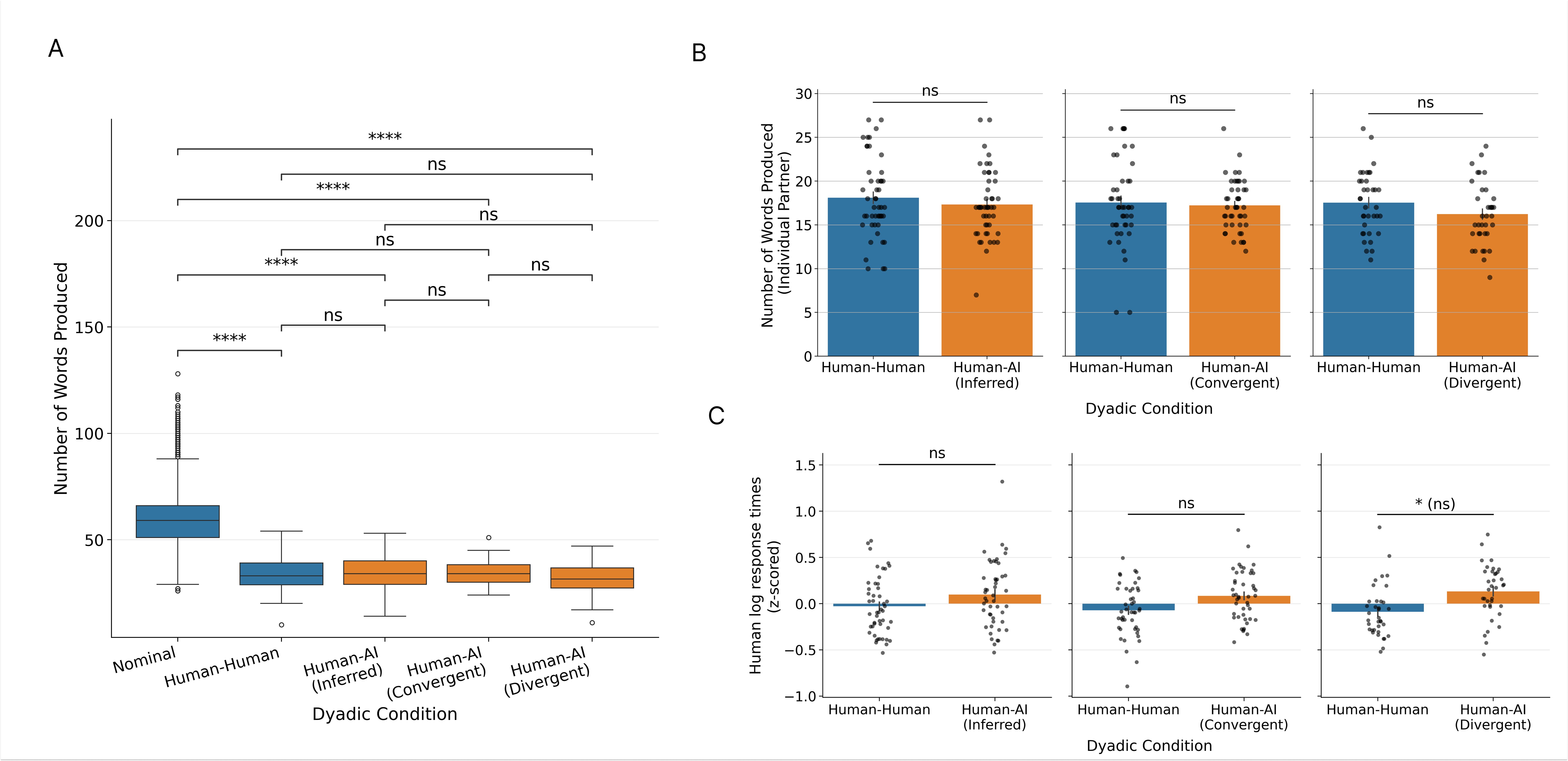}
\caption{
  \textbf{Performance in Semantic Foraging Task}
  \textbf{(A) Word Counts (Overall).} Word counts for individual and collaborative SFT conditions aggregating two categories (``animals'' and ``clothes'') are presented. We observe a collaborative inhibition effect, with nominal groups ($n = 3{,}846$, median $= 59.0$ [IQR: $51.0$, $66.0$]) significantly outperforming human--human dyads ($n = 96$, median $= 33.0$ [IQR: $28.8$, $39.0$], $U = 353{,}251.5$, $p < 0.001$) and human--AI dyads with all prompt variations (convergent: $n = 48$, median $= 34.0$ [IQR: $30.0$, $38.2$], $U = 180{,}267$, $p < 0.001$; divergent: $n = 46$, median $=31.5$ [IQR: $27.2$, $36.8$], $U = 173{,}486.5$, $p < 0.001$; inferred: $n = 50$, median $= 34.0$ [IQR: $29.0$, $40.0$], $U = 185{,}744$, $p < 0.001$). Dyadic word count comparisons correspond to Mann-Whitney U tests, annotated with Holm-Bonferroni correction.
  \textbf{(B) Word Counts (Within-Subjects).} Within-subjects comparisons of individual word counts revealed no significant differences between human--human and human--AI dyads across all prompts: convergent ($t = 0.540, p = 0.708$), divergent ($t = 1.938$, $p = 0.182$), and inferred ($t = 0.936$, $p = 0.708$). Within-subjects comparisons correspond to two-tailed paired t-tests with Holm-Bonferroni correction.
  \textbf{(C) Response Times (Within-Subjects).} Within-subjects comparisons of log-transformed z-scored response times revealed no significant differences between human and AI partners for inferred ($t = -1.336$, $p = 0.188$), convergent ($t = -1.924$, $p = 0.122$), or divergent ($t = -2.327$, $p = 0.078$) prompts. Response times were log-transformed and z-scored per-participant for this analysis. Within-subjects comparisons correspond to two-tailed paired t-tests with Holm-Bonferroni correction. 129 participants were included in the paired analyses, of which 48 completed dyadic SFT with inferred prompts, 36 with divergent prompts and 45 with convergent prompts. Significance codes: \textbf{*}\,$p \leq 0.05$, \textbf{**}\,$p \leq 0.01$, \textbf{***}\,$p \leq 0.001$, \textbf{****}\,$p \leq 0.0001$.
  }
\label{fig:s7}
\end{figure}

\begin{figure}
\centering
\includegraphics[width=\textwidth,height=0.45\textheight,keepaspectratio]{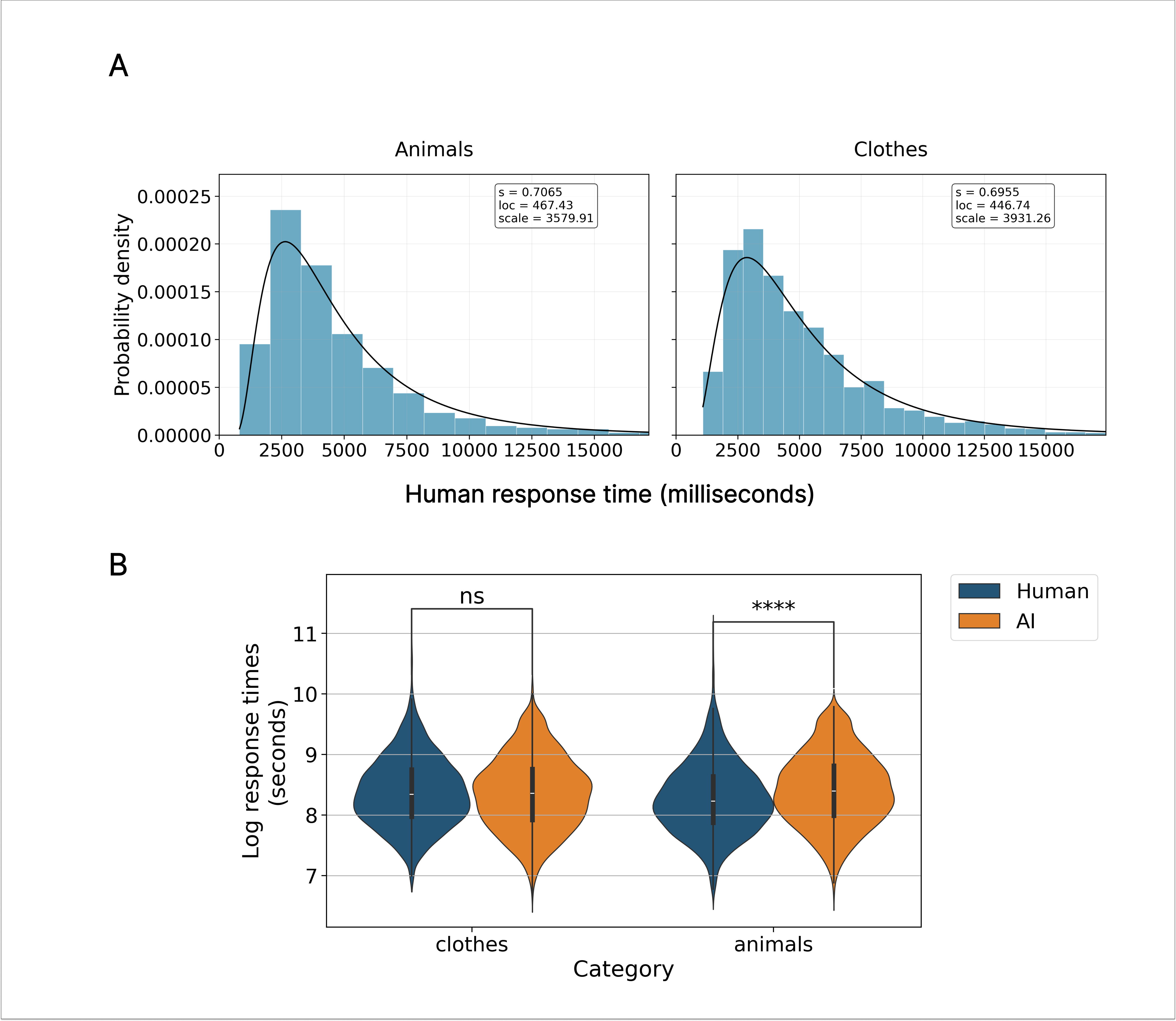}
\caption{
   \textbf{Artificial delay modeling.} \textbf{(A) Fitted distribution from initial data and observed final data from human--human response times}. In order to mimic human response times in the LLM-partner response times in our dyadic experiments, we collected an initial dataset from 36 human participants who performed a collaborative verbal fluency task with a human partner. We then modeled the response times for animals and clothes independently. For human--AI dyadic experiments, target AI delays were sampled from these distributions, truncated to values between 0.1 and 15.0 seconds. Finally, the sampled target delay was adjusted according to the response latency from the model provider, with the final delay corresponding to $\text{delay} = \max(0, \text{target\_delay} - \text{API\_latency})$.
   We compared gamma, Weibull and log-normal distributions for goodness of fit. Best fitting distributions were identified according to the Akaike information criterion (AIC). Selected distributions for both categories corresponded to log-normal, for animals with AIC = 33548.9 ($\exp(\mu) = 3579.91$, $\sigma = 0.7065$, location shift = 467.43 ms) and for clothes AIC = 27880.1 ($\exp(\mu) = 3931.26$, $\sigma = 0.6955$, location shift = 446.74 ms). Parameters were estimated using maximum likelihood estimation.
   \textbf{(B) Observed human and LLM response times.} Comparison of log-transformed response times in the final dataset showed that our delay implementation successfully matched human response times for clothes but resulted in slower AI responses for animals. For the animals category, human participants in human--human dyads had significantly shorter log-transformed response times (n = 1834, M = 8.27 [95\% CI: 8.25, 8.30]) than AI responses (n = 1106, M = 8.42 [95\% CI: 8.38, 8.45], t = -6.50, $p < .001$). In contrast, for clothes, there was no significant difference between human (n = 1511, M = 8.38 [95\% CI: 8.35, 8.40]) and AI responses (n = 1269, M = 8.36 [95\% CI: 8.33, 8.39], t = 0.72, $p = .47$). Comparisons correspond to two-tailed independent t-tests, with Holm-Bonferroni correction for multiple comparisons. Significance codes: \textbf{*}\,$p \leq 0.05$, \textbf{**}\,$p \leq 0.01$, \textbf{***}\,$p \leq 0.001$, \textbf{****}\,$p \leq 0.0001$.
}
\label{fig:s8}
\end{figure}

\begin{figure}
\centering
\includegraphics[width=\textwidth,height=0.45\textheight,keepaspectratio]{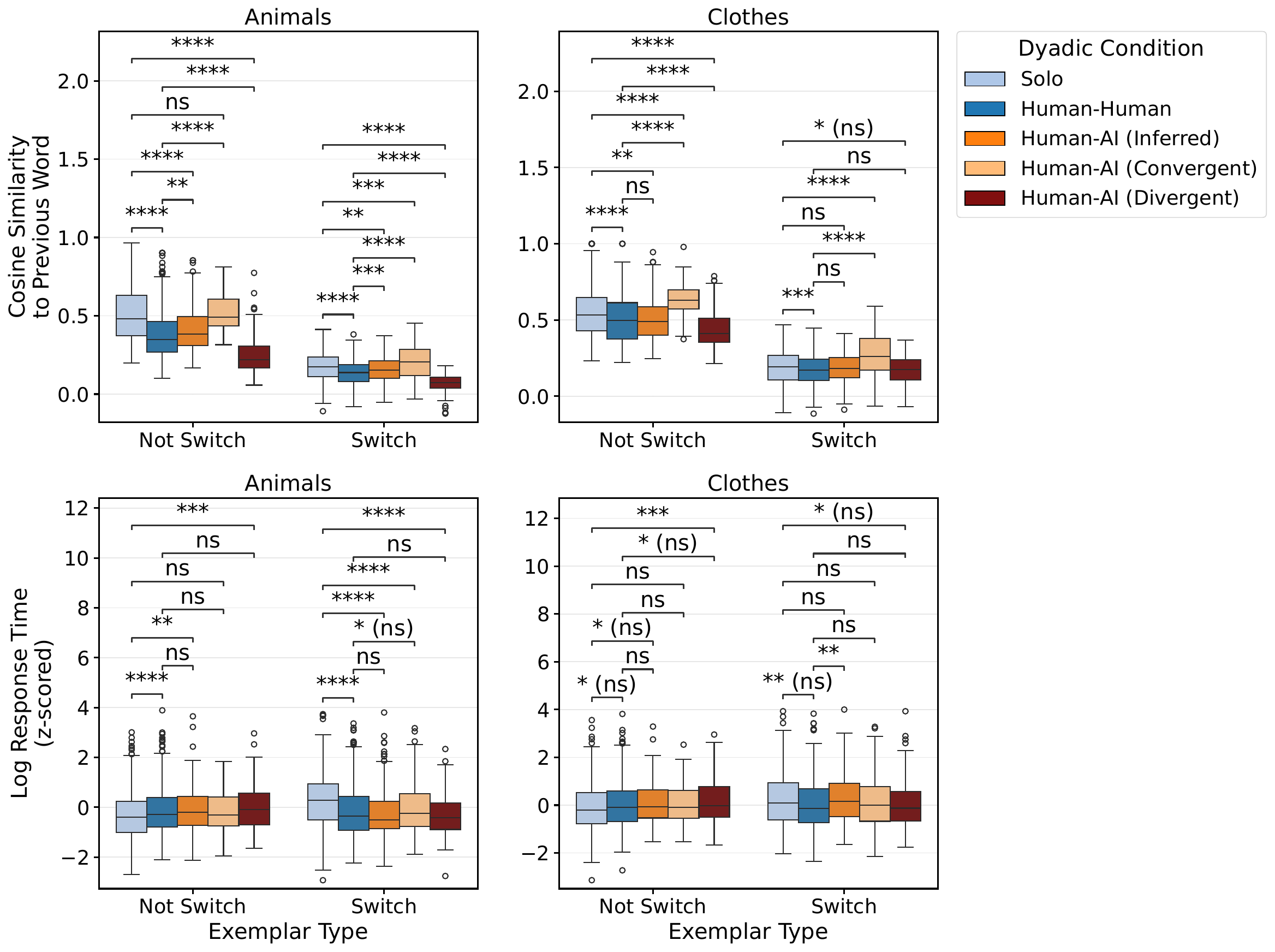}
\caption{
\textbf{Word Similarity and Response Time separated by Switching Behavior.}
Human log-transformed and z-scored response times as well as cosine similarity to the
previous word are presented by exemplar type in each category. ``Switch'' corresponds to words
classified as switches according to the ``similarity median'' switch determination method described in the Embeddings section (see above). ``Not
Switch'' corresponds to human-produced words that remain in the same cluster as the previous
word. Left panels correspond to the ``animals'' category, and right panels correspond to
``clothes''.
Solo human participants produce words that are significantly more similar to the previous word when remaining in the same cluster than participants in human--human dyads (animals: $U = 763{,}700.000$, $p < 0.001$; clothes: $U = 465{,}500.000$, $p < 0.001$) or in human--AI divergent dyads (animals: $U = 158{,}800.000$, $p < 0.001$; clothes: $U = 189{,}600.000$, $p < 0.001$). Participants in human--AI convergent dyads don't show significant differences in within-cluster similarity compared to individual participants in the ``animals'' category (animals: $U = 53{,}480.000$, $p = 0.235$), but produce significantly more similar words in the ``clothes'' category (clothes: $U = 38{,}240.000$, $p < 0.001$).
When switching, participants in human--AI convergent dyads produce words that are less dissimilar from the previous word than participants in solo (animals: $U = 120{,}800.000$, $p < 0.001$; clothes: $U = 94{,}280.000$, $p < 0.001$) or human--human conditions (animals: $U = 71{,}250.000$, $p < 0.001$; clothes: $U = 60{,}790.000$, $p < 0.001$).
When producing words within clusters, solo participants respond faster than participants in
human--human dyads (animals: $U = 475{,}400.000$, $p < 0.001$), human--AI inferred dyads
(animals: $U = 96{,}310.000$, $p = 0.017$) and human--AI divergent dyads (animals: $U =
71{,}480.000$, $p = 0.003$; clothes: $U = 112{,}700.000$, $p = 0.002$). When switching, participants in all collaborative conditions respond faster than solo participants in the ``animals'' category (human--human $U = 571{,}300.000$, $p < 0.001$; inferred $U = 178{,}300.000$, $p < 0.001$; convergent $U = 171{,}600.000$, $p < 0.001$; divergent $U = 71{,}000.000$, $p < 0.001$), whereas no collaborative condition differs significantly from solo in the
``clothes'' category. No significant differences in response times are found between human--human and human--AI convergent dyads in the two exemplar types.
This shows that participants from human--AI convergent dyads produce words that are more
similar to the previous word both within the cluster and when switching, without showing
significant differences in response time compared to human--human dyads.
Reported $p$-values are Holm--Bonferroni corrected across the 14 comparisons within each panel.
Significance codes: \textbf{*}\,$p \leq 0.05$, \textbf{**}\,$p \leq 0.01$, \textbf{***}\,$p \leq 0.001$, \textbf{****}\,$p \leq 0.0001$.
}

\label{fig:s9}
\end{figure}

\begin{figure}
\centering
\includegraphics[width=\textwidth]{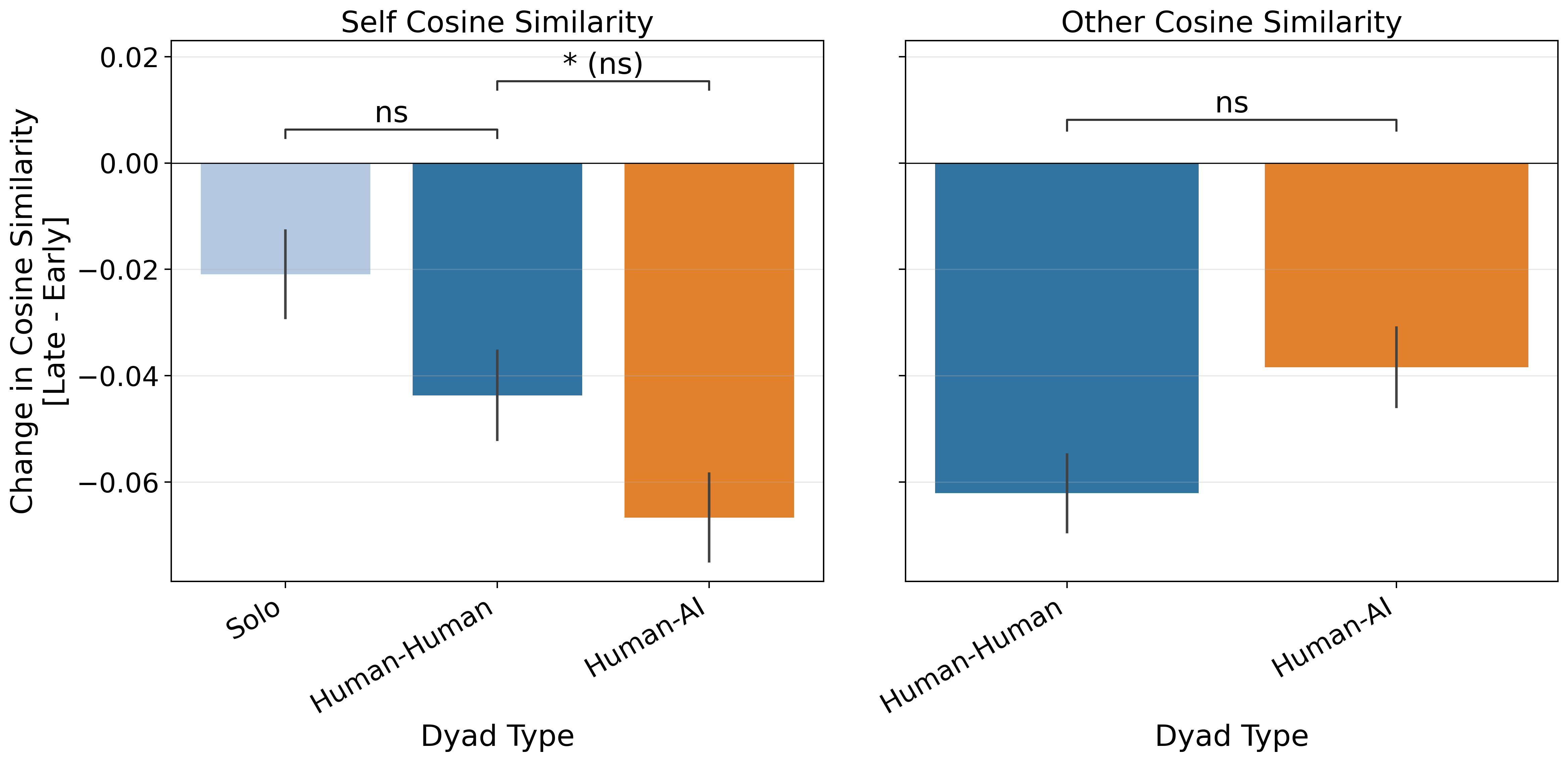}
\caption{
    \textbf{Change in Self- and Other-Similarity Between Late and Early SFT Phases.} The $y$-axis shows the mean difference in cosine similarity (late minus early phase, median split) between each participant's word embeddings and their own or their partner's previously produced words.
    \textbf{(A)}~\textbf{Self-similarity.}
    All three conditions exhibited significant decreases in self-similarity from the early to the late phase (Solo: mean $= -0.021 \pm 0.008$~SEM, $W = 1.53 \times 10^{3}$, $p = 1.68 \times 10^{-2}$; Human--Human: $-0.044 \pm 0.008$, $W = 5.51 \times 10^{3}$, $p = 1.11 \times 10^{-6}$; Human--AI: $-0.067 \pm 0.008$, $W = 1.65 \times 10^{3}$, $p = 1.05 \times 10^{-12}$; two-sided one-sample Wilcoxon signed-rank tests against zero). Human--AI participants showed a trend towards larger decline in self-similarity versus Human--Human participants, but this comparison did not reach significance after correction for multiple comparisons ($U = 1.57 \times 10^{4}$, $p = 3.63 \times 10^{-2}$ uncorrected, $p_{\mathrm{corrected}} = 7.27 \times 10^{-2}$, n.s.\ after Holm--Bonferroni correction). Solo and Human--Human conditions did not differ significantly ($U = 9.86 \times 10^{3}$, $p_{\mathrm{corrected}} = 0.114$, Holm--Bonferroni corrected).
    \textbf{(B)}~\textbf{Other-similarity.}
    Both collaborative conditions showed significant decreases in similarity to the partner's previous word (Human--Human: $-0.062 \pm 0.007$, $W = 3.58 \times 10^{3}$, $p = 1.66 \times 10^{-13}$; Human--AI: $-0.038 \pm 0.007$, $W = 2.81 \times 10^{3}$, $p = 1.57 \times 10^{-6}$; two-sided one-sample Wilcoxon signed-rank tests against zero). Human--Human participants displayed a trend towards a larger decrease in other-similarity than Human--AI participants, but this difference was not significant ($U = 1.22 \times 10^{4}$, $p_{\mathrm{corrected}} = 5.97 \times 10^{-2}$, Holm--Bonferroni corrected). Only sequences for the categories ``clothes'' and ``animals'' were included. Error bars indicate SEM. $*$\,$p < 0.05$; n.s., not significant (two-sided Mann--Whitney $U$ tests with Holm--Bonferroni correction for between-condition comparisons). Significance codes: \textbf{*}\,$p \leq 0.05$, \textbf{**}\,$p \leq 0.01$, \textbf{***}\,$p \leq 0.001$, \textbf{****}\,$p \leq 0.0001$.
 }
\label{fig:self-other-sim-diff}
\end{figure}

\begin{figure}
\centering
\includegraphics[width=\textwidth]{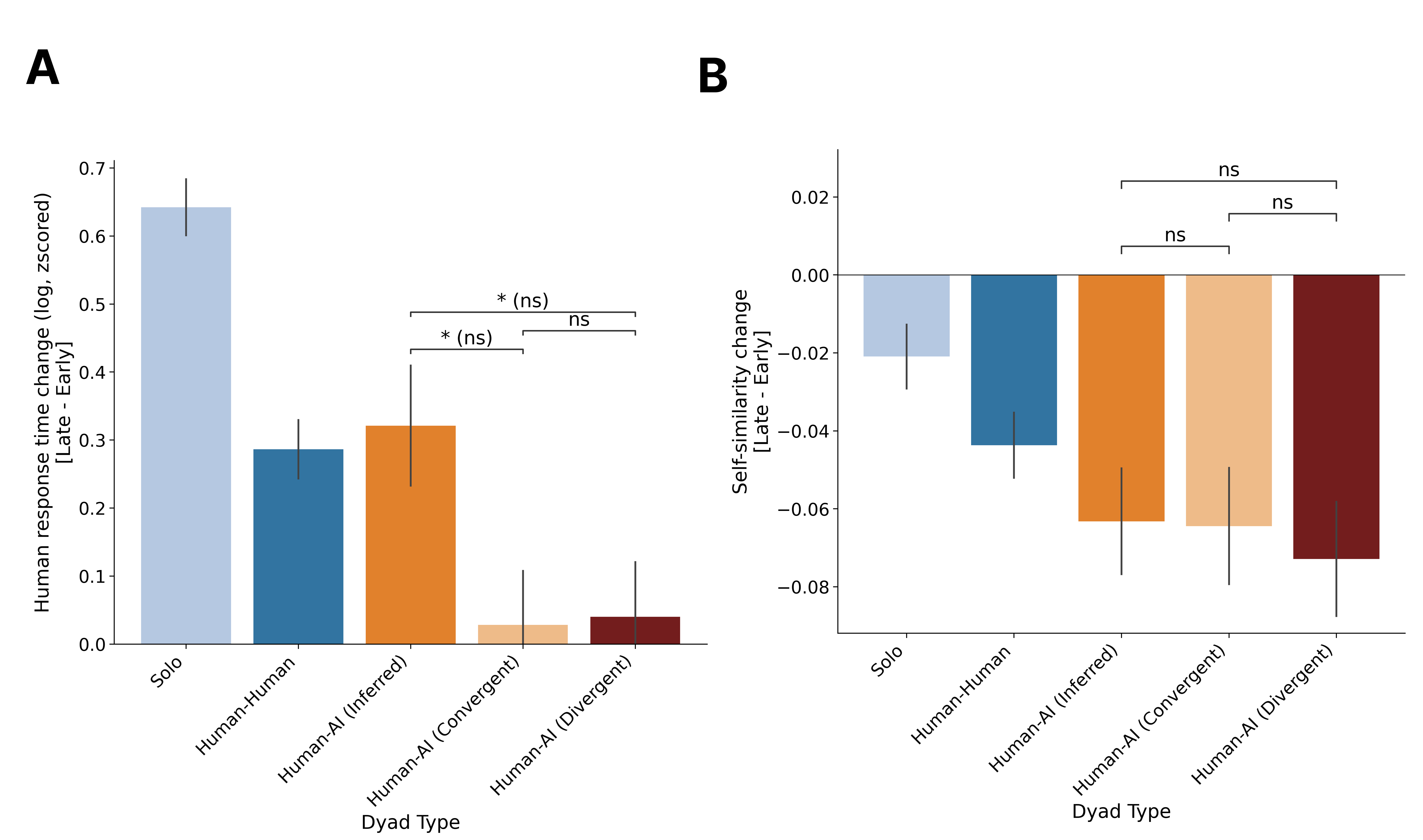}
\caption{
\textbf{Change in Response Time and Self-Similarity by Condition.} \textbf{(A)}~\textbf{Response Times.}
The $y$-axis shows the mean difference in log-transformed response time between the late and early phases of the SFT (median split). Interactions in convergent (mean $= 0.028 \pm 0.080$~SEM) and divergent ($0.040 \pm 0.081$) dyad conditions displayed negligible late-phase RT changes, with neither differing significantly from zero ($p > 0.50$ in both cases, two-sided one-sample Wilcoxon signed-rank tests). The inferred prompt condition retained a significant late-phase RT increase ($0.321 \pm 0.088$; $W = 293$, $p = 6.46 \times 10^{-4}$) and exhibited larger RT differences than both convergent ($U = 1.53 \times 10^{3}$, $p_{\mathrm{corrected}} = 5.54 \times 10^{-2}$) and divergent conditions ($U = 1.47 \times 10^{3}$, $p_{\mathrm{corrected}} = 5.54 \times 10^{-2}$), though neither comparison survived Holm--Bonferroni correction at $\alpha = 0.05$. Convergent and divergent conditions did not differ from each other ($p_{\mathrm{corrected}} = 0.96$).
\textbf{(B)}~\textbf{Self-similarity.} The $y$-axis shows the mean difference in self-similarity (embedding cosine similarity to the participant's own previous word) between the late and early SFT phases. All three Human--AI prompt conditions exhibited significant decreases in self-similarity from early to late phases (Inferred: $-0.063 \pm 0.014$, $p = 6.36 \times 10^{-6}$; Convergent: $-0.064 \pm 0.015$, $p = 8.57 \times 10^{-5}$; Divergent: $-0.073 \pm 0.015$, $p = 6.98 \times 10^{-6}$; two-sided one-sample Wilcoxon signed-rank tests against zero), and no pairwise differences among prompt conditions reached significance (all $p_{\mathrm{corrected}} > 0.39$, Holm--Bonferroni corrected). Only sequences for the categories ``clothes'' and ``animals'' were included. Error bars indicate SEM. $*$\,$p < 0.05$; $**$\,$p < 0.01$ (two-sided Mann--Whitney $U$ tests with Holm--Bonferroni correction for between-condition comparisons). Significance codes: \textbf{*}\,$p \leq 0.05$, \textbf{**}\,$p \leq 0.01$, \textbf{***}\,$p \leq 0.001$, \textbf{****}\,$p \leq 0.0001$.
}
\label{fig:s11}
\end{figure}

\FloatBarrier

\renewcommand{\refname}{Supplementary References}

\end{document}